\documentclass[fleqn,10pt]{wlscirep}
\usepackage[utf8]{inputenc}
\usepackage[T1]{fontenc}
\usepackage{soul}
\usepackage{lmodern}
\usepackage{kpfonts}
\usepackage[scaled]{beramono}
\usepackage{microtype}
\usepackage{parskip}
\usepackage{nicefrac}
\usepackage{wrapfig}
\usepackage{fontawesome5}
\usepackage{subcaption}

\usepackage{geometry}
\usepackage{float}
\usepackage{multicol}
\usepackage{wrapfig}
\usepackage{adjustbox}
\usepackage{array}
\usepackage{rotating}
\usepackage{longtable}
\usepackage{appendix}

\usepackage{amsmath}
\usepackage{amssymb}
\usepackage{amsfonts}
\usepackage{amsthm}

\usepackage{booktabs}
\usepackage{tabularx}
\usepackage{multirow}
\usepackage{threeparttable}
\usepackage{caption}
\usepackage{subcaption}
\usepackage{colortbl}
\usepackage{makecell}

\usepackage{graphicx}
\usepackage{svg}
\usepackage{tikz}
\usetikzlibrary{positioning,calc}

\usepackage{enumitem}
\usepackage{pifont}
\usepackage{fontawesome5}
\usepackage{siunitx}
\usepackage{algorithm}
\usepackage{algpseudocode}

\usepackage[table,dvipsnames,svgnames,xcdraw]{xcolor}
\definecolor{categorygray}{gray}{0.95}

\definecolor{cellbest}   {rgb}{0.47, 0.85, 0.55}
\definecolor{cellbetter}  {rgb}{0.67, 0.91, 0.72}
\definecolor{cellgood}    {rgb}{0.82, 0.95, 0.84}
\definecolor{cellneutral} {rgb}{0.99, 0.97, 0.83}
\definecolor{cellworse}   {rgb}{0.98, 0.82, 0.77}
\definecolor{cellbad}     {rgb}{0.96, 0.61, 0.55}

\definecolor{pillDark}   {HTML}{534AB7}
\definecolor{pillDeep}   {HTML}{3C3489}
\definecolor{boxBg}      {HTML}{F5F4FE}
\definecolor{boxBorder}  {HTML}{7F77DD}
\definecolor{shadowCol}  {HTML}{C8C4E8}
\definecolor{bodyText}   {HTML}{26215C}
\definecolor{cusppurple} {RGB}{90,70,160}
\definecolor{cusplight}  {RGB}{245,243,252}
\definecolor{cuspborder} {RGB}{140,120,200}
\definecolor{myorange}   {RGB}{37,79,150}

\usepackage[most]{tcolorbox}
\tcbuselibrary{skins,breakable}

\newtcolorbox{benchbox}[1][]{
  enhanced,
  breakable,
  colback        = boxBg,
  colframe       = boxBorder,
  arc            = 4pt,
  boxrule        = 0.8pt,
  top            = 10pt,
  bottom         = 0pt,
  left           = 10pt,
  right          = 10pt,
  fontupper      = \small\color{bodyText},
  drop shadow    = {shadowCol!70!white},
  borderline west= {4pt}{0pt}{pillDark},
  attach boxed title to top left = {xshift=10pt, yshift=-\tcboxedtitleheight/2},
  boxed title style = {
    colback   = pillDeep,
    colframe  = pillDeep,
    arc       = 10pt,
    boxrule   = 0pt,
    fonttitle = \bfseries\small,
    coltitle  = white,
    left      = 10pt,
    right     = 10pt,
    top       = 3pt,
    bottom    = 3pt,
  },
  title = {#1},
  lowerbox        = visible,
  colbacklower    = boxBg!50!white,
  fontlower      = \small\itshape\color{pillDeep},
  bottomtitle     = 0pt,
}

\newtcolorbox{airemovesbox}[2][]{
  benchbox={#2},
  colbacklower = boxBg!50!white, 
  fonttlower   = \small\itshape\color{pillDeep},
}

\newtcolorbox{humanremovesbox}[2][]{
  benchbox={#2},
  colbacklower = boxBg!50!white,
  fonttlower   = \small\itshape\color{pillDeep},
}

\newtcolorbox{takeawaybox}[1][]{
  colback=yellow!3!white,
  colframe=yellow!50!black,
  boxrule=0.6pt,
  arc=2mm,
  left=6pt,
  right=6pt,
  top=6pt,
  bottom=4pt,
  fontupper=\small,
  before skip=6pt,
  after skip=6pt,
  overlay={
    \node[anchor=north west, font=\small, xshift=2pt, yshift=-2pt]
    at (frame.north west) {\faLightbulb};
  },
  #1
}

\newcommand{\llmicon}[1]{\raisebox{-0.15\height}{\includegraphics[height=8pt]{#1}}}

\usepackage{xspace}

\usepackage{xcolor}
\usepackage{xspace}

\newcommand{\CUSP}{\textbf{\textcolor{RoyalPurple}{\textsc{CUSP}}\xspace}}

\definecolor{pastelblue}{HTML}{E3F2FD}
\definecolor{pastelgreen}{HTML}{E8F5E9}
\definecolor{pastellavender}{HTML}{EDE7F6}
\definecolor{pastelpeach}{HTML}{FFF3E0}
\definecolor{pastelrose}{HTML}{FCE4EC}
\definecolor{pastelmint}{HTML}{E0F2F1}

\newtcolorbox{reviewbox}[2][]{%
  enhanced,
  breakable,
  colback=#2,
  colframe=#2!60!black!20,
  boxrule=0pt,
  arc=4mm,
  outer arc=4mm,
  left=5mm, right=5mm, top=4mm, bottom=4mm,
  shadow={1.5mm}{-1.5mm}{0mm}{black!8},
  attach boxed title to top left={yshift=-3mm, xshift=5mm},
  boxed title style={
    colback=white,
    colframe=white,
    arc=2mm,
    outer arc=2mm,
    boxrule=0pt,
    left=2mm, right=2mm, top=1mm, bottom=1mm,
    shadow={1mm}{-1mm}{0mm}{black!6},
  },
  fonttitle=\normalsize\bfseries,
  coltitle=black,
  title={Human Expert Review~---~#1},
}

\definecolor{artifactbg}{HTML}{ffffff}
\definecolor{artifactframe}{HTML}{d0d0d0}
\definecolor{artifacttitle}{HTML}{333333}

\definecolor{codegreen}{HTML}{228B22}
\definecolor{codeblue}{HTML}{0000CD}
\definecolor{codegray}{HTML}{808080}

\definecolor{promptbg}{HTML}{f9f6f2}
\definecolor{promptframe}{HTML}{d4c4b0}
\definecolor{promptaccent}{HTML}{5a4a3a}

\definecolor{artifactbg}{HTML}{ffffff}
\definecolor{artifactframe}{HTML}{d0d0d0}
\definecolor{artifacttitle}{HTML}{333333}

\definecolor{codegreen}{HTML}{228B22}
\definecolor{codeblue}{HTML}{0000CD}
\definecolor{codegray}{HTML}{808080}

\definecolor{promptbg}{HTML}{f9f6f2}
\definecolor{promptframe}{HTML}{d4c4b0}
\definecolor{promptaccent}{HTML}{5a4a3a}

\usepackage{upquote}
\usepackage{listingsutf8}
\lstdefinestyle{artifactstyle}{
    basicstyle=\ttfamily\scriptsize,
    backgroundcolor=\color{artifactbg},
    commentstyle=\color{codegray}\itshape,
    keywordstyle=\color{codeblue},
    stringstyle=\color{codegreen},
    numberstyle=\tiny\color{codegray},
    breakatwhitespace=false,
    breaklines=true,
    keepspaces=true,
    numbers=left,
    numbersep=8pt,
    showspaces=false,
    showstringspaces=false,
    showtabs=false,
    tabsize=4,
    xleftmargin=12pt,
    framexleftmargin=12pt,
    aboveskip=0pt,
    belowskip=0pt,
    inputencoding=utf8,
    columns=fullflexible,
    upquote=true,
    literate=
        {→}{{$\rightarrow$}}1
        {←}{{$\leftarrow$}}1
        {↔}{{$\leftrightarrow$}}1
        {≤}{{$\leq$}}1
        {≥}{{$\geq$}}1
        {≠}{{$\neq$}}1
        {−}{{-}}1
        {—}{{--}}1
        {–}{{-}}1
        {"}{{\textquotedblleft}}1
        {"}{{\textquotedblright}}1
        {'}{{\textquoteleft}}1
        {'}{{\textquoteright}}1
        {…}{{...}}1
        {×}{{$\times$}}1
        {÷}{{$\div$}}1
        {±}{{$\pm$}}1
        {∞}{{$\infty$}}1
        {α}{{$\alpha$}}1
        {β}{{$\beta$}}1
        {γ}{{$\gamma$}}1
        {δ}{{$\delta$}}1
        {ε}{{$\epsilon$}}1
        {λ}{{$\lambda$}}1
        {π}{{$\pi$}}1
        {σ}{{$\sigma$}}1
        {∑}{{$\sum$}}1
        {∏}{{$\prod$}}1
        {√}{{$\sqrt{}$}}1
        {∈}{{$\in$}}1
        {∉}{{$\notin$}}1
        {⊂}{{$\subset$}}1
        {⊃}{{$\supset$}}1
        {∩}{{$\cap$}}1
        {∪}{{$\cup$}}1
        {∀}{{$\forall$}}1
        {∃}{{$\exists$}}1
        {¬}{{$\neg$}}1
        {∧}{{$\land$}}1
        {∨}{{$\lor$}}1,
}

\lstdefinestyle{promptstyle}{
    basicstyle=\small\scriptsize,
    backgroundcolor=\color{promptbg},
    breakatwhitespace=false,
    breaklines=true,
    breakindent=0pt,
    breakautoindent=false,
    keepspaces=true,
    numbers=none,
    showspaces=false,
    showstringspaces=false,
    showtabs=false,
    tabsize=4,
    aboveskip=0pt,
    belowskip=0pt,
    xleftmargin=0pt,
    framexleftmargin=0pt,
    extendedchars=true,
    inputencoding=utf8,
    upquote=true,
    literate=
        {→}{{$\rightarrow$}}1
        {←}{{$\leftarrow$}}1
        {↔}{{$\leftrightarrow$}}1
        {≤}{{$\leq$}}1
        {≥}{{$\geq$}}1
        {≠}{{$\neq$}}1
        {−}{{-}}1
        {—}{{--}}1
        {–}{{-}}1
        {"}{{\textquotedblleft}}1
        {"}{{\textquotedblright}}1
        {'}{{\textquoteleft}}1
        {'}{{\textquoteright}}1
        {…}{{...}}1
        {×}{{$\times$}}1
        {÷}{{$\div$}}1
        {±}{{$\pm$}}1
        {∞}{{$\infty$}}1
        {α}{{$\alpha$}}1
        {β}{{$\beta$}}1
        {γ}{{$\gamma$}}1
        {δ}{{$\delta$}}1
        {ε}{{$\epsilon$}}1
        {λ}{{$\lambda$}}1
        {π}{{$\pi$}}1
        {σ}{{$\sigma$}}1
        {∑}{{$\sum$}}1
        {∏}{{$\prod$}}1
        {√}{{$\sqrt{}$}}1
        {∈}{{$\in$}}1
        {∉}{{$\notin$}}1
        {⊂}{{$\subset$}}1
        {⊃}{{$\supset$}}1
        {∩}{{$\cap$}}1
        {∪}{{$\cup$}}1
        {∀}{{$\forall$}}1
        {∃}{{$\exists$}}1
        {¬}{{$\neg$}}1
        {∧}{{$\land$}}1
        {∨}{{$\lor$}}1,
}

\newtcblisting{artifact}[2][python]{%
    enhanced,
    breakable,
    colback=artifactbg,
    colframe=artifactframe,
    boxrule=0.5pt,
    arc=3mm,
    outer arc=3mm,
    left=0mm, right=3mm, top=2mm, bottom=2mm,
    shadow={1.5mm}{-1.5mm}{0mm}{black!10},
    attach boxed title to top left={yshift=-3mm, xshift=5mm},
    boxed title style={
        colback=white,
        colframe=artifactframe,
        arc=2mm,
        outer arc=2mm,
        boxrule=0.5pt,
        left=2mm, right=2mm, top=1mm, bottom=1mm,
    },
    fonttitle=\normalsize\bfseries,
    coltitle=artifacttitle,
    title={#2},
    listing only,
    listing options={
        style=artifactstyle,
        language=#1,
    },
}

\newtcblisting{prompt}[1]{%
    enhanced,
    breakable,
    colback=promptbg,
    colframe=promptframe,
    boxrule=0.5pt,
    arc=3mm,
    outer arc=3mm,
    left=5mm, right=5mm, top=4mm, bottom=4mm,
    shadow={1.5mm}{-1.5mm}{0mm}{black!8},
    attach boxed title to top left={yshift=-3mm, xshift=5mm},
    boxed title style={
        colback=white,
        colframe=promptframe,
        arc=2mm,
        outer arc=2mm,
        boxrule=0.5pt,
        left=2mm, right=2mm, top=1mm, bottom=1mm,
    },
    fonttitle=\normalsize,
    coltitle=promptaccent,
    title={#1},
    listing only,
    listing options={
        style=promptstyle,
    },
}

\title{Scientific reasoning does not reliably translate into scientific forecasting in frontier AI}

\author[1,*]{Sean Wu}
\author[2,*]{Pan Lu}
\author[1 ]{Yupeng Chen}
\author[3 ]{Jonathan Bragg}
\author[4 ]{Yutaro Yamada}
\author[3 ]{Peter Clark}
\author[1 ]{David Clifton}
\author[1,$\dagger$]{Philip Torr}
\author[2,$\dagger$]{James Zou}
\author[1,$\dagger$]{Junchi Yu}

\affil[1]{Department of Engineering Science, University of Oxford, Oxford, UK}
\affil[2]{Department of Computer Science, Stanford University, Stanford, CA, USA}
\affil[3]{Allen Institute for AI, Seattle, WA, USA}
\affil[4]{Sakana AI, Tokyo, Japan}

\affil[*]{Equal contribution}
\affil[$\dagger$]{To whom the correspondence should be addressed:\newline
Junchi Yu (\href{junchi.yu@eng.ox.ac.uk}{\textcolor{blue}{junchi.yu@eng.ox.ac.uk}});\newline
Philip Torr (\href{philip.torr@eng.ox.ac.uk}{\textcolor{blue}{philip.torr@eng.ox.ac.uk}});\newline
James Zou (\href{jamesz@stanford.edu}{\textcolor{blue}{jamesz@stanford.edu}})
}

\begin{abstract}
AI systems are increasingly used to support forward-looking scientific judgment, but it remains unclear whether they can form reliable expectations about future scientific advances. Here we show that strong scientific reasoning does not reliably translate into accurate forecasting of future scientific advances. To study this question, we introduce \CUSP, a temporally grounded evaluation suite for event-level scientific forecasting across eight scientific disciplines. Across six frontier AI models, we observe a striking asymmetry in forecasting performance together with systematic error patterns. Models often identify plausible mechanisms underlying future scientific advances, yet perform near chance on feasibility assessment, generate solution strategies that only weakly align with realized advances, and systematically predict scientific advances later than they become publicly observable. Providing additional pre-cutoff scientific knowledge improves performance but does not eliminate these forecasting limitations. These findings suggest that current AI systems possess substantial retrospective scientific competence but limited forward-looking predictive capability. Scientific forecasting should therefore be evaluated as a complementary dimension of AI scientific capability when deploying AI systems for research prioritization and scientific decision-making.
\end{abstract}

\begin{document}

\flushbottom

\renewcommand{\rmdefault}{phv}
\renewcommand{\sfdefault}{phv}
\renewcommand{\familydefault}{\sfdefault}
\fontsize{11}{13.2}\selectfont
\normalfont
\setlength{\parskip}{6pt}

\maketitle
\thispagestyle{empty}

\fancyhf{}
\rfoot{\small\sffamily\bfseries\thepage}
\renewcommand{\headrulewidth}{0pt}
\renewcommand{\footrulewidth}{0pt}

\section*{Introduction}

Anticipating future scientific progress plays an important role in scientific decision making \cite{uzzi2013atypical,shapere1964structure}. 
Throughout modern science, researchers have used existing scientific knowledge to form expectations about future advances and guide scientific efforts. 
Empirical regularities such as Moore’s Law \cite{moore1965moore} in semiconductors and scaling relationships in deep learning \cite{kaplan2020scaling,xiao2025densing} demonstrate that observations of past progress can sometimes be transformed into expectations about future developments. 
These expectations have informed research roadmaps, funding decisions, and scientific exploration \cite{park2023papers,miles2010development}.

Advances in large language models demonstrate that artificial intelligence (AI) is becoming increasingly embedded in scientific discovery \cite{jumper2021highly, abramson2024accurate, novikov2025alphaevolve, merchant2023scaling, swanson2025virtual}. 
Across domains including biology, chemistry, and materials science, AI systems are accelerating research in literature synthesis \cite{yamada2025ai}, experimental design \cite{huang2025biomni}, and data analysis \cite{gao2024empowering}.
As AI becomes more integrated into scientific workflows, it is also beginning to inform future research directions \cite{marwitz2026predicting,krenn2023forecasting}.
Research organizations are beginning to explore AI-assisted systems for research prioritization and resource allocation \cite{adam2025ai,rees2026could}, while scientists use AI to evaluate competing directions and prioritize costly experiments \cite{penades2025ai,gottweis2025towards}.
Many of these applications require judgment about how future scientific fields may evolve and which advances are most likely to occur. 
If such forward-looking scientific judgments are unreliable, scientific attention, resources, and efforts may be misdirected \cite{messeri2024artificial}.

In this study, we investigate the forward-looking scientific judgment of frontier AI models by focusing on their ability to forecast scientific advances beyond their knowledge cutoff.
This setting provides a direct and measurable way to evaluate whether models' expectations align with subsequently realized scientific outcomes.
Despite strong performance on a wide range of scientific reasoning tasks, it remains unclear whether current AI models can reliably anticipate future scientific advances. 
Existing evaluations assess scientific knowledge understanding, expert-level question answering, and scientific problem solving \cite{lu2022learn, rein2024gpqa, bragg2025astabench, phan2025humanity}. 
While these studies measure how effectively models reason about established scientific knowledge, they provide limited insight into whether models can form reliable expectations about scientific advances beyond the knowledge cutoff.
Recent work on hypothesis generation \cite{liu2025researchbench} and scientific contribution prediction \cite{ajith2026prescience} begins to explore future-oriented aspects of scientific reasoning. 
However, these studies investigate the promise of AI-generated ideas or research directions, rather than whether model expectations align with subsequently realized scientific advances.


To study this question, we introduce \CUSP\ (\textbf{C}utoff-conditioned \textbf{U}nseen \textbf{S}cientific \textbf{P}rogress), a new temporally grounded evaluation suite comprising 4,760 verifiable scientific events across 8 disciplines.
\CUSP\ operationalizes the anticipation of future scientific advances through four complementary dimensions of event-level scientific forecasting: 
whether an advance will be realized (feasibility assessment), 
which technical pathway will enable it (mechanistic forecasting), 
how a viable solution can be designed (solution design), and when the event will become publicly observable (temporal prediction).

Using \CUSP, we evaluate six frontier AI models and identify a striking asymmetry in how they anticipate future scientific advances.
Models often identify plausible mechanisms underlying future advances and generate technically detailed scientific solutions, yet remain substantially less reliable at forecasting whether advances will occur, when they will become publicly observable, and how they will ultimately be realized.

This asymmetry is consistent across frontier AI models we evaluate. 
Performance is strongest on mechanistic forecasting, but substantially weaker on open-ended solution design, feasibility assessment, and temporal prediction. Although models frequently generate scientifically plausible solution strategies, these strategies remain only weakly aligned with the approaches underlying realized scientific advances. 
Feasibility assessment remains near chance across models, while temporal prediction exhibits large and systematic errors.

We further find that these limitations cannot be explained solely by access to scientific knowledge. 
Providing additional pre-cutoff scientific knowledge consistently improves forecasting performance, revealing a knowledge gap in how models retrieve and utilize existing knowledge. 
However, substantial forecasting limitations remain even when relevant historical information is provided, suggesting that reliable scientific forecasting requires more than access to and reasoning over existing scientific knowledge.

Beyond limited accuracy, forecasting failures exhibit strong systematic patterns.
Models exhibit persistent response biases, substantial overconfidence, and a consistent tendency to predict scientific advances later than they ultimately become publicly observable. 
Together, these findings suggest that forecasting failures arise not only from incomplete knowledge but also from systematic limitations in how current AI systems form calibrated expectations about future scientific advances.

Taken together, our findings suggest that current frontier AI systems possess substantial retrospective scientific competence but limited forward-looking predictive capability. 
While they can reason effectively about existing scientific knowledge, they remain substantially less effective at anticipating future scientific advances.
As AI systems become increasingly involved in research prioritization, opportunity evaluation, and scientific decision-making, understanding these limitations becomes as important as measuring scientific reasoning itself. \CUSP\ is publicly available at \href{https://huggingface.co/datasets/SeanWu25/CUSP}{Huggingface}, and source code is available at \href{https://github.com/SeanWu25/cusp-scientific-foresight}{Github}.

\begin{figure*}[h]
    \centering
    \includegraphics[width=\textwidth]{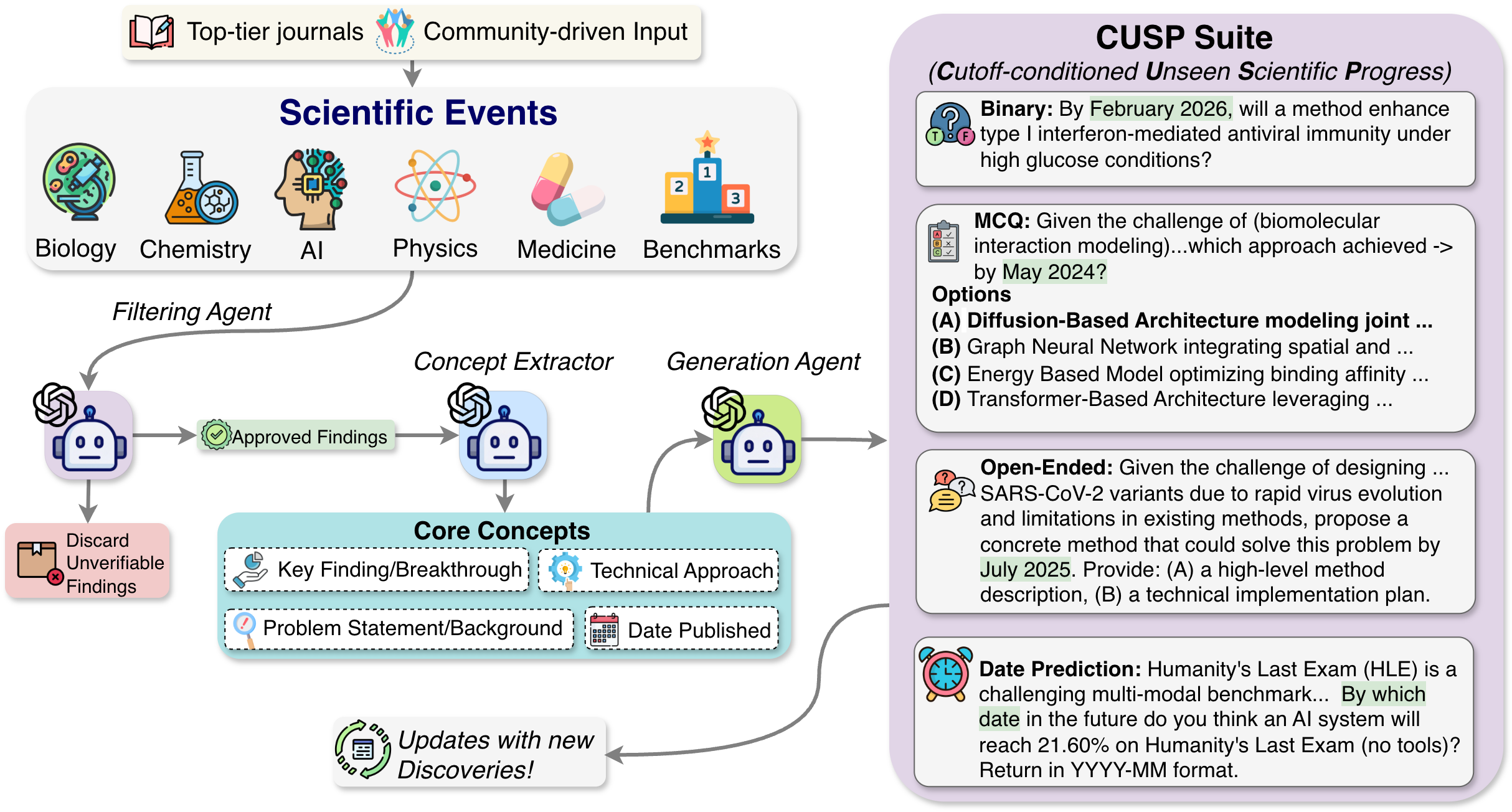}
    \caption{
    We construct \CUSP\ suite by aggregating scientific events from top-tier journals and community-driven sources across multiple domains. \CUSP\ is continuously updated with newly published discoveries, enabling an event-level, cross-disciplinary, and temporally grounded evaluation of AI systems’ ability to forecast scientific events beyond a knowledge cutoff.
    }
    \label{fig:cusp_pipeline}
\end{figure*}

\section*{Results}
We first introduce \CUSP, a temporally grounded evaluation suite that operationalizes anticipation of scientific advances as event-level scientific forecasting. 
We then evaluate proprietary models, including GPT-5.4 \cite{openai2026gpt54}, GPT-4o \cite{hurst2024gpt}, and Claude Sonnet 4.5 (Claude S4.5) \cite{claudesonnet}, with open-source models such as LLaMA-3.3-70B-Instruct (LLaMA 3.3) \cite{grattafiori2024llama}, GPT-OSS-20B (GPT-OSS) \cite{agarwal2025gpt}, and DeepSeek R1 \cite{guo2025deepseek}.

\subsection*{Operationalizing anticipation of scientific advances by event-level scientific forecasting}
To operationalize the anticipation of future scientific advances, we introduce \CUSP, a temporally grounded evaluation suite built from 4,760 verifiable scientific events spanning January 2024 to March 2026 across eight scientific disciplines.
These scientific events are collected from top-tier journals and community-driven repositories, with publication dates cross-verified across multiple academic platforms to establish strict temporal references of their first publicly observable dates.

Each scientific event is transformed into four complementary forecasting tasks that evaluate different aspects of future scientific events (\hyperref[fig:cusp_pipeline]{Fig.~\ref*{fig:cusp_pipeline}}). 
Binary prediction (Binary) evaluates feasibility assessment by requiring models to predict whether a scientific advance will be realized. 
Multiple-choice questions (MCQ) evaluate mechanistic forecasting by requiring models to identify which technical pathway will enable the advance. 
Free-response questions (FRQ) evaluate solution design by assessing whether models can generate a viable approach for achieving the advance. 
Date prediction (Date) evaluates temporal prediction by requiring models to predict when a scientific event will become publicly observable.

To ensure that these tasks evaluate forecasting rather than post-hoc reconstruction, all generated questions are reviewed by a critic agent that performs task-specific question validation. 
The agent verifies that forecasting targets are faithful to the source events, objectively verifiable, and free from unsupported perturbations or trivial post-event cues. 
For MCQ, it additionally checks the question choices to ensure that the correct choice is supported by the source event while distractors remain scientifically plausible alternatives rather than easily eliminated options. 
For binary prediction, it removes perturbations that are under-specified or insufficiently distinguishable from the original claim (\hyperref[fig:eval_pipeline]{Fig.~\ref*{fig:eval_pipeline}a}).

A human study on a randomly sampled 200 scientific events shows that the critic agent achieves high precision across question formats (Binary: 93.7\%, MCQ: 85.8\%, Date: 96.8\%, FRQ: 89.5\%), indicating that the retained questions are highly consistent with human judgment.
The resulting evaluation suite comprises 17,429 forecasting questions, including 6,411 binary questions, 4,128 MCQs, 4,135 FRQs, and 2,755 date prediction questions. Representative examples and model responses are shown in \hyperref[fig:examples]{Fig.~\ref*{fig:examples}a}.

Forecasting performance is evaluated using a two-track framework consisting of an outcome track for deterministic forecasting tasks (Binary, MCQ, and Date) and a reasoning track for open-ended solution design (FRQ) (Fig.~\ref{fig:eval_pipeline}b). 
The outcome track compares model forecasts against realized scientific outcomes, whereas the reasoning track evaluates the scientific quality of generated solution strategies. 
Together, these components enable a systematic evaluation of event-level scientific forecasting across feasibility, mechanism, solution strategy, and public-observation timing.

\begin{table*}[t]
\centering
\caption{\textbf{Model performance on \CUSP\, with change across the training cutoff.}
Overall performance on \CUSP\ post-cutoff instances. Each metric shows the post-cutoff value; the small grey number beneath ($\Delta$) is the change from the pre-cutoff partition (post\,$-$\,pre) for models that have one.
\emph{Binary} and \emph{MCQ} report merged accuracy on original and negation-flipped variants, corrected for directional response bias (chance\,$=$\,0.50 and 0.25).
\emph{FRQ} is an LLM rubric score (0--10); \emph{Pass\,\%} the fraction scoring $\geq5$.
\emph{Temporal prediction} reports mean predicted publication date (ground-truth mean 2025-02), signed error in months (predicted\,$-$\, actual; positive\,$=$\, later than truth), the fraction of predictions within $N$ calendar months of truth, and an exponential-decay score $e^{-0.1|\Delta t|}$ (Date; 1.0\,$=$\, exact month).
All models show positive signed error, systematically over-estimating how recently papers were published.
Models are sorted by MCQ accuracy; bold marks the best value per column.
$\Delta$ shading: \colorbox{cellgood}{\strut green} $>+0.05$ improvement, \colorbox{cellworse}{\strut orange} $<-0.03$ degradation, \colorbox{cellbad}{\strut red} $<-0.10$ strong degradation.
Models with cutoff $\leq$\,Dec 2023 (GPT-4o, LLaMA\,3.3) have no pre-cutoff instances ($\Delta=$\,---); $n_{\text{pre}}$ is the number of binary-task pre-cutoff instances. Benchmark papers span Jan~2024--Mar~2026.}
\label{tab:main_results}  

\begin{adjustbox}{max width=\textwidth}
\footnotesize\renewcommand{\arraystretch}{1.3}
\begin{tabular}{@{}ll r cc cc l cc rrrr c@{}}
\toprule
 & & & \multicolumn{2}{c}{Closed-form\,$\uparrow$} & \multicolumn{2}{c}{Open-ended\,$\uparrow$} & \multicolumn{8}{c}{Temporal prediction} \\
\cmidrule(lr){4-5}\cmidrule(lr){6-7}\cmidrule(lr){8-15}
 & & & & & & & & \multicolumn{2}{c}{Signed error (mo)} & \multicolumn{4}{c}{Within $N$ months (\%)\,$\uparrow$} & \\
\cmidrule(lr){9-10}\cmidrule(lr){11-14}
\textbf{Model} & \textbf{Cutoff} & $n_{\text{pre}}$ & \textbf{Binary} & \textbf{MCQ} & \textbf{FRQ} & \textbf{Pass\,\%} & \textbf{Mean pred.} & \textbf{Mean} & \textbf{Median} & $\leq 3$ & $\leq 6$ & $\leq 12$ & $\leq 24$ & \textbf{Date\,$\uparrow$} \\
\midrule

\llmicon{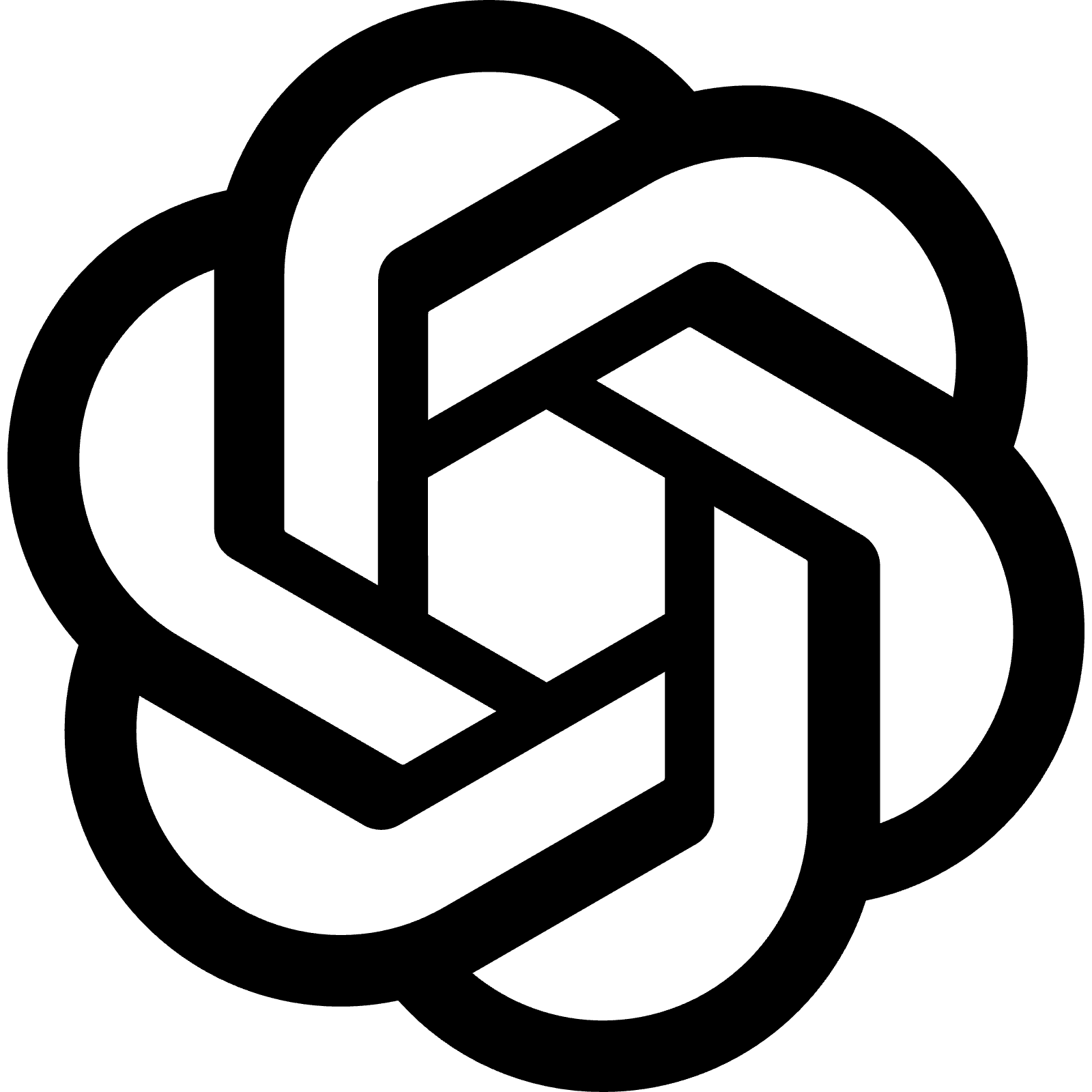} GPT-5.4 & Aug 2025 & 4577
  & 0.491 & \textbf{0.792} & \textbf{5.04} & \textbf{60.0}
  & 2026-08 & +9.5 & +6.5 & 15.0 & 22.5 & 35.7 & 54.4 & 0.270 \\
  & & & {\scriptsize\textcolor{gray}{$-$0.012}} & \cellcolor{cellworse}{\scriptsize\textcolor{gray}{$-$0.038}} & {\scriptsize\textcolor{gray}{0.00}} & & & & & & & & & {\scriptsize\textcolor{gray}{$+$0.039}} \\

\llmicon{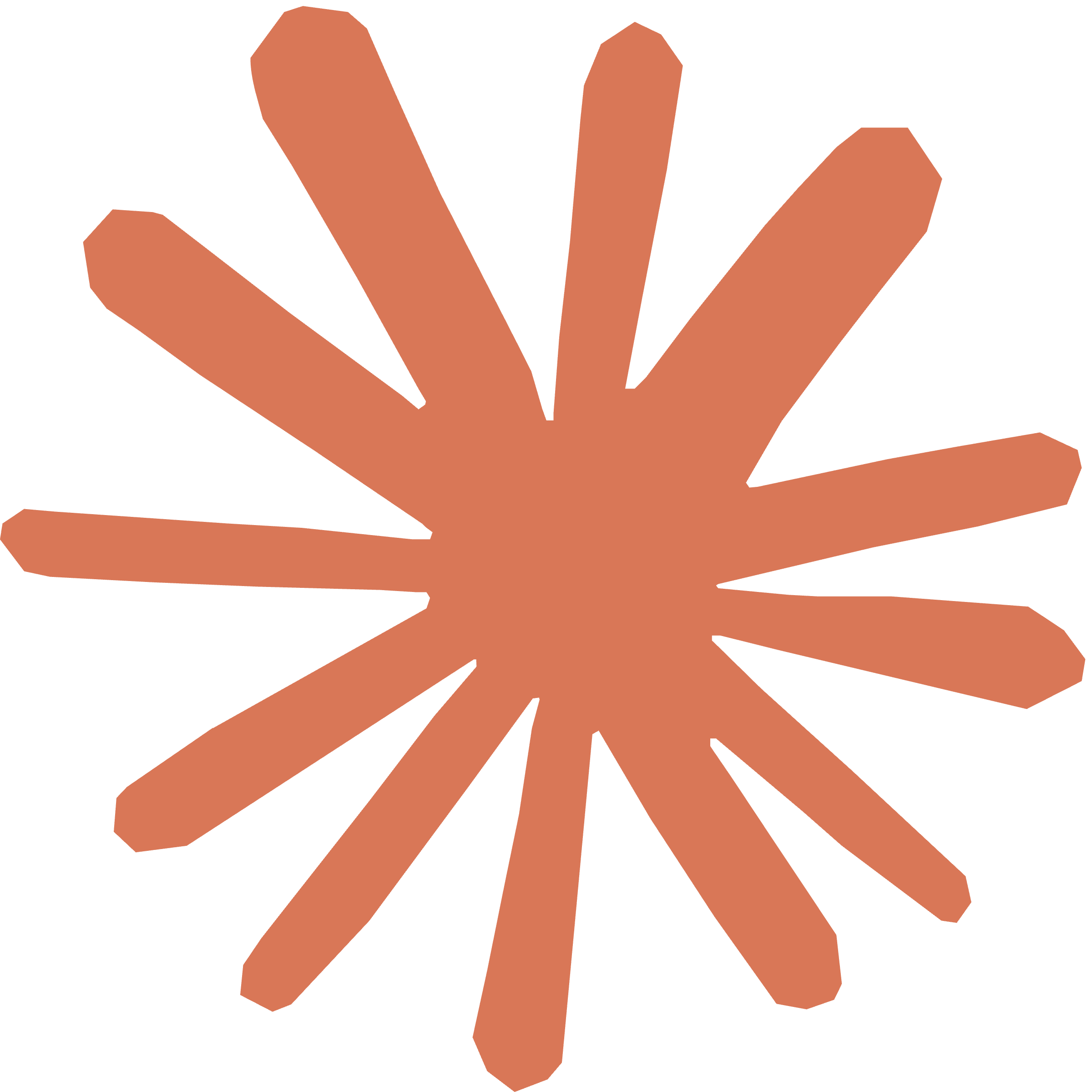} Claude S4.5 & Jan 2025 & 2516
  & 0.526 & 0.699 & 4.02 & 14.0
  & 2026-06 & +12.6 & +13.0 & 10.7 & 19.8 & 35.5 & 67.0 & 0.270 \\
  & & & {\scriptsize\textcolor{gray}{$+$0.031}} & \cellcolor{cellworse}{\scriptsize\textcolor{gray}{$-$0.063}} & \cellcolor{cellgood}{\scriptsize\textcolor{gray}{$+$0.07}} & & & & & & & & & \cellcolor{cellgood}{\scriptsize\textcolor{gray}{$+$0.069}} \\

\llmicon{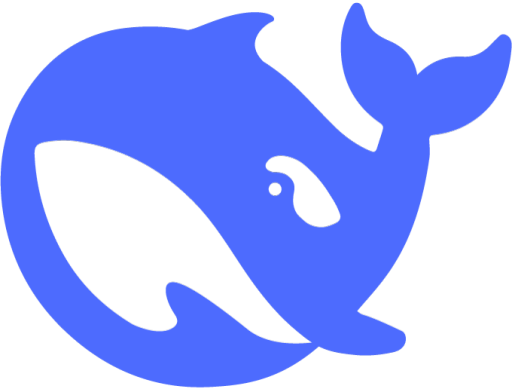} DeepSeek R1 & Jul 2024 & 1198
  & 0.480 & 0.589 & 4.18 & 20.1
  & 2026-08 & +15.7 & +10.0 & 16.7 & 32.4 & 57.6 & 83.7 & 0.328 \\
  & & & {\scriptsize\textcolor{gray}{$-$0.008}} & \cellcolor{cellworse}{\scriptsize\textcolor{gray}{$-$0.030}} & {\scriptsize\textcolor{gray}{$+$0.01}} & & & & & & & & & \cellcolor{cellgood}{\scriptsize\textcolor{gray}{$+$0.183}} \\

\llmicon{icons/openai_logo.png} GPT-4o & Oct 2023 & {\textemdash}
  & \textbf{0.519} & 0.530 & 3.26 & 3.9
  & 2028-02 & +35.9 & +26.0 & 6.8 & 13.2 & 25.5 & 46.8 & 0.178 \\
  & & & {\scriptsize\textcolor{gray}{---}} & {\scriptsize\textcolor{gray}{---}} & {\scriptsize\textcolor{gray}{---}} & & & & & & & & & {\scriptsize\textcolor{gray}{---}} \\

\llmicon{icons/openai_logo.png} GPT-OSS & Jun 2024 & 949
  & 0.526 & 0.466 & 3.86 & 11.4
  & 2027-06 & +25.6 & +12.0 & 17.7 & 30.8 & 50.5 & 63.7 & 0.336 \\
  & & & {\scriptsize\textcolor{gray}{$+$0.050}} & \cellcolor{cellworse}{\scriptsize\textcolor{gray}{$-$0.031}} & {\scriptsize\textcolor{gray}{$-$0.02}} & & & & & & & & & \cellcolor{cellgood}{\scriptsize\textcolor{gray}{$+$0.210}} \\

\llmicon{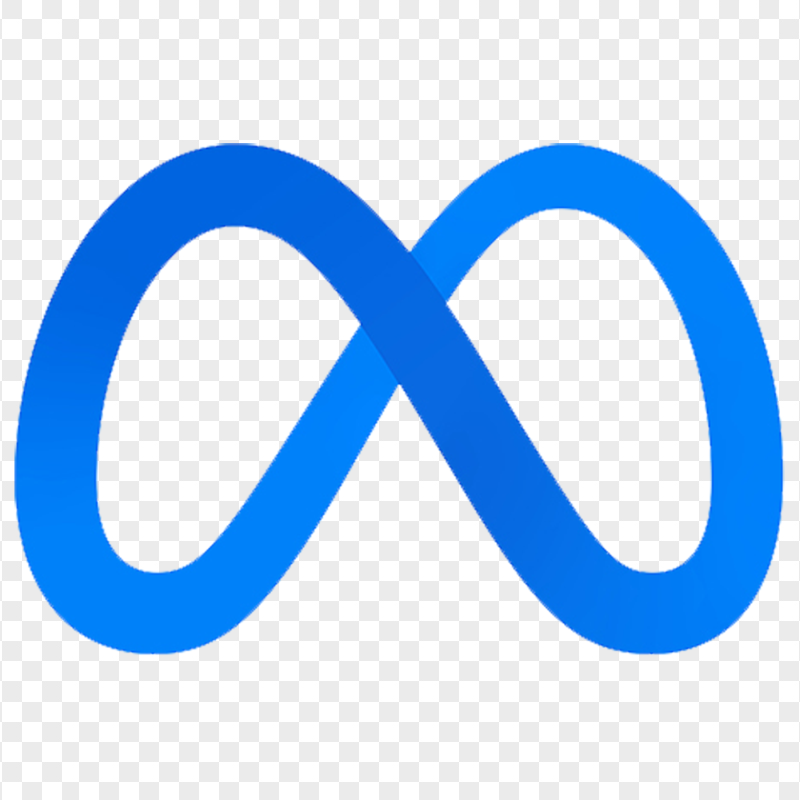} LLaMA 3.3 & Dec 2023 & {\textemdash}
  & 0.453 & 0.434 & 3.49 & 2.5
  & 2025-07 & \textbf{+4.9} & \textbf{+4.0} & \textbf{25.2} & \textbf{45.7} & \textbf{74.6} & \textbf{93.7} & \textbf{0.500} \\
  & & & {\scriptsize\textcolor{gray}{---}} & {\scriptsize\textcolor{gray}{---}} & {\scriptsize\textcolor{gray}{---}} & & & & & & & & & {\scriptsize\textcolor{gray}{---}} \\

\bottomrule
\end{tabular}
\end{adjustbox}
\end{table*}
\subsection*{Overall forecasting performance on \CUSP}
\label{sec:overall_model_performance}
We first evaluate frontier AI models on scientific advances occurring beyond each model's knowledge cutoff (\hyperref[tab:main_results]{Tab.~\ref*{tab:main_results}}). Across all evaluated models, we observe a striking asymmetry in event-level scientific forecasting. Models often identify plausible mechanisms underlying future scientific advances, yet remain substantially less effective at anticipating whether advances will occur, how they will ultimately be realized, and when they will become publicly observable.

This asymmetry is consistently reflected across the four forecasting dimensions. 
Models achieve their strongest performance on mechanistic forecasting (MCQ), with GPT-5.4 reaching 79.2\% accuracy and all evaluated models substantially outperforming the random-guess baseline (25\%). In contrast, performance is considerably lower on open-ended solution design (FRQ). 
Even GPT-5.4, the strongest model on mechanistic forecasting, achieves only 5.04 out of 10 on the FRQ evaluation.

Performance deteriorates further on the feasibility assessment and temporal prediction. On Binary prediction, all evaluated models perform close to chance (50\%). Despite leading mechanistic forecasting, GPT-5.4 achieves only 49.1\% accuracy, while all frontier models range from 43.5\% to 52.6\%, even underperforming the trivial always-No baseline (57.03\%). Temporal prediction exhibits similar limitations. Although LLaMA~3.3 achieves the highest date score (0.500), all models struggle to accurately anticipate when scientific advances become publicly observable.

Taken together, these results reveal a consistent asymmetry in how frontier AI models anticipate future scientific advances. Current models perform substantially more effectively on identifying plausible scientific pathways than forecasting whether advances will occur, how they will ultimately be realized, and when they will be observed. This suggests that event-level scientific forecasting requires predictive capabilities that are not captured by existing scientific reasoning evaluations.

\subsection*{Forecasting realized approaches is harder than generating plausible ones}
The large performance gap between MCQ and FRQ tasks suggests that forecasting scientific approaches underlying scientific events remains substantially more difficult when models must generate solution strategies without candidate guidance.

To better understand this gap, we analyze the sub-dimensions of FRQ evaluation (\hyperref[fig:radar_areas]{Fig.~\ref*{fig:radar_areas}b}). Across all evaluated models, alignment is consistently the lowest-scoring dimension. GPT-5.4 achieves the highest alignment score among all models, yet still reaches only 3.3 out of 10. In contrast, feasibility scores are substantially higher, ranging from 4.8 to 6.0 across models. This alignment--feasibility gap suggests that forecasting failures do not primarily arise from an inability to construct scientifically viable solutions. Instead, the generated approaches remain only weakly aligned with the methods underlying realized scientific advances.

Recent improvements in frontier models further reinforce this pattern (\hyperref[fig:frq_area]{Fig.~\ref*{fig:frq_area}b}). From GPT-4o to GPT-5.4, specificity increases from 2.9 to 6.2 and novelty from 2.4 to 4.9, indicating substantial gains in producing more detailed and creative scientific solutions. By comparison, alignment improves only modestly, from 2.2 to 3.3, while feasibility remains relatively stable (4.8--6.0). These trends suggest that recent advances primarily improve the ability to generate richer and more scientifically plausible solutions, rather than to anticipate the approaches that are ultimately realized. Detailed FRQ sub-dimension results are provided in \hyperref[fig:frq_radar]{Fig.~\ref*{fig:frq_radar}}.

Taken together, these findings indicate that current frontier AI models increasingly excel at generating plausible scientific pathways, yet remain substantially less effective at forecasting which pathways will ultimately lead to realized scientific advances.

\begin{figure*}[!htbp]
    \centering
    \includegraphics[
        width=\textwidth,
        height=\textheight,
        keepaspectratio
    ]{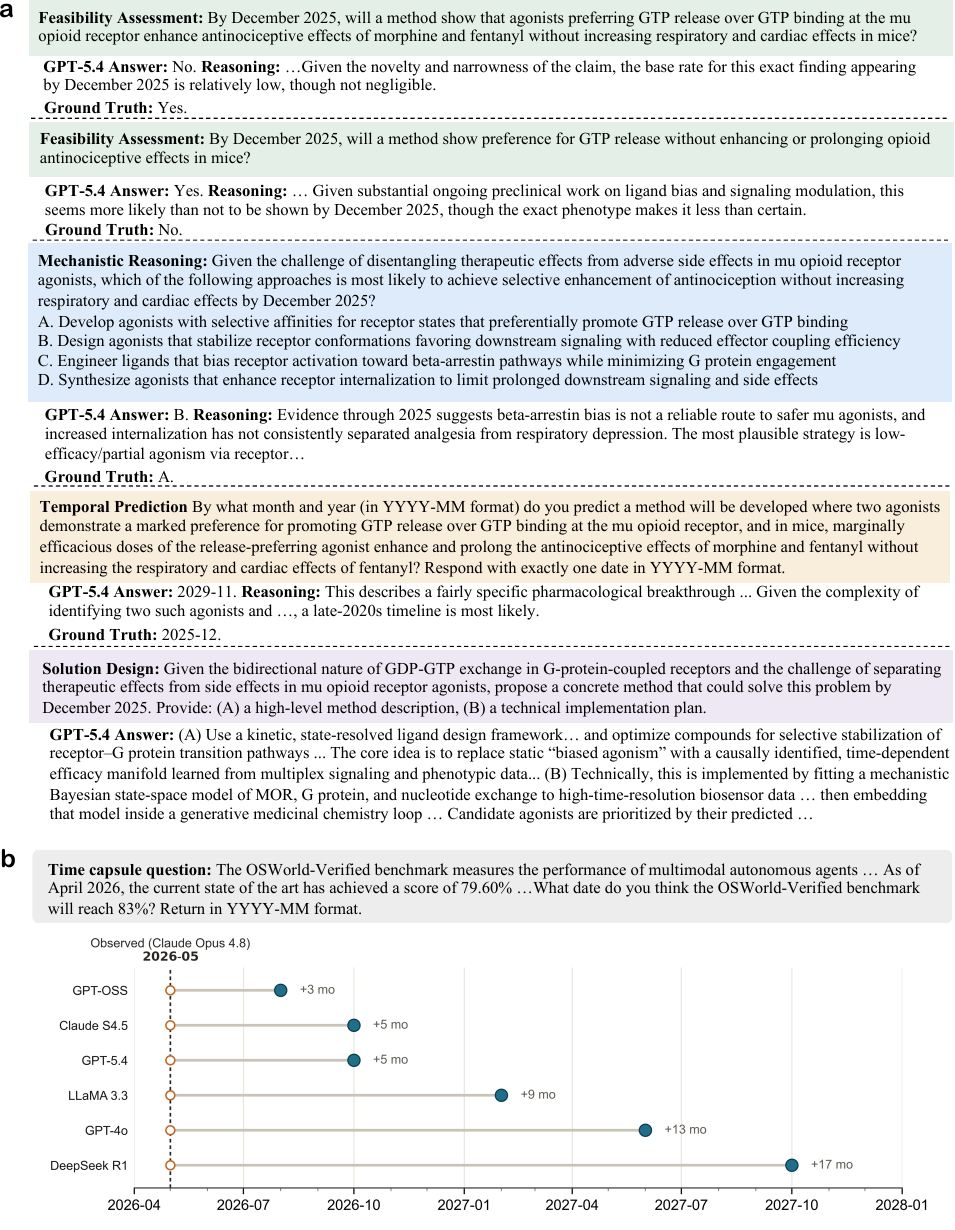}
\caption{\textbf{a.} Examples of model responses to four types of tasks in \CUSP. \textbf{b.} Example from \CUSP\  time capsule. The OSWorld-Verified score surpasses 83\% in 2026-05 (Claude Opus 4.8, 83.4\%), all models predicted a later date.}
    \label{fig:examples}
\end{figure*}

\begin{figure}[!htb]
    \centering
    \includegraphics[
        width=\linewidth,
    ]{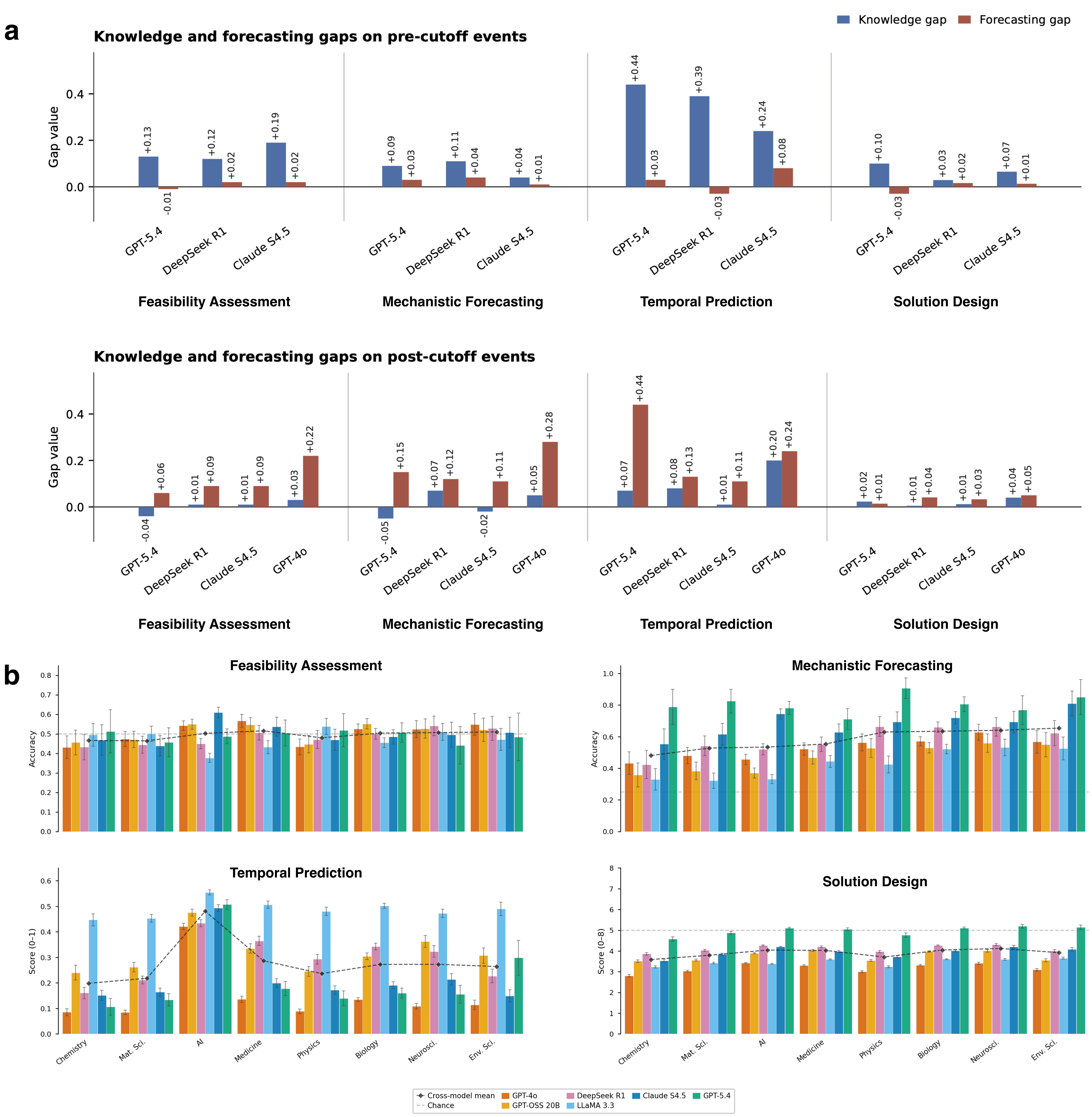}
    \caption{\textbf{Knowledge and forecasting gaps diverge across the knowledge cutoff.} \textbf{a}. Knowledge gaps (blue) and forecasting gaps (red) across four task types on pre-cutoff (top) and post-cutoff (bottom) events. Knowledge gaps dominate before the cutoff; after it the pattern inverts, with forecasting gaps largest for Temporal Prediction. \textbf{b}. Per-domain performance of six models across eight scientific fields for each task; diamonds, cross-model mean; dashed lines, chance. Error bars, s.e.m.}
    \label{fig:gaps}
\end{figure}
  
\subsection*{Knowledge alone is insufficient for forecasting scientific events}
The limited forecasting performance may arise for two distinct reasons. Models may lack access to relevant pre-event scientific knowledge, or they may possess such knowledge but struggle to use it to reliably forecast future scientific events. 
Under CUSP's event-level formulation, we find only modest performance differences between pre- and post-cutoff events (\hyperref[tab:main_results]{Tab.~\ref*{tab:main_results}}), suggesting that whether a scientific event occurs before or after a model's knowledge cutoff is insufficient to fully explain forecasting performance.

To separate these effects, we evaluate models under controlled information access on a random subset of \CUSP\ (500 scientific events). We compare three settings: the base model, web search restricted to pre-cutoff information (WS+Cutoff), and unrestricted web search (WS). This design allows us to decompose performance into a knowledge gap ($\Delta_{\text{know}}$), representing gains from additional pre-cutoff information, and a forecasting gap ($\Delta_{\text{fore}}$), representing the remaining difference between model performance under pre-cutoff information and full-information access.

For pre-cutoff events, knowledge gaps dominate forecasting gaps across most tasks (\hyperref[fig:gaps]{Fig.~\ref*{fig:gaps}a}). Providing additional pre-cutoff information consistently improves performance, indicating that models do not fully access or utilize relevant scientific knowledge in the base setting. The largest improvements are observed for temporal prediction and feasibility assessment, suggesting that better access to historical scientific knowledge substantially enhances performance when the underlying scientific events are already within models' knowledge horizon.

In contrast, the pattern reverses for post-cutoff events. Although additional pre-cutoff information continues to improve performance, forecasting gaps frequently exceed knowledge gaps by a substantial margin (\hyperref[fig:gaps]{Fig.~\ref*{fig:gaps}a}). This effect is particularly pronounced for temporal prediction and mechanistic forecasting. For example, GPT-5.4 exhibits a forecasting gap more than six times larger than its knowledge gap on temporal prediction ($\Delta_{\text{fore}}=0.44$ versus $\Delta_{\text{know}}=0.07$), with similar trends observed for Claude Sonnet and DeepSeek R1. These results indicate that even when models are provided with relevant historical scientific knowledge, substantial forecasting limitations remain.

Taken together, these findings indicate that limited forecasting performance cannot be explained solely by incomplete access to scientific knowledge. While additional historical information consistently improves forecasting performance, substantial forecasting gaps persist even when relevant pre-cutoff knowledge is available. This suggests that reliable event-level scientific forecasting requires capabilities beyond simply accessing and reasoning over existing scientific knowledge.

\subsection*{Human performance under identical information conditions}
A remaining alternative explanation is that forecasting future scientific advances is inherently difficult given the available evidence, making the observed model limitations largely unavoidable. 
To examine this possibility, we compare human participants and a representative frontier model under identical information conditions.

We conduct a human study using the same evidence packets provided to GPT-4o under constrained web search.
To ensure comparable expertise across participants, the study is restricted to AI-related scientific advances.
Since participants are recruited after many scientific events have already occurred, they are instructed to skip any question for which they believe that they have recognized the underlying paper, benchmark result, or scientific event. 
A total of 100 events are used for human study. We evaluated Binary, MCQ, and Date prediction tasks using the same scoring procedures as the AI evaluation.

Human participants achieve consistently better performance on feasibility assessment, mechanistic forecasting, and predicting the timing of scientific advances than GPT-4o.
Humans achieved 73.0\% accuracy on binary questions compared with 52.5\% for GPT-4o, and 67.0\% accuracy on MCQs compared with 40.0\%. 
Humans also obtained a higher Date score than GPT-4o (0.65 versus 0.54). 

Together with the controlled knowledge-access experiments, these results indicate that forecasting limitations cannot be attributed solely to task difficulty under controlled information access. 
Under identical information conditions, humans remain substantially more effective at assessing whether scientific advances will occur, identifying the approaches underlying those advances, and anticipating when they will become publicly observable than GPT-4o.

\subsection*{Forecasting errors are systematic rather than random}
\label{sec:bias}

Forecasting failures are not merely a consequence of limited predictive accuracy. 
Instead, frontier AI models exhibit consistent and structured error patterns across forecasting tasks.
These biases are not random errors, but consistent patterns that reveal how current AI models represent feasibility, uncertainty, and timing when forecasting future scientific events.

\begin{figure*}[!htbp]
    \centering
    \includegraphics[
        width=\textwidth,
        height=0.9\textheight,
        keepaspectratio
    ]{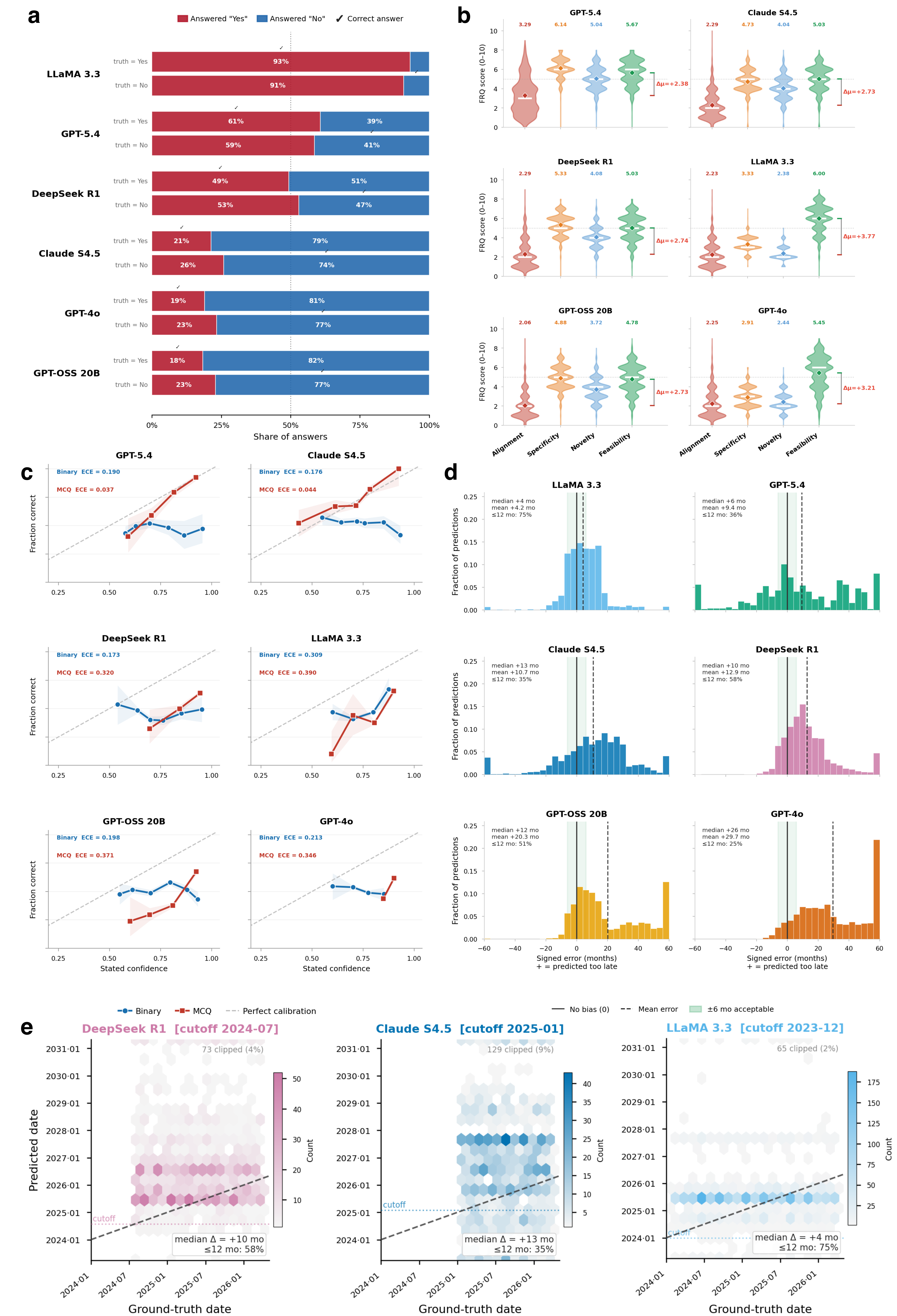}
    \caption{\textbf{a.} Distribution of feasibility assessment responses across models. \textbf{b.} Distribution of FRQ scores across evaluation dimensions and models. \textbf{c.} Calibration curves for binary and multiple-choice questions. \textbf{d.} Distributions of temporal prediction errors. \textbf{e.} Anchor-point effect on temporal prediction.}
    \label{fig:radar_areas}
\end{figure*}

Models exhibit strong response priors when assessing the feasibility of future scientific advances (\hyperref[fig:radar_areas]{Fig.~\ref*{fig:radar_areas}a}). 
For instance, LLaMA~3.3 and GPT-5.4 favor affirmative responses while most other models display a persistent tendency toward negative predictions. 
DeepSeek R1 is the only model that remains comparatively balanced between positive and negative predictions.
These response tendencies contribute to the near-chance performance observed on feasibility assessment tasks and suggest that model predictions are influenced by systematic response priors.

Models also become substantially overconfident when forecasting future scientific advances (\hyperref[fig:radar_areas]{Fig.~\ref*{fig:radar_areas}c}). 
While GPT-5.4 (ECE: 0.033) and Claude S4.5 (ECE: 0.022) remain relatively well calibrated on mechanistic forecasting (MCQ), all other models exhibit high calibration errors (ECE$\geq$0.2).
All models become overconfident and exhibit high calibration errors (0.16$\leq$ECE$\leq$0.31) in feasibility assessment.
These results indicate confidence estimates become unreliable when models forecast future scientific advances.

Temporal prediction exhibits delayed realization (\hyperref[fig:radar_areas]{Fig.~\ref*{fig:radar_areas}d}). 
Across all models, predicted publicly observable dates are systematically later than the true dates, with median prediction errors ranging from four months (LLaMA~3.3) to more than two years (GPT-4o).
An example of time capsule question also provide additional evidence to the delayed prediction (\hyperref[fig:examples]{Fig.~\ref*{fig:examples}b}).
Although all models predict that the scientific event will eventually be achieved, they consistently overestimate the time required.

Moreover, the predicted dates are not uniformly distributed over the time span but exhibit the anchor-point effect, with most predicted dates falling on specific dates (\hyperref[fig:radar_areas]{Fig.~\ref*{fig:radar_areas}e}).
To contextualize this behavior, we compare model performance against a trivial baseline that always predicts the mean publication date of the post-cutoff events.
This simple baseline achieves substantially higher date scores than all evaluated models (GPT-5.4: 0.857, Claude S4.5: 0.733, DeepSeek R1: 0.650, GPT-OSS: 0.637, GPT-4o: 0.578, LLaMA 3.3: 0.578), suggesting that temporal predictions are influenced by the anchor-point effect.
Full visualization of anchor-point effect is shown in \hyperref[fig:date_clusters]{Fig.~\ref*{fig:date_clusters}}.

Taken together, these findings show that forecasting failures arise from systematic prediction biases rather than random error. Current frontier AI systems exhibit persistent response priors, overconfidence, and temporal conservatism, indicating that limitations in anticipating future scientific advances reflect characteristic biases in how these systems represent uncertainty in forecasting.


\subsection*{Capability-dependent variation across scientific domains}
The asymmetry in event-level scientific forecasting persists across scientific domains, but the extent of domain dependence varies substantially across forecasting capabilities (\hyperref[fig:gaps]{Fig.~\ref*{fig:gaps}b}). 
While some dimensions of forecasting benefit from domain-specific knowledge, others remain consistently challenging regardless of scientific domain.

Domain-dependent variation emerges primarily in mechanistic forecasting and temporal prediction. 
For mechanistic forecasting, models achieve substantially higher performance in environmental science (65.4\%), neuroscience (64.0\%), and biology (63.5\%) than in chemistry (48.1\%) and AI (53.4\%), suggesting that models are better at recognizing technical approaches underlying scientific advances in some domains than others. 
Temporal prediction exhibits an even stronger pattern. 
Across models, model performance on predicting the timing of AI advances (0.481) is substantially higher than in biology (0.272), chemistry (0.198), physics (0.236), and medicine (0.286), with scores nearly doubling those observed in most other domains.

In contrast, feasibility assessment and solution design remain consistently challenging. 
Feasibility assessment stays close to chance across all domains, with accuracy ranging from 46.4\% (material science) to 51.6\% (medicine), indicating a persistent inability to assess whether scientific advances will be realized. 
Similarly, open-ended solution design remains challenging across all domains, with even the highest-performing domain (biology, 4.05) achieving less than half of the maximum attainable score. 
Notably, these limitations persist even in domains where mechanistic forecasting and temporal prediction perform relatively well.

These findings further reinforce the capability-dependent asymmetry observed throughout this study.
While recognizing the mechanisms and estimating the timing of future scientific advances depend substantially on the scientific domain, assessing whether advances will occur and generating the approaches that ultimately lead to realized scientific advances remain consistently challenging across domains.

\section*{Discussion}

As AI systems become increasingly integrated into scientific discovery, understanding their ability to support forward-looking scientific judgment is becoming increasingly important. Using \CUSP, a temporally grounded evaluation suite spanning 4,760 scientific events across eight scientific disciplines, we find that current frontier AI systems exhibit substantial limitations in event-level scientific forecasting. While models often identify plausible mechanisms underlying future scientific advances and generate technically detailed scientific solutions, they remain substantially less reliable at forecasting whether advances will occur, when they will become publicly observable, and how they will ultimately be realized. These limitations persist across scientific domains, cannot be explained solely by access to scientific knowledge, and exhibit consistent systematic error patterns.

Our results suggest that scientific reasoning and scientific forecasting should not be treated as interchangeable evaluations. Existing benchmarks largely measure how effectively models reason about scientific knowledge through question answering, knowledge retrieval, and scientific problem solving. By contrast, forecasting future scientific advances requires models to form expectations about scientific outcomes that have not yet occurred. Although frontier models increasingly generate scientifically coherent and technically plausible solution strategies, these strategies frequently diverge from the approaches underlying realized scientific advances. Providing additional pre-cutoff scientific knowledge consistently improves forecasting performance, but substantial forecasting limitations remain even when relevant historical information is available. Together, these findings indicate that access to scientific knowledge and scientific reasoning are important but insufficient for forecasting future scientific advances.


The practical implications arise because their forecasting failures exhibit systematic patterns instead of frontier models forecasting imperfectly. Overconfidence may cause AI systems to overestimate confidence in uncertain scientific opportunities, potentially mis-ranking competing research directions. Persistent response priors may bias go/no-go decisions for speculative research ideas. Temporal conservatism may delay investment in rapidly developing scientific areas by systematically predicting advances later than they ultimately become publicly observable. Weak alignment between generated solution strategies and realized scientific advances may encourage researchers to pursue scientifically plausible yet ultimately unrealized approaches. These systematic failure modes therefore have implications beyond forecasting accuracy itself, influencing how AI systems may shape research prioritization, opportunity evaluation, and scientific planning.

Our findings should nevertheless be interpreted within the scope of the forecasting setting studied here. \CUSP\ evaluates event-level scientific forecasting rather than broader scientific foresight. Whether similar limitations extend to forecasting the evolution of entire scientific fields, long-term research trajectories, or scientific paradigms remains an open question. Furthermore, although our study spans multiple scientific disciplines and forecasting dimensions, the evaluation suite necessarily reflects its event selection process. Controlled information-access experiments separate knowledge-access limitations from forecasting limitations but do not fully characterize the internal mechanisms by which frontier models form future-oriented expectations. In addition, while our automated evaluation framework demonstrates moderate agreement with human evaluation, free-response assessment remains inherently more subjective than deterministic forecasting tasks. Finally, our controlled human study is limited to AI-related scientific advances and cannot completely eliminate residual familiarity with previously published work.

An important direction for future research is the development of AI systems capable of forming better-calibrated expectations about future scientific advances. Beyond improving forecasting accuracy, future work should investigate the relationship between event-level scientific forecasting and broader forms of scientific foresight, as well as the mechanisms underlying systematic forecasting failures. More broadly, scientific forecasting may emerge as a complementary evaluation paradigm alongside scientific reasoning, helping characterize what AI systems know about science and how well they anticipate its future development. Understanding when AI systems can reliably anticipate future scientific advances and when they cannot may become increasingly important as these systems play larger roles in scientific discovery and decision-making.

\section*{Methods}

\section*{The \CUSP\ suite}

We develop \CUSP\ using a temporally stratified corpus of scientific events, spanning January 2024 to March 2026, to evaluate the capabilities of forecasting scientific progress in current AI systems under controlled temporal knowledge constraints. 
\CUSP\ is designed to rigorously evaluate predictive performance and calibrated expectation on scientific development across a broad spectrum of scientific disciplines. We construct the dataset using domain-specific inclusion criteria to account for the diverse publication dynamics and evidentiary standards inherent to these fields, ensuring that all incorporated events represent verifiable and definitively resolved advances. Full inclusion criteria are provided in Appendix~\ref{sec:criteria}.


We source natural science events from \textit{Nature}, \textit{Science}, and \textit{Cell}, restricting this subset to high-impact peer-reviewed publications with clearly measurable outcomes. To prevent temporal leakage, we query Crossref, Semantic Scholar, OpenAlex, Europe PMC, arXiv, and bioRxiv/medRxiv, and use the earliest observed date associated with each manuscript DOI as the relevant knowledge boundary. For artificial intelligence, we include high-visibility papers from community-driven repositories, including weekly top-paper lists and the Hugging Face Top Papers hub, together with time-resolved leaderboard records from widely used benchmarks such as GPQA Diamond \cite{rein2024gpqa}, MMLU-Pro \cite{wang2024mmlu}, and Humanity's Last Exam \cite{phan2025humanity}. Details of source selection, ranking criteria, and leaderboard construction are provided in Appendix~\ref{sec:data_acquisition}.


\subsection*{Question Types and Synthesis}
To operationalize scientific forecasting as a measurable capability, we decompose it into four core dimensions and design task formats that probe each aspect. For each accepted milestone, we construct four complementary evaluation tasks.

Binary prediction (including its perturbed variant) assesses feasibility and calibration by testing whether models can determine if a concrete scientific claim will be achieved and distinguish realized advances from plausible but unrealized alternatives. Multiple-choice questions probe mechanistic forecasting by asking models to identify the technical approach that later enabled the discovery from competing candidates. Free-response questions evaluate generative solution design by requiring models to propose a concrete solution strategy based on the scientific problem context. Finally, date prediction tasks assess temporal reasoning by asking models to forecast when a scientific advance will be first publicly observable.

To construct these tasks, we first decompose each abstract into three structured components: a problem statement, a technical approach, and a results summary. We explicitly remove post-cutoff identifiers and narratives, such as newly introduced acronyms or methodology names, to prevent information leakage. Details of this decomposition procedure are provided in Appendix~\ref{sec:task_synthesis}. The full task taxonomy is summarized in 
\hyperref[tab:task_taxonomy]{Tab.~\ref*{tab:task_taxonomy}}, and representative examples are provided in Appendix~\ref{sec:task_examples}. \hyperref[fig:cusp_pipeline]{Fig.~\ref*{fig:cusp_pipeline}} illustrates the end-to-end construction pipeline.

To ensure the validity of forecasting tasks, we apply a two-stage validation procedure combining an independent LLM judge with human expert review. The validation process verifies that each task is faithful to the source abstract, objectively verifiable, and free from unsupported perturbations or trivial distractors. This step ensures that all tasks correspond to well-defined and reliable targets for evaluating predictive capabilities. Full validation criteria, prompts, and human-LLM agreement analyses are provided in Appendix~\ref{sec:benchmark_validation}.


\subsection*{\CUSP\ Time Capsule Construction}
We further introduce \CUSP\ Time Capsule, a set of prospective forecasting tasks whose outcomes are not known at evaluation time but will become verifiable in the future.  
Each question in  \CUSP\ Time Capsule is constructed so that the outcome corresponds to a real-world, measurable event and can be resolved by an authoritative source without ambiguity. 
Because ground truth is not yet available, the Time Capsule is not used for accuracy evaluation, but for analyzing prediction consistency, confidence calibration, and agreement across models.

We design questions across domains, including scientific benchmarks, institutional recognitions, and future technological events, curated in collaboration with human experts to ensure their relevance and verifiability (see Appendix~\ref{app:time_capsule}). 
In addition, we include AI capability forecasting by conditioning on current state-of-the-art results. 
Together, the Time Capsule extends \CUSP\ from retrospective evaluation to prospective analysis of how models form predictions about future scientific progress.

\subsection*{Key Statistics of \CUSP}

\CUSP\ is an event-based, multi-disciplinary, and temporally grounded evaluation suite designed to evaluate AI systems’ ability to anticipate scientific progress. It consists of 4,760 scientific events, from which we construct 17,429 structured forecasting tasks spanning multiple evaluation formats. \CUSP\ spans eight top-level scientific domains and 4,245 distinct subcategories, reflecting highly specialized research topics across disciplines. The dataset is dominated by biology (1,234 papers) and artificial intelligence (1,141 papers), followed by medicine (746), neuroscience (403), materials science (375), physics (359), environmental science (235), chemistry (203), and other domains (64). See
\hyperref[fig:single_column]{Fig.~\ref*{fig:single_column}}
for further details on temporal, source, and subsection details of our evaluation suite. See Appendix~\ref{sec:statistics} for key statistics and distributional properties of \CUSP.

\subsection*{Comparison to Related Evaluation}
Two lines of prior work approach related problems, but neither evaluates scientific forecasting under temporal knowledge constraints. The first targets general-world forecasting over news, markets, or geopolitical events and does not enforce knowledge cutoffs tied to scientific discovery \cite{karger2025forecastbench, zeng2025futurex, yuan2025introducing, tao2025prophet, goel2026futuresimreplayingworldevents}, leaving its targets disconnected from verifiable scientific events. 
The second measures the retrospective reasoning via scientific reasoning \cite{lu2022learn, rein2024gpqa} and problem-solving tasks \cite{bragg2025astabench, phan2025humanity} against resolved ground truth.
Although some prior work \cite{ajith2026prescience, liu2025researchbench} explores prospective reasoning and future prediction, existing evaluations lack a temporally stratified, event-level framework for disentangling access to scientific knowledge from the ability to forecast future scientific progress.
\CUSP\ grounds tasks in temporally referenced scientific events, enabling controlled disentanglement between access to scientific knowledge and the ability to forecast future scientific progress.
Moreover, \CUSP\ remains substantially unsaturated across all evaluated frontier systems 
(\hyperref[fig:saturation]{Fig.~\ref*{fig:saturation}}), consistent with our finding that forecasting scientific progress is harder than retrospective reasoning. 
\hyperref[tab:benchmark_comparison]{Tab.~\ref*{tab:benchmark_comparison}}
summarizes the comparison, and Appendix~\ref{sec:comparison_to_benchmarks} provides details.

 \newcommand{\cmark}{\textcolor{green!60!black}{\ding{51}}}
\newcommand{\xmark}{\textcolor{red!70!black}{\ding{55}}}

\definecolor{cusppurple}{RGB}{88, 66, 155}

\begin{table}[t]
\centering
\small
\setlength{\tabcolsep}{3pt}
\renewcommand{\arraystretch}{1.1}

\resizebox{\textwidth}{!}{%
\begin{tabular}{lccccccc}
\specialrule{0.5pt}{0pt}{0pt}

\rowcolor{cuspborder}
\textcolor{white}{\textbf{Benchmark}} &
\textcolor{white}{\textbf{Scientific}} &
\textcolor{white}{\textbf{Forecasting}} &
\textcolor{white}{\textbf{Time}} &
\textcolor{white}{\textbf{Cutoff}} &
\textcolor{white}{\textbf{Multi}} &
\textcolor{white}{\textbf{Scale}} &
\textcolor{white}{\textbf{Key}} \\

\specialrule{0.5pt}{0pt}{0pt}

ForecastBench \cite{karger2025forecastbench} & \xmark & \cmark & \cmark & \xmark & \xmark & $\sim$1K & Events \\
FutureX \cite{zeng2025futurex}      & \xmark & \cmark & \cmark & \xmark & \cmark & --- & Agents \\
FOReCAst \cite{yuan2025introducing}     & \xmark & \cmark & \cmark & \xmark & \cmark & --- & Calib \\
PROPHET \cite{tao2025prophet}    & \xmark & \cmark & \cmark & \xmark & \xmark & --- & Retr. \\

\midrule

Humanity’s Last Exam \cite{phan2025humanity}& \cmark & \xmark & \xmark & \xmark & \xmark & $\sim$2.5K & QA \\
AstaBench \cite{bragg2025astabench}         & \cmark & \xmark & \xmark & \xmark & \cmark & $\sim$2.4K & Agents \\
PreScience \cite{ajith2026prescience}        & \cmark & \cmark & \xmark & \xmark & \cmark & 98K & Papers \\
ResearchBench \cite{liu2025researchbench}       & \cmark & \xmark & \xmark & \xmark & \cmark & 1.3K & Decomp \\
ScienceQA \cite{lu2022learn}           & \cmark & \xmark & \xmark & \xmark & \cmark & 21K & Reason \\
Matter-of-Fact \cite{jansen2025matter}           & \cmark & \xmark & \cmark & \cmark & \xmark & 8.4K & Feasibility \\
\midrule

\textbf{\CUSP\ (ours)} & \cmark & \cmark & \cmark & \cmark & \cmark 
& \textbf{4.7K / 17K$^\dagger$} & \textbf{Forecasting} \\

\bottomrule
\end{tabular}%
}

\caption{Comparison of \CUSP\ with forecasting and scientific benchmarks. \CUSP\ uniquely combines scientific grounding, temporal prediction, cutoff conditioning, and multi-task evaluation. $^\dagger$Scale grows periodically over time as new scientific events are continuously incorporated.}
\label{tab:benchmark_comparison}
\end{table}

\section{Model Evaluation}
\label{sec:model_evaluation}

We evaluate model predictions on \CUSP\ using a two-track evaluation framework that combines deterministic outcome scoring with rubric-based scientific reasoning evaluation
(\hyperref[fig:eval_pipeline]{Fig.~\ref*{fig:eval_pipeline}b}). This design is motivated by prior work showing that language models can produce correct final answers while relying on flawed, unfaithful, or post-hoc reasoning processes \cite{lu2025solving,lightman2023let,matton2025walk}. \CUSP\ therefore evaluates not only whether a model can correctly forecast future scientific developments, but also whether its generated scientific proposals remain temporally consistent and scientifically plausible under a strict knowledge cutoff.

Each scientific advance may contain a subset of four task types, including binary classification, multiple-choice questions (MCQ), free-response questions (FRQ), and date prediction. For binary classification, we include both the original statement and a negation-perturbed variant and report a merged score averaged across the two in order to mitigate directional response bias. The evaluation framework automatically detects which tasks are present for each scientific event and scores only the corresponding outputs.

\paragraph{Track I. Deterministic outcome evaluation.}
The first evaluation track measures forecasting accuracy using deterministic grading procedures tailored to each task type. Binary tasks are evaluated using exact yes or no agreement with the ground-truth label. MCQ tasks are scored through deterministic answer extraction with semantic matching fallbacks when explicit option selection cannot be reliably parsed. Date prediction is evaluated using an exponential-decay distance metric, $\exp(-0.1d)$, where $d$ denotes the absolute month difference between the predicted and ground-truth publication dates. This formulation assigns partial credit to temporally proximate forecasts while smoothly penalizing large temporal errors.

\paragraph{Track II. Free-response scientific reasoning evaluation.}
Free-response questions (FRQs) evaluate whether models can generate scientifically plausible solution strategies for open research problems under a strict temporal cutoff. Because these tasks admit multiple potentially valid solutions, exact-match evaluation is inappropriate. Instead, we employ a rubric-based LLM-as-a-judge protocol using GPT-5.4-mini augmented with agentic web search.

FRQ evaluation proceeds in two stages. First, the judge performs explicit leakage detection using web search to determine whether a generated response contains information, terminology, methods, datasets, or discoveries that were unavailable prior to the knowledge cutoff date. Responses identified as containing post-cutoff information are treated as contaminated and do not receive valid forecasting credit. This separation between leakage detection and proposal-quality evaluation prevents memorization from being conflated with scientific reasoning ability.

Second, non-contaminated responses are evaluated along four complementary dimensions designed to capture scientific proposal quality. Alignment measures whether the proposal directly addresses the stated scientific problem. Specificity evaluates the presence of concrete technical mechanisms or implementation details. Novelty measures whether the response introduces non-trivial methodological ideas relative to pre-cutoff scientific knowledge. Feasibility assesses whether the proposed approach is scientifically and technically plausible under the stated constraints. Each dimension is scored on a rubric-anchored 0--10 scale using explicit evaluation criteria and reference anchors provided in Appendix~\ref{sec:llm_judge_prompt}. The final FRQ score is computed as the normalized mean across all four dimensions. To ensure the reliability of this automated protocol, we conducted a human expert correlation study; detailed methodology and results comparing human versus LLM-as-a-judge scoring are provided in Appendix~\ref{sec:judge_human_eval}. 

\clearpage
\renewcommand{\figurename}{Supplementary Figure}  
\setcounter{figure}{0}   

\section*{Acknowledgements}

Sean Wu is supported by the Rhodes Scholarship. Junchi Yu and Philip Torr are funded by the UKRI grant: Turing AI Fellowship EP/W002981/1 and the Schmidt Science Foundation. Philip Torr is a Schmidt Science AI 2050 Senior Fellow. David Clifton is funded by an NIHR Research Professorship (NIHR302440), a Royal Academy of Engineering Research Chair, and the InnoHK Hong Kong Centre for Cerebro-Cardiovascular Engineering, and was supported by the National Institute for Health Research Oxford Biomedical Research Centre and the Pandemic Sciences Institute at the University of Oxford. This work is partially supported by the Hoffman-Yee Research Grants program at Stanford HAI and the AI for Math Fund by Renaissance Philanthropy. We thank Tinglin Huang, Shenxu Chang, Aoxi Liu, Baicheng Chen, Xiaoyu Zhang, Guanzong Wu, Bo Zheng, Zhun Zhang, Yijie Sun, and Ruiyang Lu for their assistance with human evaluation of our study. We thank Fabien Scalzo for support with GPU compute, and Y Combinator for providing API compute credits.

\section*{Ethics declarations}
The authors declare no competing interests.

\section*{Author contributions statement}

Sean Wu and Pan Lu contributed equally. Junchi Yu conceived the study, with 
input from Philip Torr and James Zou. Sean Wu, Pan Lu, and Junchi Yu 
designed the evaluation suite. Sean Wu led data curation, task synthesis, and 
validation, and ran the experiments, with contributions from Pan Lu and 
Yupeng Chen. Peter Clark, Yutaro Yamada, and Jonathan Bragg provided 
valuable guidance on the analysis and evaluation design. David Clifton 
contributed to scientific discussion and manuscript revision. Sean Wu, Pan 
Lu, Philip Torr, James Zou, and Junchi Yu interpreted the results and 
drafted the manuscript. All authors revised and approved the final 
manuscript. Philip Torr, James Zou, and Junchi Yu jointly supervised the 
project.
\section*{Data Availability}
\CUSP\ is a curated evaluation suite derived from recent scientific literature and distributed under the MIT License, available at \href{https://huggingface.co/datasets/SeanWu25/CUSP}{https://huggingface.co/datasets/SeanWu25/CUSP}.
\section*{Code Availability}
Source code for \CUSP\ is available at \href{https://github.com/SeanWu25/cusp-scientific-foresight}{https://github.com/SeanWu25/cusp-scientific-foresight}.

\bibliography{references}

@article{krenn2023forecasting,
  title={Forecasting the future of artificial intelligence with machine learning-based link prediction in an exponentially growing knowledge network},
  author={Krenn, Mario and Buffoni, Lorenzo and Coutinho, Bruno and Eppel, Sagi and Foster, Jacob Gates and Gritsevskiy, Andrew and Lee, Harlin and Lu, Yichao and Moutinho, Jo{\~a}o P and Sanjabi, Nima and others},
  journal={Nature Machine Intelligence},
  volume={5},
  number={11},
  pages={1326--1335},
  year={2023},
  publisher={Nature Publishing Group UK London}
}

@article{marwitz2026predicting,
  title={Predicting new research directions in materials science using large language models and concept graphs},
  author={Marwitz, Thomas and Colsmann, Alexander and Breitung, Ben and Brabec, Christoph and Kirchlechner, Christoph and Blasco, Eva and Marques, Gabriel Cadilha and Hahn, Horst and Hirtz, Michael and Levkin, Pavel A and others},
  journal={Nature Machine Intelligence},
  pages={1--10},
  year={2026},
  publisher={Nature Publishing Group UK London}
}

@article{uzzi2013atypical,
  title={Atypical combinations and scientific impact},
  author={Uzzi, Brian and Mukherjee, Satyam and Stringer, Michael and Jones, Ben},
  journal={Science},
  volume={342},
  number={6157},
  pages={468--472},
  year={2013},
  publisher={American Association for the Advancement of Science}
}

@article{xiao2025densing,
  title={Densing law of llms},
  author={Xiao, Chaojun and Cai, Jie and Zhao, Weilin and Lin, Biyuan and Zeng, Guoyang and Zhou, Jie and Zheng, Zhi and Han, Xu and Liu, Zhiyuan and Sun, Maosong},
  journal={Nature Machine Intelligence},
  pages={1--11},
  year={2025},
  publisher={Nature Publishing Group UK London}
}

@article{messeri2024artificial,
  title={Artificial intelligence and illusions of understanding in scientific research},
  author={Messeri, Lisa and Crockett, Molly J},
  journal={Nature},
  volume={627},
  number={8002},
  pages={49--58},
  year={2024},
  publisher={Nature Publishing Group UK London}
}

@article{shapere1964structure,
  title={The structure of scientific revolutions},
  author={Shapere, Dudley},
  journal={The Philosophical Review},
  volume={73},
  number={3},
  pages={383--394},
  year={1964},
  publisher={JSTOR}
}

@article{park2023papers,
  title={Papers and patents are becoming less disruptive over time},
  author={Park, Michael and Leahey, Erin and Funk, Russell J},
  journal={Nature},
  volume={613},
  number={7942},
  pages={138--144},
  year={2023},
  publisher={Nature Publishing Group UK London}
}

@inproceedings{bragg2025astabench,
  title={Astabench: Rigorous benchmarking of ai agents with a scientific research suite},
  author={Bragg, Jonathan and D'Arcy, Mike and Balepur, Nishant and Bareket, Dan and Dalvi, Bhavana and Feldman, Sergey and Haddad, Dany and Hwang, Jena D and Jansen, Peter and Kishore, Varsha and others},
  booktitle={International conference on learning representations},
  year={2026}
}

@article{merchant2023scaling,
  title={Scaling deep learning for materials discovery},
  author={Merchant, Amil and Batzner, Simon and Schoenholz, Samuel S and Aykol, Muratahan and Cheon, Gowoon and Cubuk, Ekin Dogus},
  journal={Nature},
  volume={624},
  number={7990},
  pages={80--85},
  year={2023},
  publisher={Nature Publishing Group UK London}
}

@article{novikov2025alphaevolve,
  title={Alphaevolve: A coding agent for scientific and algorithmic discovery},
  author={Novikov, Alexander and V{\~u}, Ng{\^a}n and Eisenberger, Marvin and Dupont, Emilien and Huang, Po-Sen and Wagner, Adam Zsolt and Shirobokov, Sergey and Kozlovskii, Borislav and Ruiz, Francisco JR and Mehrabian, Abbas and others},
  journal={arXiv preprint arXiv:2506.13131},
  year={2025}
}

@article{liu2025researchbench,
  title={Researchbench: Benchmarking llms in scientific discovery via inspiration-based task decomposition},
  author={Liu, Yujie and Yang, Zonglin and Xie, Tong and Ni, Jinjie and Gao, Ben and Li, Yuqiang and Tang, Shixiang and Ouyang, Wanli and Cambria, Erik and Zhou, Dongzhan},
  journal={arXiv preprint arXiv:2503.21248},
  year={2025}
}

@inproceedings{wang2024mmlu,
  title={Mmlu-pro: A more robust and challenging multi-task language understanding benchmark},
  author={Wang, Yubo and Ma, Xueguang and Zhang, Ge and Ni, Yuansheng and Chandra, Abhranil and Guo, Shiguang and Ren, Weiming and Arulraj, Aaran and He, Xuan and Jiang, Ziyan and others},
  booktitle={Advances in neural information processing systems},
  year={2024}
}

@inproceedings{rein2024gpqa,
  title={Gpqa: A graduate-level google-proof q\&a benchmark},
  author={Rein, David and Hou, Betty Li and Stickland, Asa Cooper and Petty, Jackson and Pang, Richard Yuanzhe and Dirani, Julien and Michael, Julian and Bowman, Samuel R},
  booktitle={First conference on language modeling},
  year={2024}
}

@inproceedings{hendrycksmeasuring,
  title={Measuring Massive Multitask Language Understanding},
  author={Hendrycks, Dan and Burns, Collin and Basart, Steven and Zou, Andy and Mazeika, Mantas and Song, Dawn and Steinhardt, Jacob},
  booktitle={International conference on learning representations},
  year={2021}
}

@article{zeng2025futurex,
  title={Futurex: An advanced live benchmark for llm agents in future prediction},
  author={Zeng, Zhiyuan and Liu, Jiashuo and Chen, Siyuan and He, Tianci and Liao, Yali and Tian, Yixiao and Wang, Jinpeng and Wang, Zaiyuan and Yang, Yang and Yin, Lingyue and others},
  journal={arXiv preprint arXiv:2508.11987},
  year={2025}
}

@inproceedings{
karger2025forecastbench,
title={ForecastBench: A Dynamic Benchmark of {AI} Forecasting Capabilities},
author={Ezra Karger and Houtan Bastani and Chen Yueh-Han and Zachary Jacobs and Danny Halawi and Fred Zhang and Philip Tetlock},
booktitle={International conference on learning representations},
year={2025},
}

@article{bianchi2025exploring,
  title={Exploring the use of AI authors and reviewers at Agents4Science},
  author={Bianchi, Federico and Queen, Owen and Thakkar, Nitya and Sun, Eric and Zou, James},
  journal={Nature Biotechnology},
  pages={1--4},
  year={2025},
  publisher={Nature Publishing Group US New York}
}

@article{yamada2025ai,
  title={The ai scientist-v2: Workshop-level automated scientific discovery via agentic tree search},
  author={Yamada, Yutaro and Lange, Robert Tjarko and Lu, Cong and Hu, Shengran and Lu, Chris and Foerster, Jakob and Clune, Jeff and Ha, David},
  journal={arXiv preprint arXiv:2504.08066},
  year={2025}
}

@article{gottweis2025towards,
  title={Towards an AI co-scientist},
  author={Gottweis, Juraj and Weng, Wei-Hung and Daryin, Alexander and Tu, Tao and Palepu, Anil and Sirkovic, Petar and Myaskovsky, Artiom and Weissenberger, Felix and Rong, Keran and Tanno, Ryutaro and others},
  journal={arXiv preprint arXiv:2502.18864},
  year={2025}
}

@article{mitchener2025kosmos,
  title={Kosmos: An ai scientist for autonomous discovery},
  author={Mitchener, Ludovico and Yiu, Angela and Chang, Benjamin and Bourdenx, Mathieu and Nadolski, Tyler and Sulovari, Arvis and Landsness, Eric C and Barabasi, Daniel L and Narayanan, Siddharth and Evans, Nicky and others},
  journal={arXiv preprint arXiv:2511.02824},
  year={2025}
}

@article{channing2025ai,
  title={AI for Scientific Discovery is a Social Problem},
  author={Channing, Georgia and Ghosh, Avijit},
  journal={arXiv preprint arXiv:2509.06580},
  year={2025}
}

@article{grace2018will,
  title={When will AI exceed human performance? Evidence from AI experts},
  author={Grace, Katja and Salvatier, John and Dafoe, Allan and Zhang, Baobao and Evans, Owain},
  journal={Journal of Artificial Intelligence Research},
  volume={62},
  pages={729--754},
  year={2018}
}

@article{bengio2024international,
  title={International scientific report on the safety of advanced ai (interim report)},
  author={Bengio, Yoshua and Mindermann, S{\"o}ren and Privitera, Daniel and Besiroglu, Tamay and Bommasani, Rishi and Casper, Stephen and Choi, Yejin and Goldfarb, Danielle and Heidari, Hoda and Khalatbari, Leila and others},
  journal={arXiv preprint arXiv:2412.05282},
  year={2024}
}

@article{phan2025humanity,
      title = {A benchmark of expert-level academic questions to assess {AI} capabilities},
      author = {{Center for AI Safety} and {Scale AI} and {HLE Contributors Consortium}},
      journal = {Nature},
      volume = {649},
      pages = {1139--1146},
      year = {2026},
      doi = {10.1038/s41586-025-09962-4},
}

@article{kaplan2020scaling,
  title={Scaling laws for neural language models},
  author={Kaplan, Jared and McCandlish, Sam and Henighan, Tom and Brown, Tom B and Chess, Benjamin and Child, Rewon and Gray, Scott and Radford, Alec and Wu, Jeffrey and Amodei, Dario},
  journal={arXiv preprint arXiv:2001.08361},
  year={2020}
}

@article{miles2010development,
  title={The development of technology foresight: A review},
  author={Miles, Ian},
  journal={Technological forecasting and social change},
  volume={77},
  number={9},
  pages={1448--1456},
  year={2010},
  publisher={Elsevier}
}

@article{moore1965moore,
  title={Moore’s law},
  author={Moore, Gordon},
  journal={Electronics Magazine},
  volume={38},
  number={8},
  pages={114},
  year={1965}
}

@article{gao2024empowering,
  title={Empowering biomedical discovery with AI agents},
  author={Gao, Shanghua and Fang, Ada and Huang, Yepeng and Giunchiglia, Valentina and Noori, Ayush and Schwarz, Jonathan Richard and Ektefaie, Yasha and Kondic, Jovana and Zitnik, Marinka},
  journal={Cell},
  volume={187},
  number={22},
  pages={6125--6151},
  year={2024},
  publisher={Elsevier}
}

@article{swanson2025virtual,
  title={The Virtual Lab of AI agents designs new SARS-CoV-2 nanobodies},
  author={Swanson, Kyle and Wu, Wesley and Bulaong, Nash L and Pak, John E and Zou, James},
  journal={Nature},
  volume={646},
  number={8085},
  pages={716--723},
  year={2025},
  publisher={Nature Publishing Group UK London}
}

@article{rees2026could,
  title={Could agentic AI topple grant-funding systems?},
  author={Rees, Geraint and Wilsdon, James},
  journal={Nature},
  volume={652},
  number={8112},
  pages={1119--1121},
  year={2026},
  publisher={Nature Publishing Group UK London}
}

@article{adam2025ai,
  title={When AI rejects your grant proposal: algorithms are helping to make funding decisions},
  author={Adam, David},
  journal={Nature},
  volume={645},
  number={8082},
  pages={832--833},
  year={2025},
  publisher={Nature}
}

@article{penades2025ai,
  title={AI mirrors experimental science to uncover a mechanism of gene transfer crucial to bacterial evolution},
  author={Penad{\'e}s, Jos{\'e} R and Gottweis, Juraj and He, Lingchen and Patkowski, Jonasz B and Daryin, Alexander and Weng, Wei-Hung and Tu, Tao and Palepu, Anil and Myaskovsky, Artiom and Pawlosky, Annalisa and others},
  journal={Cell},
  volume={188},
  number={23},
  pages={6654--6665},
  year={2025},
  publisher={Elsevier}
}

@article{jumper2021highly,
  title={Highly accurate protein structure prediction with AlphaFold},
  author={Jumper, John and Evans, Richard and Pritzel, Alexander and Green, Tim and Figurnov, Michael and Ronneberger, Olaf and Tunyasuvunakool, Kathryn and Bates, Russ and {\v{Z}}{\'\i}dek, Augustin and Potapenko, Anna and others},
  journal={Nature},
  volume={596},
  number={7873},
  pages={583--589},
  year={2021},
  publisher={Nature Publishing Group UK London}
}

@article{huang2025biomni,
  title={Biomni: A general-purpose biomedical ai agent},
  author={Huang, Kexin and Zhang, Serena and Wang, Hanchen and Qu, Yuanhao and Lu, Yingzhou and Roohani, Yusuf and Li, Ryan and Qiu, Lin and Li, Gavin and Zhang, Junze and others},
  journal={biorxiv},
  year={2025}
}

@inproceedings{jansen2025matter,
  title={Matter-of-fact: A benchmark for verifying the feasibility of literature-supported claims in materials science},
  author={Jansen, Peter and Hassan, Samiah and Wang, Ruoyao},
  booktitle={Empirical methods in natural language processing},
  year={2025}
}

@inproceedings{
yuan2025introducing,
title={Introducing {FOR}e{CA}st: The Future Outcome Reasoning and Confidence Assessment Benchmark},
author={Moy Yuan and Zifeng Ding and Andreas Vlachos},
booktitle={Advances in neural information processing systems datasets and benchmarks track},
year={2025},
}

@article{tao2025prophet,
  title={Prophet: An inferable future forecasting benchmark with causal intervened likelihood estimation},
  author={Tao, Zhengwei and Wu, Pu and Jin, Zhi and Bai, Xiaoying and Zhao, Haiyan and Dou, Chengfeng and Chen, Xiancai and Li, Jia and Li, Linyu and Tao, Chongyang and others},
  journal={arXiv preprint arXiv:2504.01509},
  year={2025}
}

@article{ajith2026prescience,
  title={PreScience: A Benchmark for Forecasting Scientific Contributions},
  author={Ajith, Anirudh and Singh, Amanpreet and DeYoung, Jay and Kunievsky, Nadav and Kozlowski, Austin C and Tafjord, Oyvind and Evans, James and Weld, Daniel S and Hope, Tom and Downey, Doug},
  journal={arXiv preprint arXiv:2602.20459},
  year={2026}
}

@inproceedings{lu2022learn,
  title={Learn to explain: Multimodal reasoning via thought chains for science question answering},
  author={Lu, Pan and Mishra, Swaroop and Xia, Tanglin and Qiu, Liang and Chang, Kai-Wei and Zhu, Song-Chun and Tafjord, Oyvind and Clark, Peter and Kalyan, Ashwin},
  booktitle={Advances in neural information processing systems},
  year={2022}
}

@article{lu2026towards,
  title={Towards end-to-end automation of AI research},
  author={Lu, Chris and Lu, Cong and Lange, Robert Tjarko and Yamada, Yutaro and Hu, Shengran and Foerster, Jakob and Ha, David and Clune, Jeff},
  journal={Nature},
  volume={651},
  number={8107},
  pages={914--919},
  year={2026},
  publisher={Nature Publishing Group UK London}
}

@inproceedings{lu2025solving,
  title={Solving inequality proofs with large language models},
  author={Sheng, Jiayi and Lyu, Luna and Jin, Jikai and Xia, Tanglin and Gu, Alex and Zou, James and Lu, Pan},
  booktitle={Advances in neural information processing systems},
  year={2025}
}

@inproceedings{lightman2023let,
  title={Let's verify step by step},
  author={Lightman, Hunter and Kosaraju, Vineet and Burda, Yuri and Edwards, Harrison and Baker, Bowen and Lee, Teddy and Leike, Jan and Schulman, John and Sutskever, Ilya and Cobbe, Karl},
  booktitle={International conference on learning representations},
  year={2023}
}

@inproceedings{matton2025walk,
  title={Walk the talk? measuring the faithfulness of large language model explanations},
  author={Matton, Katie and Ness, Robert and Guttag, John and Kiciman, Emre},
  booktitle={International conference on learning representations},
  year={2025}
}

@article{bank1971protein,
  title={Protein data bank},
  author={Bank, Protein Data},
  journal={Nature New Biol},
  volume={233},
  number={223},
  pages={10--1038},
  year={1971}
}

@article{wang2024nature,
  title={Nature of metal-support interaction for metal catalysts on oxide supports},
  author={Wang, Tairan and Hu, Jianyu and Ouyang, Runhai and Wang, Yutao and Huang, Yi and Hu, Sulei and Li, Wei-Xue},
  journal={Science},
  volume={386},
  number={6724},
  pages={915--920},
  year={2024},
  publisher={American Association for the Advancement of Science}
}

@article{schulz2026functional,
  title={Functional gradients facilitate tactile sensing in elephant whiskers},
  author={Schulz, Andrew K and Kaufmann, Lena V and Smith, Lawrence T and Philip, Deepti S and David, Hilda and Lazovic, Jelena and Brecht, Michael and Richter, Gunther and Kuchenbecker, Katherine J},
  journal={Science},
  volume={391},
  number={6786},
  pages={712--718},
  year={2026},
  publisher={American Association for the Advancement of Science}
}

@article{luo2025rpg,
  title={RPG: A Repository Planning Graph for Unified and Scalable Codebase Generation},
  author={Luo, Jane and Zhang, Xin and Liu, Steven and Wu, Jie and Liu, Jianfeng and Huang, Yiming and Huang, Yangyu and Yin, Chengyu and Xin, Ying and Zhan, Yuefeng and others},
  journal={arXiv preprint arXiv:2509.16198},
  year={2025}
}

@article{heng2024smart,
  title={A smart mask for exhaled breath condensate harvesting and analysis},
  author={Heng, Wenzheng and Yin, Shukun and Min, Jihong and Wang, Canran and Han, Hong and Shirzaei Sani, Ehsan and Li, Jiahong and Song, Yu and Rossiter, Harry B and Gao, Wei},
  journal={Science},
  volume={385},
  number={6712},
  pages={954--961},
  year={2024},
  publisher={American Association for the Advancement of Science}
}

@article{vukadinovic2026comprehensive,
  title={Comprehensive echocardiogram evaluation with view primed vision language AI},
  author={Vukadinovic, Milos and Chiu, I-Min and Tang, Xiu and Yuan, Neal and Chen, Tien-Yu and Cheng, Paul and Li, Debiao and Cheng, Susan and He, Bryan and Ouyang, David},
  journal={Nature},
  volume={650},
  number={8103},
  pages={970--977},
  year={2026},
  publisher={Nature Publishing Group UK London}
}

@article{yu2026infinidepth,
  title={InfiniDepth: Arbitrary-Resolution and Fine-Grained Depth Estimation with Neural Implicit Fields},
  author={Yu, Hao and Lin, Haotong and Wang, Jiawei and Li, Jiaxin and Wang, Yida and Zhang, Xueyang and Wang, Yue and Zhou, Xiaowei and Hu, Ruizhen and Peng, Sida},
  journal={arXiv preprint arXiv:2601.03252},
  year={2026}
}

@article{kang2025electromagnetic,
  title={Electromagnetic interference shielding using metal and MXene thin films},
  author={Kang, Geosan and Kwon, Guhyeon and Jeon, Jiwoon and Kwon, Jisung and Kim, Myung-Ki and Hong, Junpyo and Lee, Albert S and Lee, Seongi and Lee, Binhyung and Kim, Yujin and others},
  journal={Nature},
  pages={1--8},
  year={2025},
  publisher={Nature Publishing Group UK London}
}

@article{abramson2024accurate,
  title={Accurate structure prediction of biomolecular interactions with AlphaFold 3},
  author={Abramson, Josh and Adler, Jonas and Dunger, Jack and Evans, Richard and Green, Tim and Pritzel, Alexander and Ronneberger, Olaf and Willmore, Lindsay and Ballard, Andrew J and Bambrick, Joshua and others},
  journal={Nature},
  volume={630},
  number={8016},
  pages={493--500},
  year={2024},
  publisher={Nature Publishing Group UK London}
}

@misc{openai2026gpt54,
  title={Introducing {GPT}-5.4},
  author={OpenAI},
  year={2026},
  month={March},
  howpublished={\url{https://openai.com/index/introducing-gpt-5-4/}}
}

@article{hurst2024gpt,
  title={Gpt-4o system card},
  author={Hurst, Aaron and Lerer, Adam and Goucher, Adam P and Perelman, Adam and Ramesh, Aditya and Clark, Aidan and Ostrow, AJ and Welihinda, Akila and Hayes, Alan and Radford, Alec and others},
  journal={arXiv preprint arXiv:2410.21276},
  year={2024}
}

@misc{claudesonnet,
  title={Introducing Claude Sonnet 4.5},
  author={Anthropic},
  year={2025},
  month={September},
  howpublished={\url{https://www.anthropic.com/news/claude-sonnet-4-5}}
}

@article{grattafiori2024llama,
  title={The llama 3 herd of models},
  author={Grattafiori, Aaron and Dubey, Abhimanyu and Jauhri, Abhinav and Pandey, Abhinav and Kadian, Abhishek and Al-Dahle, Ahmad and Letman, Aiesha and Mathur, Akhil and Schelten, Alan and Vaughan, Alex and others},
  journal={arXiv preprint arXiv:2407.21783},
  year={2024}
}

@article{agarwal2025gpt,
  title={gpt-oss-120b \& gpt-oss-20b model card},
  author={Agarwal, Sandhini and Ahmad, Lama and Ai, Jason and Altman, Sam and Applebaum, Andy and Arbus, Edwin and Arora, Rahul K and Bai, Yu and Baker, Bowen and Bao, Haiming and others},
  journal={arXiv preprint arXiv:2508.10925},
  year={2025}
}

@article{guo2025deepseek,
  title={DeepSeek-R1 incentivizes reasoning in LLMs through reinforcement learning},
  author={Guo, Daya and Yang, Dejian and Zhang, Haowei and Song, Junxiao and Wang, Peiyi and Zhu, Qihao and Xu, Runxin and Zhang, Ruoyu and Ma, Shirong and Bi, Xiao and others},
  journal={Nature},
  volume={645},
  number={8081},
  pages={633--638},
  year={2025},
  publisher={Nature Publishing Group UK London}
}

@misc{goel2026futuresimreplayingworldevents,
      title={FutureSim: Replaying World Events to Evaluate Adaptive Agents}, 
      author={Shashwat Goel and Nikhil Chandak and Arvindh Arun and Ameya Prabhu and Steffen Staab and Moritz Hardt and Maksym Andriushchenko and Jonas Geiping},
      year={2026},
      eprint={2605.15188},
      archivePrefix={arXiv},
      primaryClass={cs.LG},
      url={https://arxiv.org/abs/2605.15188}, 
}

\clearpage
\label{MainTextLastPage}
\renewcommand{\MainLastPage}{MainTextLastPage}
\clearpage
\appendix

\setcounter{page}{1}

\renewcommand{\figurename}{Supplementary Figure}
\renewcommand{\tablename}{Supplementary Table}
\renewcommand{\thefigure}{S\arabic{figure}}
\renewcommand{\thetable}{S\arabic{table}}
\setcounter{figure}{0}
\setcounter{table}{0}

\fancyhf{}                      
\rfoot{\small\sffamily\bfseries\thepage}
\renewcommand{\headrulewidth}{0pt}
\renewcommand{\footrulewidth}{0pt}

\pagestyle{fancy}
\thispagestyle{fancy}


\renewcommand{\rmdefault}{phv}   
\renewcommand{\sfdefault}{phv}   
\renewcommand{\familydefault}{\sfdefault}
\setlength{\parindent}{0pt}      

{\LARGE\bfseries Supplementary Information for}\\[6pt]
{\LARGE\bfseries Scientific reasoning does not reliably translate into scientific forecasting in frontier AI}

\vspace{2em}

\renewcommand{\figurename}{Supplementary Figure}  
\renewcommand{\tablename}{Supplementary Table}  
\setcounter{figure}{0}
\setcounter{table}{0}


\section{\CUSP\ Construction Details}

\subsection{Data acquisition and source construction}
\label{sec:data_acquisition}
\paragraph{Natural Science Data:} We extract primary claims, experimental outcomes, and contextual data from the publication logs of \textit{Nature}, \textit{Science}, and \textit{Cell} to capture foundational breakthroughs in the physical and life sciences. We restrict the natural science corpus strictly to these high-impact, peer-reviewed journals to guarantee that the forecasted milestones in biology, chemistry, and physics represent rigorously validated empirical discoveries rather than preliminary hypotheses or unverified preprints. To account for the potential temporal lag between initial preprints and formal journal publication, we query a combination of academic APIs, including Crossref, Semantic Scholar, OpenAlex, Europe PMC, arXiv, and bioRxiv/medRxiv, and define the earliest observed date of each manuscript’s DOI across these sources as a strict knowledge cutoff to prevent temporal leakage.

\paragraph{Artificial Intelligence Data:} We incorporate artificial intelligence as a core evaluation of \CUSP\ because algorithmic advancements increasingly dictate the pace and direction of discovery across the natural sciences. 
For example, breakthroughs like AlphaFold have been fundamentally catalyzed by both prior computational innovations--especially attention mechanisms and transformer architectures--and the availability of large-scale dataset resources such as the Protein Data Bank \cite{bank1971protein}.
Accurately anticipating milestones such as new architectures, algorithms, and datasets is therefore essential for forecasting the broader trajectory of scientific progress. Furthermore, evaluating frontier models on AI-specific targets provides a rigorous empirical basis for assessing their capacity to reason about the evolution of their underlying technologies.

\CUSP\ includes high-visibility AI papers from dynamic, community-driven repositories, specifically incorporating widely acknowledged “Top 10 AI Papers of the Week” lists and the Hugging Face Top Papers hub. To filter top papers from Hugging Face, we rank the most impactful research using a hybrid impact score that balances community engagement (upvotes) against academic traction (citations). The score is given by $\text{upvotes} + 5 \times [\text{citations} / (\text{months old} + 1)]$ and provides an age-adjusted metric to ensure that recent, high-velocity publications remain competitive with older, more highly cited works. We subsequently select the highest-scoring papers from each month.

\paragraph{Leaderboard Forecasting Data: }In addition to publication-based milestones, we incorporate forecasting targets from widely used AI benchmarks and leaderboards, which provide standardized and time-resolved measurements of progress in machine learning. Representative benchmarks include GPQA Diamond, MMLU-Pro, and Humanity’s Last Exam, along with domain-specific leaderboards tracking rapid capability improvements. Forecasting these benchmarks tests whether AI systems can extrapolate how capabilities evolve over time, rather than recall past results. Compared to heterogeneous scientific discoveries, leaderboard progress is more structured and temporally dense, offering a complementary setting to assess whether models can internalize and predict technological advancement. \CUSP\ combines these targets with publication-derived milestones to evaluate whether AI systems exhibit a coherent sense of progress across both performance scaling and scientific breakthroughs.

\subsection{Corpus Extraction and Automated Filtering}
To construct a benchmark to evaluate scientific forecasting, we focus on extracting \emph{verifiable scientific milestones} from papers that contain concrete results for predictive evaluation. 
We collect top-tier publications across multiple disciplines and restrict data extraction to the title, publication metadata, and abstract of each paper to ensure a consistent and standardized representation of each milestone. 
Abstracts provide a concise and structured representation of scientific contributions in papers, summarizing the primary claims, core methods, and key quantitative outcomes.
This design also ensures reproducibility and accessibility, as abstracts are uniformly available across publication venues.

We design an LLM-based agentic pipeline to extract and filter candidate milestones from the collected abstracts automatically. 
The extraction stage identifies statements corresponding to concrete scientific results or capabilities, while the filtering stage operates as a strict binary classifier that accepts only entries containing at least one verifiable and measurable outcome. 
This process ensures that each retained instance corresponds to a well-defined prediction target.

To improve precision, we incorporate domain-aware filtering by classifying each abstract into a scientific domain (e.g., AI, Chemistry, Biology, Physics, or General Science) and applying domain-specific acceptance criteria. 
For computational research, accepted entries must report validated technical advances on recognized benchmarks or provide explicit performance metrics. 
For experimental sciences, entries must include measurable quantities (e.g., binding affinities, fold changes) or clearly defined physical or biological properties.

We explicitly reject abstracts that are purely descriptive, speculative, or review-oriented, as these lack quantitative outcomes suitable for predictive evaluation. 
For each decision, the pipeline generates a concise summary of the extracted result for accepted entries or a justification for rejection. 
This automated procedure ensures scalability and consistency while minimizing subjective annotation bias. 
Box~\ref{box:rejected_example} illustrates that even high-impact publications may fail to meet the criteria of verifiable scientific milestones.

\subsection{Task synthesis procedure}
\label{sec:task_synthesis}

To synthesize forecasting tasks from each accepted scientific milestone, we first decompose the source abstract into three structured fields. This decomposition separates the scientific problem, the technical solution and the resulting measurable outcome before any task generation step. We avoid novel acronyms, method names and system names introduced in the source paper, so that models evaluated under a historical knowledge cutoff cannot identify a discovery from post-cutoff terminology.

\begin{description}
    \item[Problem Statement:] A technical description focusing exclusively on the research problem and the limitations of preceding methods.
    \item[Technical Approach:] A detailed, method-oriented specification of the mechanism, experimental design or architectural innovation.
    \item[Results and Metrics:] A single-sentence summary capturing only quantitative outcomes, performance numbers or benchmark results.
\end{description}

The task-generation pipeline uses these fields to construct five evaluation formats. Binary questions assess whether the milestone claim is feasible by the target date. Perturbed binary questions modify the original claim to create plausible but unsupported alternatives. Multiple-choice questions ask models to identify the correct technical approach among expert-level distractors. Free-response questions ask models to propose an implementation strategy from the problem context. Date prediction tasks ask models to forecast the month and year in which the milestone is realized. We use GPT-4o to convert structured abstract decompositions into candidate binary, perturbed binary, MCQ, FRQ, and date-prediction tasks, including extraction of the candidate correct answer from the source abstract. Final task inclusion and answer validity are determined by the source publication itself and independently checked by Grok-3 and human review.

Representative examples of each task format are provided in Appendix~\ref{sec:task_examples}.

\subsection{Key Statistics and Distributional Properties}
\label{sec:statistics}

\begin{figure}[t]
    \centering
    \includegraphics[width=\textwidth]{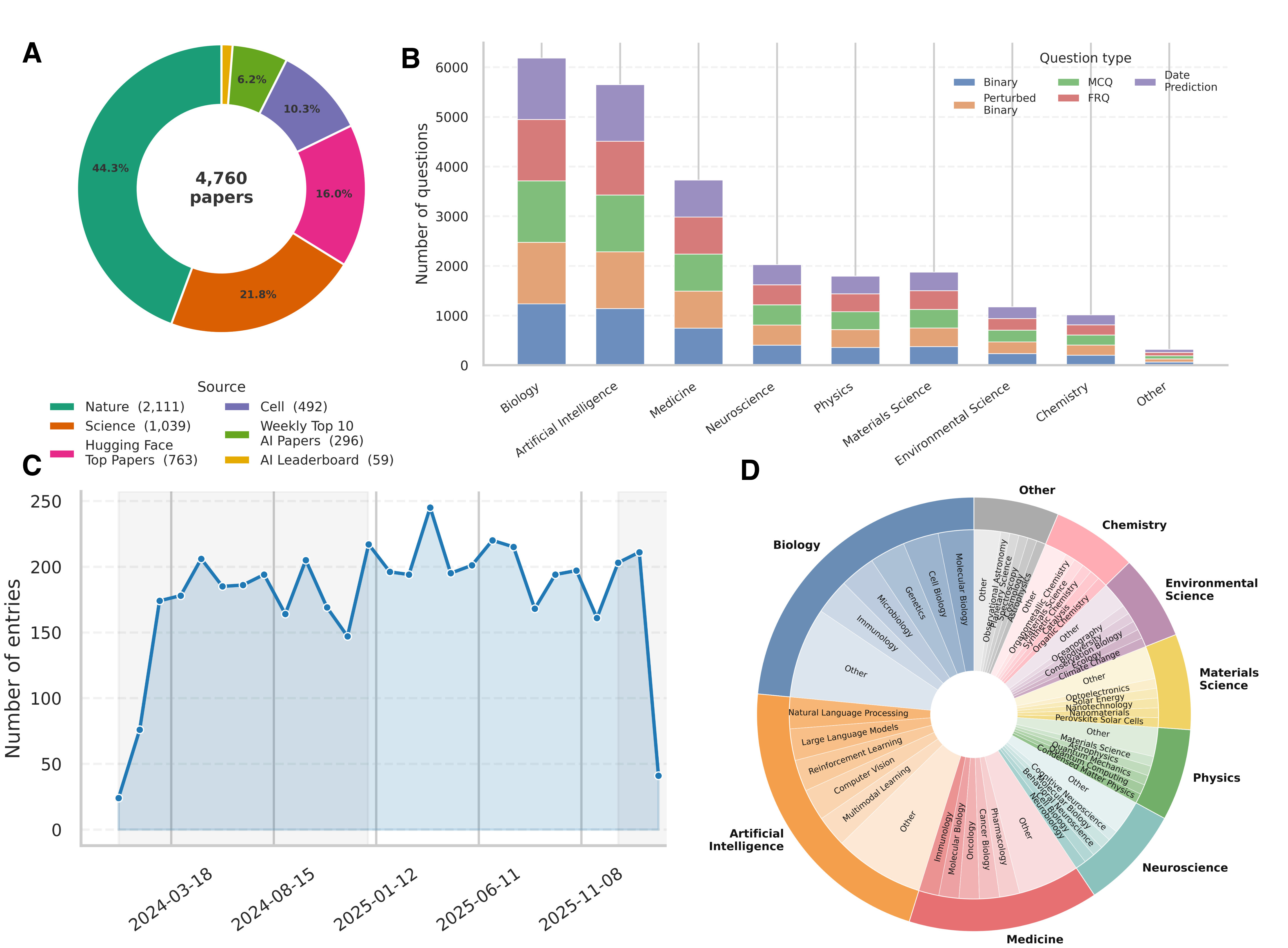}
 \caption{\textbf{A) Source Distribution:} Breakdown of the 4,760 scientific milestones by publication venue. \textbf{B) Task Density by Domain:} Distribution of the 17,429 validated tasks across eight top-level domains. \textbf{C) Temporal Information:} Longitudinal count of entries from January 2024 to March 2026. \textbf{D) Multi-Disciplinary Taxonomy:} Sunburst visualization of distinct subcategories. \textbf{E) Human vs. AI Keep Rates:} Calibration of the Grok-3 validation pipeline against graduate-level human experts. \textbf{F) Validation Agreement:} Reliability metrics of the automated judge, showing high precision across all question modalities.}    \label{fig:single_column}
 \end{figure}
We analyze the temporal, structural, and linguistic properties of \CUSP\ to characterize the diversity and complexity of the benchmark.

\paragraph{Temporal Distribution.}
\CUSP\ spans a continuous time horizon from January 2024 to March 2026, covering 27 active months of scientific progress. The dataset includes 4,760 milestones with valid publication timestamps, with all 27 months represented. The distribution of milestones over time is relatively stable, with an average of 176.3 papers per month ($\pm$ 51.5). This temporal structure enables controlled evaluation across different forecasting horizons, supporting analysis of short-term versus long-term prediction and the degradation of model performance as temporal distance increases. 

\paragraph{Task Composition.}
Each scientific milestone is decomposed into multiple task formats designed to probe complementary aspects of scientific forecasting. In total, the validated dataset contains 17,429 task instances across four task types. The distribution of task types is non-uniform, reflecting strict validation and filtering constraints: 4,128 multiple-choice questions (MCQs), 4,135 free-response (FRQ) prompts, 3,656 perturbed binary tasks, and 2,755 each of binary and date prediction tasks. 
Task availability varies across milestones, resulting in a heterogeneous task density. 
Specifically, 54.0\% of papers admit all four task types, while 27.1\%, 15.9\%, and 3.0\% of papers admit three, two, and one task(s), respectively. 
This sparsity arises from strict LLM-as-a-judge validation filtering, which removes tasks that lack verifiability, faithfulness, or logical consistency under perturbation. As a result, \CUSP\ prioritizes task reliability and scientific validity over uniformity.

\paragraph{Label Distribution.}
For binary questions, the ground-truth answer is \emph{yes}, whereas perturbed binary questions have the ground-truth answer \emph{no}, resulting in an overall yes/no label distribution of approximately 0.75. For multiple-choice questions (MCQs), the correct answer is initially assigned to option A, and answer choices are randomly shuffled during evaluation to ensure a uniform distribution of correct options. For date prediction tasks, the distribution of ground-truth dates remains relatively stable across the dataset (Figure~\ref{fig:single_column}).

\paragraph{Question Length and Complexity.}
We analyze the linguistic characteristics of generated tasks across formats. Binary questions have an average length of 29.2 words (variance 122.5), while MCQs average 36.6 words (variance 47.8) and FRQs average 41.8 words (variance 17.8). Problem statements, which encode the underlying scientific challenge, are substantially longer, with an average length of 70.6 words (variance 107.0).

These differences reflect increasing levels of reasoning complexity across task types. Problem statements provide rich scientific context, FRQs require open-ended methodological synthesis, and MCQs demand precise discriminative reasoning over expert-level distractors.

\paragraph{Domain and Subcategory Diversity.}
\CUSP\ spans eight top-level scientific domains and 4,245 distinct subcategories, reflecting highly specialized research topics across disciplines. The dataset is dominated by biology (1,234 papers) and artificial intelligence (1,141 papers), followed by medicine (746), neuroscience (403), materials science (375), physics (359), environmental science (235), chemistry (203), and other domains (64). 


\section{Benchmark validation}
\label{sec:benchmark_validation}

Because \CUSP\ evolves continuously with newly published discoveries, it is essential to maintain an automated mechanism for verifying the faithfulness and quality of its generated questions. To this end, we develop a validation framework that employs a large language model (LLM) judge to evaluate each question against its source abstract. Since \CUSP\ questions are generated using GPT-4o, we perform validation using a distinct model, Grok-3 (xAI). As a frontier model with comparable scale and capability but a different architecture and training distribution, Grok-3 provides an independent evaluation, mitigating potential self-evaluation bias.

\paragraph{Binary and Perturbed Binary Questions.}
We first validate binary and perturbed binary questions under three criteria:

\begin{itemize}
\item \textbf{Faithfulness:} The binary statement (excluding the temporally conditioned phrasing) must accurately reflect a concrete claim in the source abstract. The validator checks for discrepancies in entities, conditions, outcomes, or quantitative details, and rejects questions that introduce unsupported or altered claims.

\item \textbf{Verifiability:} The binary statement must correspond to a concrete, objectively evaluable claim. We reject questions whose underlying statements are vague, underspecified, or lack a clear operational meaning, ensuring that each question admits a well-defined yes/no answer.
\begin{figure}[t]
    \centering
    \includegraphics[width=\textwidth]{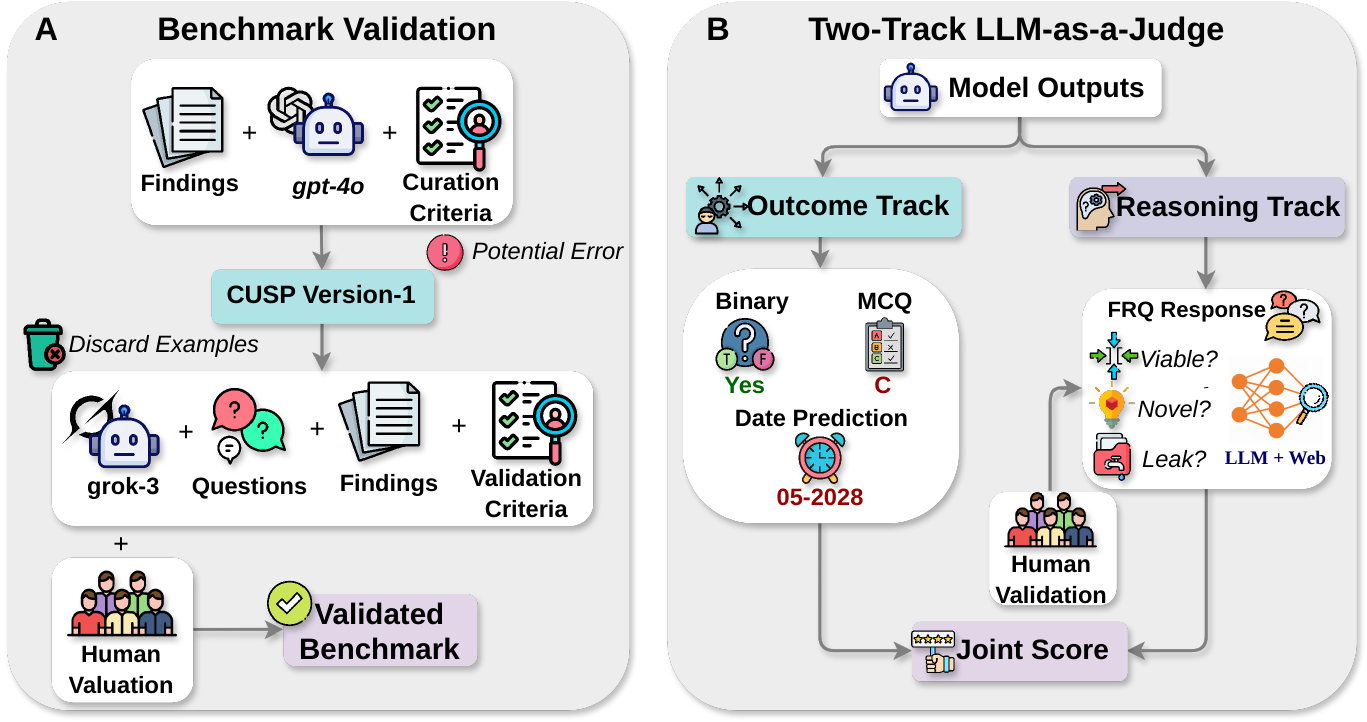}
\caption{\textbf{\CUSP\ validation and evaluation pipeline.} 
(A) Benchmark construction: scientific findings are curated and filtered via LLM-based criteria, validated by an independent model, and verified by human experts to produce a high-quality benchmark. 
(B) Two-track evaluation: model outputs are assessed for \emph{outcome correctness} (across binary, MCQ, and date prediction tasks) and \emph{reasoning quality} (viability, novelty, and leakage) for FRQ, which are combined into a joint score.}
\label{fig:eval_pipeline}
 \end{figure}

\item \textbf{Perturbation Validity:} For perturbed binary questions, we verify that the perturbation introduces a meaningful and non-trivial modification to the original claim (e.g., modifying thresholds or adding unmet constraints), such that the perturbed version is no longer directly supported by the source abstract. This prevents trivial or paraphrased perturbations.
\end{itemize}

\paragraph{Multiple-Choice Questions (MCQ).}
For MCQ tasks, we decompose validation into three components. First, we ensure that the problem statement is faithful to the source abstract and accurately captures the underlying scientific challenge. Second, we verify that the correct answer choice corresponds to a valid technical approach or mechanism that is supported or implied by the abstract. Third, we evaluate the distractor options, ensuring that incorrect choices are plausible yet not directly supported by the source. This prevents trivial elimination and enforces that MCQs require genuine mechanistic forecasting rather than superficial pattern matching.

\paragraph{Free-Response Questions (FRQ).}
For open-ended free-response tasks, direct validation of generated answers is inherently ambiguous due to the existence of multiple valid solutions. Instead, we validate the problem context and background description, ensuring that the prompt is faithful to the source abstract and accurately reflects the scientific problem being addressed.

All validation criteria are implemented as separate LLM judgments with structured outputs (verdict, score, and explanation). Based on these evaluations, we perform fine-grained filtering at the field level, removing only invalid components (e.g., a faulty binary or perturbed question) while preserving the remainder of the sample. This enables scalable quality control without sacrificing dataset coverage. Additionally, we remove date prediction tasks when the corresponding binary question is filtered out, as both depend on the same underlying verifiable result. See Appendix~\ref{sec:prompts} for the full set of validation prompts. To assess the reliability of this automated validation framework, we also conducted a human evaluation, see Appendix~\ref{sec:question_human_eval}.

\pagebreak

\section{Forecasting in a time capsule}
\label{Appendix:time_capsule}

\begin{table*}[t]
\centering
\small
\setlength{\tabcolsep}{3pt}
\renewcommand{\arraystretch}{1.10}

\caption{\textbf{Forecasted performance of frontier large language models on
reasoning benchmarks.} Predicted accuracy for six models at three forecast
horizons. Values in parentheses denote the predicted gain (percentage points)
relative to the state-of-the-art baseline measured in April~2026.
\textbf{a}, Humanity's Last Exam, reported as no-tools / with-tools.
\textbf{b}, GPQA Diamond / MMMLU.}
\label{tab:benchmark_predictions}

\begin{adjustbox}{max width=\textwidth}
\begin{tabular}{@{}lccc@{}}
\toprule

\multirow{2}{*}{\textbf{Model}}
& \multicolumn{3}{c}{\textbf{Forecast date}} \\

\cmidrule(lr){2-4}

& \textbf{2026-10}
& \textbf{2027-04}
& \textbf{2027-10} \\

\midrule
\multicolumn{4}{@{}l}{\textbf{a}\quad\textit{Humanity's Last Exam} \;|\; current state-of-the-art (2026-04): 56.8\% / 64.7\%} \\
\midrule

\llmicon{icons/openai_logo.png} GPT-5.4
& \makecell[c]{63\% {\color{green!40!black}($\uparrow$6.2)} / 74\% {\color{green!40!black}($\uparrow$9.3)}}
& \makecell[c]{72\% {\color{green!40!black}($\uparrow$15.2)} / 78\% {\color{green!40!black}($\uparrow$13.3)}}
& \makecell[c]{74\% {\color{green!40!black}($\uparrow$17.2)} / 82\% {\color{green!40!black}($\uparrow$17.3)}} \\

\llmicon{icons/claude.png} Claude S4.5
& \makecell[c]{62\% {\color{green!40!black}($\uparrow$5.2)} / 71\% {\color{green!40!black}($\uparrow$6.3)}}
& \makecell[c]{64\% {\color{green!40!black}($\uparrow$7.2)} / 72\% {\color{green!40!black}($\uparrow$7.3)}}
& \makecell[c]{68\% {\color{green!40!black}($\uparrow$11.2)} / 73\% {\color{green!40!black}($\uparrow$8.3)}} \\

\llmicon{icons/deepseek_logo.png} DeepSeek R1
& \makecell[c]{59\% {\color{green!40!black}($\uparrow$2.2)} / 69.5\% {\color{green!40!black}($\uparrow$4.8)}}
& \makecell[c]{64\% {\color{green!40!black}($\uparrow$7.2)} / 72\% {\color{green!40!black}($\uparrow$7.3)}}
& \makecell[c]{62\% {\color{green!40!black}($\uparrow$5.2)} / 78\% {\color{green!40!black}($\uparrow$13.3)}} \\

\llmicon{icons/llama.png} LLaMA 3.3
& \makecell[c]{62\% {\color{green!40!black}($\uparrow$5.2)} / 71\% {\color{green!40!black}($\uparrow$6.3)}}
& \makecell[c]{65\% {\color{green!40!black}($\uparrow$8.2)} / 75\% {\color{green!40!black}($\uparrow$10.3)}}
& \makecell[c]{65\% {\color{green!40!black}($\uparrow$8.2)} / 75\% {\color{green!40!black}($\uparrow$10.3)}} \\

\llmicon{icons/openai_logo.png} GPT-OSS
& \makecell[c]{68\% {\color{green!40!black}($\uparrow$11.2)} / 75\% {\color{green!40!black}($\uparrow$10.3)}}
& \makecell[c]{70\% {\color{green!40!black}($\uparrow$13.2)} / 73\% {\color{green!40!black}($\uparrow$8.3)}}
& \makecell[c]{70\% {\color{green!40!black}($\uparrow$13.2)} / 78\% {\color{green!40!black}($\uparrow$13.3)}} \\

\llmicon{icons/openai_logo.png} GPT-4o
& \makecell[c]{62\% {\color{green!40!black}($\uparrow$5.2)} / 68\% {\color{green!40!black}($\uparrow$3.3)}}
& \makecell[c]{65\% {\color{green!40!black}($\uparrow$8.2)} / 72\% {\color{green!40!black}($\uparrow$7.3)}}
& \makecell[c]{68\% {\color{green!40!black}($\uparrow$11.2)} / 78\% {\color{green!40!black}($\uparrow$13.3)}} \\

\midrule
\multicolumn{4}{@{}l}{\textbf{b}\quad\textit{GPQA Diamond / MMMLU} \;|\; current state-of-the-art (2026-04): 94.6\% / 92.6\%} \\
\midrule

\llmicon{icons/openai_logo.png} GPT-5.4
& \makecell[c]{96.2\% {\color{green!40!black}($\uparrow$1.6)} / 93.8\% {\color{green!40!black}($\uparrow$1.2)}}
& \makecell[c]{97.2\% {\color{green!40!black}($\uparrow$2.6)} / 95.1\% {\color{green!40!black}($\uparrow$2.5)}}
& \makecell[c]{97.8\% {\color{green!40!black}($\uparrow$3.2)} / 95.8\% {\color{green!40!black}($\uparrow$3.2)}} \\

\llmicon{icons/claude.png} Claude S4.5
& \makecell[c]{96.5\% {\color{green!40!black}($\uparrow$1.9)} / 94.2\% {\color{green!40!black}($\uparrow$1.6)}}
& \makecell[c]{97.5\% {\color{green!40!black}($\uparrow$2.9)} / 94.2\% {\color{green!40!black}($\uparrow$1.6)}}
& \makecell[c]{97.5\% {\color{green!40!black}($\uparrow$2.9)} / 95.2\% {\color{green!40!black}($\uparrow$2.6)}} \\

\llmicon{icons/deepseek_logo.png} DeepSeek R1
& \makecell[c]{96\% {\color{green!40!black}($\uparrow$1.4)} / 93.3\% {\color{green!40!black}($\uparrow$0.7)}}
& \makecell[c]{96.2\% {\color{green!40!black}($\uparrow$1.6)} / 94.1\% {\color{green!40!black}($\uparrow$1.5)}}
& \makecell[c]{96.2\% {\color{green!40!black}($\uparrow$1.6)} / 94\% {\color{green!40!black}($\uparrow$1.4)}} \\

\llmicon{icons/llama.png} LLaMA 3.3
& \makecell[c]{96.2\% {\color{green!40!black}($\uparrow$1.6)} / 94.2\% {\color{green!40!black}($\uparrow$1.6)}}
& \makecell[c]{96.2\% {\color{green!40!black}($\uparrow$1.6)} / 94.2\% {\color{green!40!black}($\uparrow$1.6)}}
& \makecell[c]{96.2\% {\color{green!40!black}($\uparrow$1.6)} / 95\% {\color{green!40!black}($\uparrow$2.4)}} \\

\llmicon{icons/openai_logo.png} GPT-OSS
& \makecell[c]{95.8\% {\color{green!40!black}($\uparrow$1.2)} / 94.2\% {\color{green!40!black}($\uparrow$1.6)}}
& \makecell[c]{95.8\% {\color{green!40!black}($\uparrow$1.2)} / 93.5\% {\color{green!40!black}($\uparrow$0.9)}}
& \makecell[c]{96.5\% {\color{green!40!black}($\uparrow$1.9)} / 94.2\% {\color{green!40!black}($\uparrow$1.6)}} \\

\llmicon{icons/openai_logo.png} GPT-4o
& \makecell[c]{95.2\% {\color{green!40!black}($\uparrow$0.6)} / 94.2\% {\color{green!40!black}($\uparrow$1.6)}}
& \makecell[c]{96.8\% {\color{green!40!black}($\uparrow$2.2)} / 94.5\% {\color{green!40!black}($\uparrow$1.9)}}
& \makecell[c]{96.8\% {\color{green!40!black}($\uparrow$2.2)} / 95.5\% {\color{green!40!black}($\uparrow$2.9)}} \\

\bottomrule
\end{tabular}
\end{adjustbox}
\end{table*}
We use \CUSP\ Time Capsule to study how frontier models anticipate future (beyond April 2026) scientific and AI progress. We evaluate both open-ended scientific forecasting and capability prediction on benchmarks, asking models to extrapolate future breakthroughs.

\begin{wrapfigure}{r}{0.5\textwidth}
    \centering
        \caption{Forecasts of global CO$_2$ emissions.}
    \label{fig:env_predict}
    \includegraphics[width=0.48\textwidth]{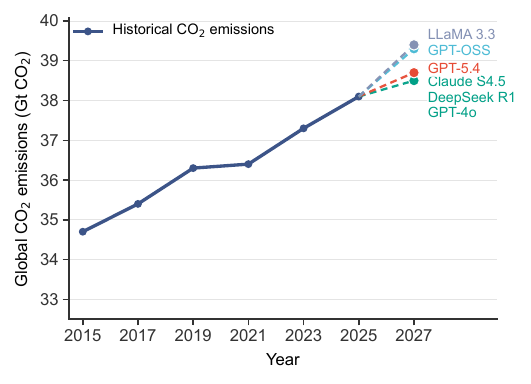}
\end{wrapfigure}

Figure~\ref{fig:env_predict} shows models' predictions of global CO$_2$ emissions in 2027. All models forecast emissions to remain above the 2025 level, indicating a shared expectation of continued near-term growth in global emissions. However, the predicted magnitude varies across models. Claude S4.5, DeepSeek R1, and GPT-4o produce comparatively conservative estimates close to the historical trend, whereas GPT-5.4 predicts a slightly larger increase. LLaMA 3.3 and GPT-OSS forecast the steepest growth, with LLaMA 3.3 producing the highest projected emissions overall. These results suggest that while models broadly agree on the direction of future emissions, they differ substantially in their expectations about the pace of global decarbonization.

Table~\ref{tab:benchmark_predictions} summarizes models' forecasts of future AI capabilities on a range of multi-disciplinary evaluation benchmarks, including Humanity's Last Exam (with and without tools), GPQA Diamond, and MMMLU \cite{hendrycksmeasuring}. Across models, we observe a shared expectation of continued capability gains over the 2026-2027 horizon, though the projected magnitude varies substantially. GPT-5.4 produces the most optimistic forecasts, especially for 2027-10, predicting Humanity's Last Exam performance to rise from 56.8\% to 74\% without tools and from 64.7\% to 82\% with tools. It also forecasts near-saturation performance on GPQA Diamond and MMMLU by late 2027. In contrast, DeepSeek R1 consistently makes more conservative predictions, with smaller gains and earlier plateaus, particularly on already high-performing benchmarks.
A broader trend is that forecasts for GPQA Diamond and MMMLU are tightly clustered near the upper performance bound, suggesting that models expect these benchmarks to saturate within the next few generations. Humanity's Last Exam, especially in the tool-augmented setting, shows substantially greater variance, indicating that models perceive it as a more open-ended and capability-sensitive benchmark. 

Together, these results highlight the potential of \CUSP\ Time Capsule as a framework for studying not only factual recall and forecasting, but also the implicit scientific priors and future-oriented world models embedded within frontier AI systems.

\pagebreak

 \section{Related Works}
 \label{sec:comparison_to_benchmarks}
 \subsection{Comparison to Related Benchmarks}
 A growing body of work has introduced benchmarks for evaluating forecasting ability and scientific reasoning in AI systems. Table~\ref{tab:benchmark_comparison} summarizes the relationship between CUSP and these prior efforts. Recent forecasting benchmarks such as ForecastBench \cite{karger2025forecastbench}, FutureX \cite{zeng2025futurex}, FOReCAst \cite{yuan2025introducing}, and PROPHET \cite{tao2025prophet} evaluate the ability of models to predict future events, often incorporating temporal reasoning and, in some cases, confidence calibration or retrieval. However, these benchmarks focus primarily on general-world or news-driven events and do not explicitly target scientific discovery. In addition, these benchmarks do not explicitly control for model knowledge cutoffs, making it difficult to distinguish genuine forecasting ability from memorization or indirect exposure to future information.

\begin{figure*}[h]
    \centering
    \includegraphics[width=\textwidth]{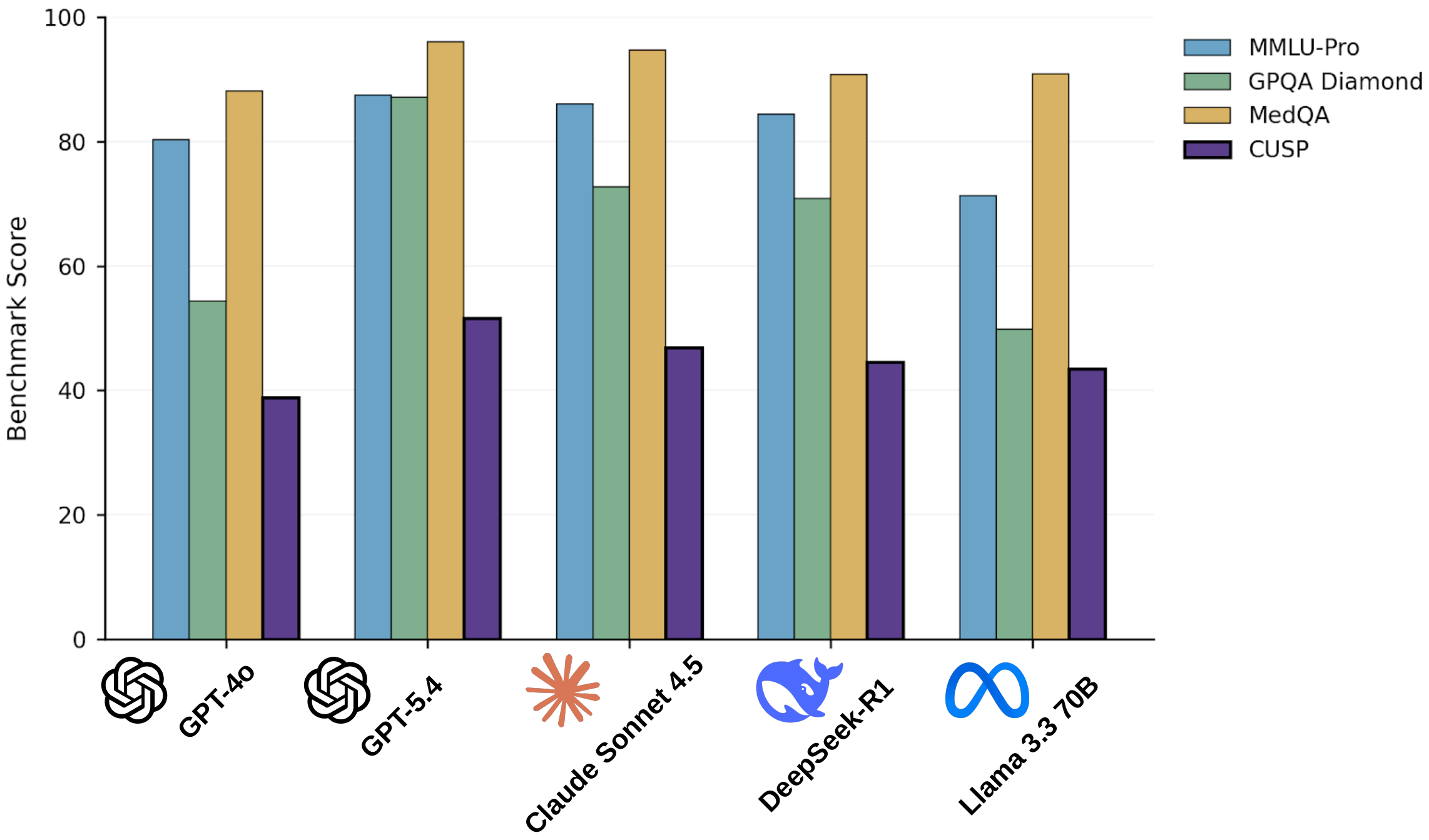}
\caption{Saturation plot of CUSP compared to other commonly used LLM benchmarks.}
    \label{fig:saturation}
\end{figure*}

 In parallel, scientific reasoning and discovery benchmarks such as Humanity’s Last Exam \cite{phan2025humanity}, AstaBench \cite{bragg2025astabench}, PreScience \cite{ajith2026prescience}, ResearchBench \cite{liu2025researchbench}, and ScienceQA \cite{lu2022learn} evaluate models on tasks including expert-level question answering, hypothesis generation, and structured reasoning over scientific content. While these benchmarks provide strong tests of domain knowledge and reasoning, they are inherently retrospective: the correct answers are known at evaluation time, and models are not required to anticipate future discoveries. In addition, many widely used scientific reasoning benchmarks are increasingly saturated by frontier AI systems, limiting their ability to discriminate further capability improvements. Benchmarks such as MMLU-Pro, GPQA Diamond, and MedQA now exhibit near-ceiling performance for leading models, despite continuing gaps in scientific forecasting ability. \CUSP\ remains substantially unsaturated across all evaluated frontier systems (Figure~\ref{fig:saturation}), suggesting that forecasting scientific progress constitutes a qualitatively more challenging setting than retrospective reasoning over established scientific knowledge.

 \CUSP\ differs fundamentally from both lines of work. Unlike forecasting benchmarks, it is explicitly grounded in scientific discovery, with all tasks derived from real, peer-reviewed breakthroughs across multiple disciplines. Unlike existing scientific benchmarks, it introduces a temporal cutoff that restricts models to knowledge available prior to each milestone, thereby requiring genuine forward-looking prediction rather than retrospective reasoning. Furthermore, \CUSP\ employs a multi-task evaluation framework that combines binary feasibility prediction, calibration under perturbation, mechanistic forecasting, generative proposal, and date prediction within a unified setting. These properties position \CUSP\ as the first benchmark to systematically evaluate AI systems as epistemic forecasters of scientific progress, bridging the gap between general forecasting and scientific reasoning benchmarks.

 \subsection{AI for Science}
 \paragraph{AI for Scientific Discovery.} 
 AI for Science is a rapidly expanding field that is increasingly integrated into scientific discovery, spanning from computational tools to automated research systems that support full research workflows \cite{bianchi2025exploring, yamada2025ai,lu2026towards, gottweis2025towards, mitchener2025kosmos}. 
 While prior work has demonstrated the potential of large language models (LLMs) to assist scientific discovery, these approaches typically rely on human researchers to define the problems and directions to explore \cite{channing2025ai, grace2018will, bengio2024international}.
 As a result, it remains unclear whether LLMs can independently identify promising and feasible research directions and proactively explore the scientific frontier.
 This work makes an initial exploration by studying the capability boundaries of LLMs in scientific forecasting.
 To enable this investigation, we introduce \CUSP, a benchmark and scalable evaluation framework that operationalizes scientific forecasting as a measurable capability, requiring models to predict concrete and verifiable discoveries that emerge only after a strict knowledge cutoff date.

\pagebreak
\section{Extended Results}
\label{sec:extended_results}

\subsection{Full Web-search Results}
\begin{table}[t]
\centering
\small
\setlength{\tabcolsep}{4pt}
\begin{tabular}{lcccccccc}
\toprule
 & & \multicolumn{3}{c}{Web Search (no cutoff)} & \multicolumn{3}{c}{Web Search (with cutoff)} \\
\cmidrule(lr){3-5}\cmidrule(lr){6-8}
Metric & Baseline & Value & $\Delta$ & $p$-value & Value & $\Delta$ & $p$-value \\
\midrule
\multicolumn{8}{l}{\llmicon{icons/openai_logo.png} \textit{GPT-4o}} \\
\addlinespace[2pt]
\quad Binary & $0.536 \pm 0.019$ & $0.779 \pm 0.016$ & $+0.243$ & $< 0.001$~\textbf{***} & $0.564 \pm 0.019$ & $+0.029$ & $0.192$ \\
\quad MCQ & $0.542 \pm 0.024$ & $0.873 \pm 0.016$ & $+0.331$ & $< 0.001$~\textbf{***} & $0.589 \pm 0.024$ & $+0.046$ & $0.062$ \\
\quad FRQ score (0--10) & $3.278 \pm 0.047$ & $4.249 \pm 0.081$ & $+0.971$ & $< 0.001$~\textbf{***} & $3.715 \pm 0.047$ & $+0.432$ & $< 0.001$~\textbf{***} \\
\quad Date score (0--1) & $0.183 \pm 0.014$ & $0.621 \pm 0.025$ & $+0.438$ & $< 0.001$~\textbf{***} & $0.382 \pm 0.018$ & $+0.201$ & $< 0.001$~\textbf{***} \\
\quad Date exact match & $0.014 \pm 0.007$ & $0.471 \pm 0.029$ & $+0.457$ & $< 0.001$~\textbf{***} & $0.061 \pm 0.014$ & $+0.047$ & $< 0.001$~\textbf{***} \\
\quad Date month error & $34.465 \pm 2.227$ & $6.408 \pm 1.004$ & $-28.057$ & $< 0.001$~\textbf{***} & $20.655 \pm 1.751$ & $-15.215$ & $< 0.001$~\textbf{***} \\
\addlinespace[4pt]
\multicolumn{8}{l}{\llmicon{icons/openai_logo.png} \textit{GPT-5.4}} \\
\addlinespace[2pt]
\quad Binary & $0.499 \pm 0.019$ & $0.599 \pm 0.019$ & $+0.100$ & $< 0.001$~\textbf{***} & $0.593 \pm 0.019$ & $+0.094$ & $< 0.001$~\textbf{***} \\
\quad MCQ & $0.841 \pm 0.018$ & $0.956 \pm 0.010$ & $+0.115$ & $< 0.001$~\textbf{***} & $0.897 \pm 0.015$ & $+0.055$ & $0.005$~\textbf{**} \\
\quad FRQ score (0--10) & $5.052 \pm 0.040$ & $5.683 \pm 0.051$ & $+0.631$ & $< 0.001$~\textbf{***} & $5.867 \pm 0.048$ & $+0.809$ & $< 0.001$~\textbf{***} \\
\quad Date score (0--1) & $0.275 \pm 0.018$ & $0.748 \pm 0.021$ & $+0.472$ & $< 0.001$~\textbf{***} & $0.643 \pm 0.021$ & $+0.369$ & $< 0.001$~\textbf{***} \\
\quad Date exact match & $0.027 \pm 0.009$ & $0.498 \pm 0.029$ & $+0.471$ & $< 0.001$~\textbf{***} & $0.341 \pm 0.028$ & $+0.314$ & $< 0.001$~\textbf{***} \\
\quad Date month error & $38.757 \pm 4.377$ & $7.533 \pm 2.592$ & $-31.224$ & $< 0.001$~\textbf{***} & $10.493 \pm 1.116$ & $-28.926$ & $< 0.001$~\textbf{***} \\
\addlinespace[4pt]
\multicolumn{8}{l}{\llmicon{icons/deepseek_logo.png} \textit{DeepSeek R1}} \\
\addlinespace[2pt]
\quad Binary & $0.476 \pm 0.019$ & $0.584 \pm 0.019$ & $+0.108$ & $< 0.001$~\textbf{***} & $0.518 \pm 0.019$ & $+0.044$ & $0.040$~\textbf{*} \\
\quad MCQ & $0.599 \pm 0.025$ & $0.771 \pm 0.021$ & $+0.172$ & $< 0.001$~\textbf{***} & $0.676 \pm 0.024$ & $+0.080$ & $0.004$~\textbf{**} \\
\quad FRQ score (0--10) & $4.232 \pm 0.042$ & $4.691 \pm 0.049$ & $+0.459$ & $< 0.001$~\textbf{***} & $4.333 \pm 0.043$ & $+0.097$ & $0.050$~\textbf{*} \\
\quad Date score (0--1) & $0.289 \pm 0.017$ & $0.544 \pm 0.020$ & $+0.255$ & $< 0.001$~\textbf{***} & $0.470 \pm 0.020$ & $+0.179$ & $< 0.001$~\textbf{***} \\
\quad Date exact match & $0.017 \pm 0.008$ & $0.214 \pm 0.024$ & $+0.197$ & $< 0.001$~\textbf{***} & $0.130 \pm 0.020$ & $+0.113$ & $< 0.001$~\textbf{***} \\
\quad Date month error & $20.372 \pm 1.913$ & $11.433 \pm 1.506$ & $-8.939$ & $< 0.001$~\textbf{***} & $15.703 \pm 1.975$ & $-4.557$ & $0.003$~\textbf{**} \\
\addlinespace[4pt]
\multicolumn{8}{l}{\llmicon{icons/claude.png} \textit{Claude Sonnet}} \\
\addlinespace[2pt]
\quad Binary & $0.534 \pm 0.020$ & $0.687 \pm 0.018$ & $+0.153$ & $< 0.001$~\textbf{***} & $0.620 \pm 0.019$ & $+0.086$ & $< 0.001$~\textbf{***} \\
\quad MCQ & $0.738 \pm 0.022$ & $0.810 \pm 0.020$ & $+0.072$ & $0.010$~\textbf{*} & $0.750 \pm 0.022$ & $+0.008$ & $0.836$ \\
\quad FRQ score (0--10) & $3.951 \pm 0.050$ & $4.540 \pm 0.056$ & $+0.589$ & $< 0.001$~\textbf{***} & $4.292 \pm 0.051$ & $+0.333$ & $< 0.001$~\textbf{***} \\
\quad Date score (0--1) & $0.258 \pm 0.017$ & $0.484 \pm 0.024$ & $+0.226$ & $< 0.001$~\textbf{***} & $0.377 \pm 0.024$ & $+0.119$ & $< 0.001$~\textbf{***} \\
\quad Date exact match & $0.029 \pm 0.010$ & $0.242 \pm 0.026$ & $+0.213$ & $< 0.001$~\textbf{***} & $0.175 \pm 0.023$ & $+0.146$ & $< 0.001$~\textbf{***} \\
\quad Date month error & $25.539 \pm 1.948$ & $15.479 \pm 6.201$ & $-10.060$ & $0.102$ & $14.453 \pm 1.586$ & $-14.194$ & $0.002$~\textbf{**} \\
\bottomrule
\end{tabular}
\caption{Web-search augmentation across models on matched 500-question subsets. Values are means $\pm$ SE. $\Delta$ is relative to the Baseline. Significance: *** $p<0.001$, ** $p<0.01$, * $p<0.05$, n.s.\ not significant.}
\label{tab:combined_web_search_cutoff}
\end{table}

\pagebreak
\begin{table}[h]
\centering
\setlength{\tabcolsep}{4pt}
\resizebox{\linewidth}{!}{
\begin{tabular}{@{}l cccccc@{}}
\toprule
Metric & \multicolumn{2}{c}{\llmicon{icons/openai_logo.png} \textit{GPT-5.4} (Aug 2025)} & \multicolumn{2}{c}{\llmicon{icons/deepseek_logo.png} \textit{DeepSeek R1} (Jul 2024)} & \multicolumn{2}{c}{\llmicon{icons/claude.png} \textit{Claude Sonnet} (Jan 2025)} \\
\cmidrule(lr){2-3} \cmidrule(lr){4-5} \cmidrule(lr){6-7}
& \textit{Pre} ($n=377$) & \textit{Post} ($n=123$) & \textit{Pre} ($n=119$) & \textit{Post} ($n=381$) & \textit{Pre} ($n=214$) & \textit{Post} ($n=286$) \\
\midrule
Binary & $0.49$ / $0.61^{***}$ / $0.62^{***}$ & $0.53$ / $0.55$ / $0.49$ & $0.47$ / $0.61^{**}$ / $0.59^{*}$ & $0.48$ / $0.58^{***}$ / $0.49$ & $0.49$ / $0.70^{***}$ / $0.68^{***}$ & $0.57$ / $0.67^{**}$ / $0.58$ \\
MCQ & $0.85$ / $0.97^{***}$ / $0.94^{***}$ & $0.82$ / $0.92^{**}$ / $0.77$ & $0.67$ / $0.82^{*}$ / $0.78^{*}$ & $0.57$ / $0.76^{***}$ / $0.64^{*}$ & $0.77$ / $0.82$ / $0.81$ & $0.72$ / $0.81^{*}$ / $0.70$ \\
FRQ (0--10) & $5.06$ / $5.76^{***}$ / $6.06^{***}$ & $5.06$ / $5.43^{***}$ / $5.29^{*}$ & $4.21$ / $4.66^{***}$ / $4.50^{**}$ & $4.24$ / $4.70^{***}$ / $4.29$ & $3.96$ / $4.74^{***}$ / $4.61^{***}$ & $3.95$ / $4.40^{***}$ / $4.07$ \\
Date (0--1) & $0.28$ / $0.75^{***}$ / $0.72^{***}$ & $0.24$ / $0.75^{***}$ / $0.31$ & $0.16$ / $0.52^{***}$ / $0.55^{***}$ & $0.35$ / $0.56^{***}$ / $0.43^{***}$ & $0.25$ / $0.57^{***}$ / $0.49^{***}$ & $0.27$ / $0.39^{***}$ / $0.28$ \\
\bottomrule
\end{tabular}
}
\caption{Per-metric scores split by each model's training knowledge cutoff. Pre-cutoff: publication date $\leq$ training cutoff; post-cutoff: publication date $>$ training cutoff (all 500 questions classified using benchmark metadata). Each cell shows Baseline / WS / WS+Cutoff (means). Significance of WS and WS+Cutoff vs.\ Baseline: $^{***}p<0.001$, $^{**}p<0.01$, $^{*}p<0.05$.}
\label{tab:cutoff_split}
\end{table}

\begin{table}[h]
\centering
\setlength{\tabcolsep}{4pt}
\resizebox{\linewidth}{!}{
\begin{tabular}{@{}ll cccccc@{}}
\toprule
Task & Metric & \multicolumn{2}{c}{\llmicon{icons/openai_logo.png} \textit{GPT-5.4} (Aug 2025)} & \multicolumn{2}{c}{\llmicon{icons/deepseek_logo.png} \textit{DeepSeek R1} (Jul 2024)} & \multicolumn{2}{c}{\llmicon{icons/claude.png} \textit{Claude S4.5} (Jan 2025)} \\
\cmidrule(lr){3-4} \cmidrule(lr){5-6} \cmidrule(lr){7-8}
& & \textit{Pre} ($n=377$) & \textit{Post} ($n=123$) & \textit{Pre} ($n=119$) & \textit{Post} ($n=381$) & \textit{Pre} ($n=214$) & \textit{Post} ($n=286$) \\
\midrule
\multirow{3}{*}{Binary} & ECE$\downarrow$ & $0.109$ / $0.027$ / $0.021$ & $0.152$ / $0.039$ / $0.060$ & $0.215$ / $0.118$ / $0.081$ & $0.221$ / $0.061$ / $0.105$ & $0.401$ / $0.147$ / $0.190$ & $0.448$ / $0.127$ / $0.426$ \\
 & Brier$\downarrow$ & $0.238$ / $0.025$ / $0.059$ & $0.242$ / $0.016$ / $0.137$ & $0.186$ / $0.135$ / $0.115$ & $0.146$ / $0.093$ / $0.164$ & $0.370$ / $0.138$ / $0.169$ & $0.391$ / $0.125$ / $0.356$ \\
 & Gap & $+0.067$ / $-0.021$ / $+0.013$ & $+0.071$ / $-0.029$ / $-0.009$ & $+0.213$ / $+0.115$ / $+0.075$ & $+0.045$ / $-0.000$ / $+0.090$ & $+0.401$ / $+0.104$ / $+0.152$ & $+0.448$ / $+0.076$ / $+0.426$ \\
\addlinespace[4pt]
\multirow{3}{*}{Binary (pert.)} & ECE$\downarrow$ & $0.338$ / $0.505$ / $0.468$ & $0.211$ / $0.527$ / $0.461$ & $0.497$ / $0.511$ / $0.487$ & $0.564$ / $0.546$ / $0.533$ & $0.187$ / $0.204$ / $0.274$ & $0.193$ / $0.219$ / $0.164$ \\
 & Brier$\downarrow$ & $0.350$ / $0.487$ / $0.486$ & $0.296$ / $0.518$ / $0.450$ & $0.397$ / $0.467$ / $0.465$ & $0.442$ / $0.484$ / $0.450$ & $0.266$ / $0.273$ / $0.282$ & $0.213$ / $0.294$ / $0.209$ \\
 & Gap & $+0.338$ / $+0.505$ / $+0.468$ & $+0.211$ / $+0.511$ / $+0.461$ & $-0.039$ / $+0.206$ / $+0.294$ & $+0.141$ / $+0.292$ / $+0.199$ & $+0.053$ / $+0.130$ / $+0.102$ & $-0.119$ / $+0.115$ / $-0.016$ \\
\addlinespace[4pt]
\multirow{3}{*}{MCQ} & ECE$\downarrow$ & $0.040$ / $0.029$ / $0.040$ & $0.041$ / $0.020$ / $0.082$ & $0.252$ / $0.120$ / $0.159$ & $0.332$ / $0.169$ / $0.309$ & $0.074$ / $0.077$ / $0.068$ & $0.040$ / $0.088$ / $0.052$ \\
 & Brier$\downarrow$ & $0.114$ / $0.028$ / $0.051$ & $0.133$ / $0.059$ / $0.169$ & $0.279$ / $0.155$ / $0.191$ & $0.353$ / $0.193$ / $0.326$ & $0.166$ / $0.063$ / $0.075$ & $0.202$ / $0.098$ / $0.185$ \\
 & Gap & $+0.003$ / $-0.014$ / $-0.015$ & $-0.005$ / $+0.007$ / $+0.049$ & $+0.246$ / $+0.110$ / $+0.159$ & $+0.332$ / $+0.169$ / $+0.309$ & $-0.051$ / $-0.029$ / $-0.041$ & $-0.008$ / $-0.003$ / $+0.030$ \\
\addlinespace[4pt]
\multirow{3}{*}{Date} & ECE$\downarrow$ & $0.321$ / $0.181$ / $0.166$ & $0.339$ / $0.272$ / $0.371$ & $0.497$ / $0.264$ / $0.236$ & $0.278$ / $0.252$ / $0.260$ & $0.311$ / $0.158$ / $0.172$ & $0.251$ / $0.317$ / $0.237$ \\
 & Brier$\downarrow$ & $0.190$ / $0.107$ / $0.106$ & $0.188$ / $0.161$ / $0.202$ & $0.280$ / $0.159$ / $0.147$ & $0.156$ / $0.156$ / $0.174$ & $0.171$ / $0.107$ / $0.104$ & $0.131$ / $0.187$ / $0.140$ \\
 & Gap & $+0.321$ / $+0.173$ / $+0.166$ & $+0.339$ / $+0.263$ / $+0.371$ & $+0.494$ / $+0.264$ / $+0.236$ & $+0.278$ / $+0.252$ / $+0.260$ & $+0.311$ / $+0.158$ / $+0.160$ & $+0.251$ / $+0.317$ / $+0.230$ \\
\bottomrule
\end{tabular}
}
\caption{Confidence calibration split by each model's training cutoff. Pre-cutoff: publication date $\leq$ training cutoff; post-cutoff: publication date $>$ training cutoff (all 500 questions classified via benchmark metadata). Each cell shows No-WS / WS / WS+Cutoff values. ECE = Expected Calibration Error; Brier = Brier score; Gap $= \bar{c} - \bar{a}$ (positive = overconfident). $\downarrow$ lower is better.}
\label{tab:calib_cutoff_split}
\end{table}

\begin{benchbox}[GPT-5.4 + Full Web Search Overlooks Specific Size Constraint]
\textbf{Perturbed Question:}\\
By November 2024, will a method show that metal-metal interactions dominate oxide support effects \textbf{\textcolor{red}{for particle sizes below 2 nm}}?

\medskip

\textbf{Perturbation Detail:}\\
Added a specific unmet constraint regarding \textbf{\textcolor{red}{particle size uniformity below 2 nm}}.

\vspace{0.5em}
\hrule
\vspace{0.5em}

\textbf{Source Abstract:}\\
\small{The metal-support interaction is one of the most important pillars in heterogeneous catalysis, but developing a fundamental theory has been challenging because of the intricate interfaces. Based on experimental data, interpretable machine learning, theoretical derivation, and first-principles simulations, we established a general theory of metal-oxide interactions grounded in metal-metal and metal-oxygen interactions. The theory applies to metal nanoparticles and atoms on oxide supports and oxide films on metal supports. We found that for late-transition metal catalysts, metal-metal interactions dominated the oxide support effects and suboxide encapsulation over metal nanoparticles. A principle of strong metal-metal interactions for encapsulation occurrence is formulated and substantiated by extensive experiments including 10 metals and 16 oxides. The valuable insights revealed on (strong) metal-support interaction advance the interfacial design of supported metal catalysts. \cite{wang2024nature}}

\vspace{0.5em}
\hrule
\vspace{0.5em}

\textbf{Model Response (GPT-5.4 + Web Search):}\\
\textit{``Yes. By November 2024, a Science paper reported a general theory/method for metal–oxide interactions and explicitly concluded that, for late-transition metal catalysts, metal–metal interactions dominate oxide support effects; \textbf{\textcolor{red}{this appears to satisfy the question’s threshold.}}''}

\medskip

\textbf{Error Analysis:}\\
The model commits a \textbf{False Positive} (Missed Constraint). It successfully matches the broad scientific themes of the text but completely ignores the artificially inserted \textbf{\textcolor{red}{$< 2$~nm}} requirement, incorrectly confirming the threshold was met.
\end{benchbox}

\begin{benchbox}[GPT-5.4 + Web Search Fails via Speculative Extrapolation]
\textbf{Perturbed Question:}\\
Aider Polyglot evaluates the ability of code language models to perform diverse code-editing tasks based on natural language instructions while retaining their general code generation and problem-solving capabilities. It measures model performance using synthetic datasets of code-editing scenarios and assesses robustness through adaptation algorithms designed to prevent overfitting. By Jun 2025, do you think there will be a publicly reported AI system that achieves \textbf{\textcolor{red}{90\% or higher}} on Aider Polyglot?

\medskip

\textbf{Perturbation Detail:}\\
Inflated threshold = \textbf{\textcolor{red}{90\%}}

\vspace{0.5em}
\hrule
\vspace{0.5em}

\textbf{Model Response (GPT-5.4 Web Search Unconstrained):}\\
\textit{``Yes. Public aider leaderboard/results pages show top polyglot \textbf{\textcolor{red}{scores already reached about 85\%}} in late 2024/early 2025, and \textbf{\textcolor{red}{given the rapid improvement pace}} in code-editing models through 2025, \textbf{\textcolor{red}{it was more likely than not}} that at least one publicly reported system would cross 90\% by June 2025.''}

\medskip

\textbf{Error Analysis:}\\
The model commits a \textbf{False Positive} (Speculative Hallucination / Overconfident Extrapolation). Despite the rigid \textbf{\textcolor{red}{90\%}} threshold, the model grounds its ``Yes'' answer in data that only supports 85\%. It improperly substitutes a forecast (``more likely than not'') for factual satisfaction of the benchmark criteria.

\end{benchbox}


\subsection{FRQ Sub-dimension Score Analysis}

\begin{figure}[H]
    \centering
    \includegraphics[width=\textwidth]{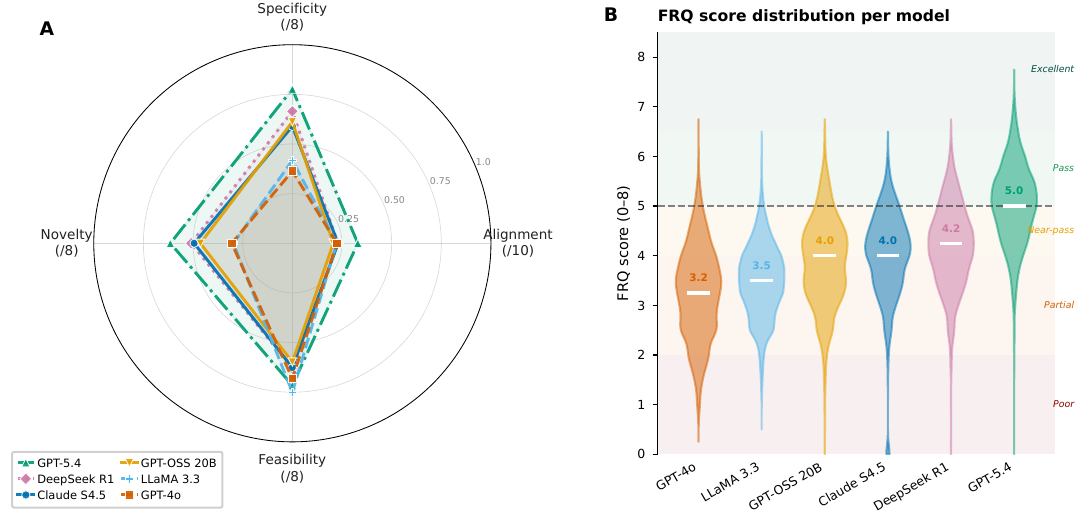}
\caption{A) Visualization of FRQ evaluation across four criteria on 6 LLMs B) FRQ score distribution per model.}
    \label{fig:frq_radar}

    \centering
    \includegraphics[width=\textwidth]{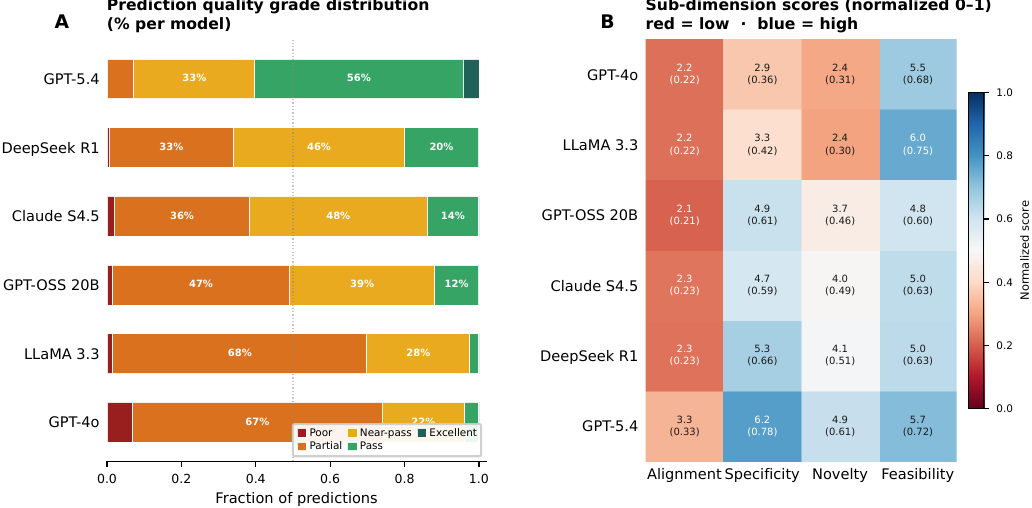}
\caption{A) Visualization of passing rates across six LLMs. B) Visualization of LLM performance across four FRQ criteria.}
    \label{fig:frq_area}
\end{figure}

\begin{table}[h]
\centering
\begin{adjustbox}{max width=\textwidth}
\small\renewcommand{\arraystretch}{1.18}
\begin{tabular}{@{}ll rrrrrr@{}}
\toprule
    \textbf{Model} & \textbf{Cutoff} & \textbf{Align.\,$\uparrow$} & \textbf{Spec.\,$\uparrow$} & \textbf{Nov.\,$\uparrow$} & \textbf{Spec.\,$-$\,Align.\ gap} & \textbf{Overall score} \\
\midrule
   \llmicon{icons/openai_logo.png} GPT-OSS & Jun 2024 & \cellcolor{cellworse}2.06 & \cellcolor{cellgood}4.89 & \cellcolor{cellneutral}3.70 & +2.8 & 3.86 \\
  \llmicon{icons/llama.png}  LLaMA 3.3 & Dec 2023 & \cellcolor{cellworse}2.23 & \cellcolor{cellneutral}3.33 & \cellcolor{cellworse}2.38 & +1.1 & 3.49 \\
 \llmicon{icons/openai_logo.png} GPT-4o & Oct 2023 & \cellcolor{cellworse}2.25 & \cellcolor{cellworse}2.91 & \cellcolor{cellworse}2.44 & +0.7 & 3.26 \\
  \llmicon{icons/deepseek_logo.png}  DeepSeek R1 & Jul 2024 & \cellcolor{cellworse}2.29 & \cellcolor{cellbetter}5.32 & \cellcolor{cellgood}4.07 & +3.0 & 4.18 \\
  \llmicon{icons/claude.png}  Claude S4.5 & Jan 2025 & \cellcolor{cellworse}2.31 & \cellcolor{cellgood}4.71 & \cellcolor{cellneutral}3.95 & +2.4 & 3.99 \\
  \llmicon{icons/openai_logo.png} GPT-5.4 & Aug 2025 & \cellcolor{cellneutral}3.30 & \cellcolor{cellbetter}6.21 & \cellcolor{cellgood}4.91 & +2.9 & 5.04 \\
\bottomrule
\end{tabular}
\end{adjustbox}
\caption{FRQ sub-dimension profile. \textbf{Alignment}: match with the actual paper method. \textbf{Specificity}: technical concreteness. \textbf{Novelty}: non-obvious insight. \textbf{Spec.\,$-$\,Align.\ gap}: positive values indicate models that write technically detailed responses but miss the specific paper approach --- a signature of plausible-sounding hallucination. Models sorted from lowest to highest alignment.}
\label{tab:frq_subdims}
\begin{minipage}{\textwidth}
\smallskip\footnotesize\textit{All sub-dimensions scored 0--10 by the LLM rubric judge. Colour shading on Alignment, Specificity, and Novelty uses the same scale: red $<$ 3, orange 3--4, yellow 4--5, light green 5--6, medium green 6--6.5, dark green $\geq$ 6.5.}
\end{minipage}
\end{table}

\subsection{Results by Research Area}
\label{sec:subcategories}
\begin{table}[H]
\centering
\begin{adjustbox}{max width=\textwidth}
\small\renewcommand{\arraystretch}{1.20}
\begin{tabular}{@{}lr rrrrrr r@{}}
\toprule
    \textbf{Research area} & $n$ & \llmicon{icons/openai_logo.png} {\textbf{GPT-5.4}} &\llmicon{icons/claude.png} {\textbf{Claude S4.5}} & \llmicon{icons/deepseek_logo.png} {\textbf{DeepSeek R1}} & \llmicon{icons/llama.png} {\textbf{LLaMA 3.3}} & \llmicon{icons/openai_logo.png} {\textbf{GPT-OSS}} & \llmicon{icons/openai_logo.png} {\textbf{GPT-4o}} & \textbf{Mean} \\
\midrule
    Biology & 1058 & \cellcolor{cellbest}\textbf{0.819} & \cellcolor{cellbetter}0.739 & \cellcolor{cellgood}0.652 & \cellcolor{cellneutral}0.522 & \cellcolor{cellneutral}0.547 & \cellcolor{cellneutral}0.570 & \cellcolor{cellgood}0.641 \\
    AI & 964 & \cellcolor{cellbest}\textbf{0.817} & \cellcolor{cellbetter}0.765 & \cellcolor{cellneutral}0.520 & \cellcolor{cellbad}0.332 & \cellcolor{cellbad}0.366 & \cellcolor{cellworse}0.456 & \cellcolor{cellneutral}0.543 \\
    Medicine & 646 & \cellcolor{cellbetter}\textbf{0.769} & \cellcolor{cellgood}0.659 & \cellcolor{cellneutral}0.567 & \cellcolor{cellworse}0.443 & \cellcolor{cellworse}0.472 & \cellcolor{cellneutral}0.522 & \cellcolor{cellneutral}0.572 \\
    Neurosci. & 351 & \cellcolor{cellbest}\textbf{0.829} & \cellcolor{cellbetter}0.727 & \cellcolor{cellgood}0.666 & \cellcolor{cellneutral}0.532 & \cellcolor{cellneutral}0.564 & \cellcolor{cellgood}0.629 & \cellcolor{cellgood}0.658 \\
    Mat.\ Sci. & 356 & \cellcolor{cellbest}\textbf{0.831} & \cellcolor{cellgood}0.680 & \cellcolor{cellneutral}0.567 & \cellcolor{cellbad}0.322 & \cellcolor{cellbad}0.374 & \cellcolor{cellworse}0.480 & \cellcolor{cellneutral}0.542 \\
    Physics & 325 & \cellcolor{cellbest}\textbf{0.840} & \cellcolor{cellbetter}0.702 & \cellcolor{cellgood}0.669 & \cellcolor{cellworse}0.425 & \cellcolor{cellneutral}0.526 & \cellcolor{cellneutral}0.563 & \cellcolor{cellgood}0.621 \\
    Env.\ Sci. & 185 & \cellcolor{cellbest}\textbf{0.897} & \cellcolor{cellbest}0.818 & \cellcolor{cellgood}0.621 & \cellcolor{cellneutral}0.525 & \cellcolor{cellneutral}0.546 & \cellcolor{cellneutral}0.568 & \cellcolor{cellgood}0.662 \\
    Chemistry & 185 & \cellcolor{cellbest}\textbf{0.822} & \cellcolor{cellgood}0.620 & \cellcolor{cellworse}0.441 & \cellcolor{cellbad}0.330 & \cellcolor{cellbad}0.346 & \cellcolor{cellworse}0.432 & \cellcolor{cellworse}0.498 \\
    Other & 58 & \cellcolor{cellbest}\textbf{0.879} & \cellcolor{cellbest}0.862 & \cellcolor{cellbetter}0.780 & \cellcolor{cellgood}0.632 & \cellcolor{cellgood}0.672 & \cellcolor{cellbetter}0.793 & \cellcolor{cellbetter}0.770 \\
\midrule
    \textit{Mean} &  & \textbf{0.834} & 0.730 & 0.609 & 0.451 & 0.490 & 0.557 & 0.612 \\
\bottomrule
\end{tabular}
\end{adjustbox}
\caption{MCQ accuracy by research area. Chance level = 0.25 (4-choice). $n$: MCQ questions per area. \textbf{Bold}: best model per row. Colour scale: \colorbox{cellbad}{\strut\,$<$0.40\,} \colorbox{cellworse}{\strut\,0.40--0.50\,} \colorbox{cellneutral}{\strut\,0.50--0.60\,} \colorbox{cellgood}{\strut\,0.60--0.70\,} \colorbox{cellbetter}{\strut\,0.70--0.80\,} \colorbox{cellbest}{\strut\,$\geq$0.80\,}.}
\label{tab:area_mcq}

\centering
\begin{adjustbox}{max width=\textwidth}
\small\renewcommand{\arraystretch}{1.20}
\begin{tabular}{@{}lr rrrrrr r@{}}
\toprule
    \textbf{Research area} & $n$ & {\llmicon{icons/openai_logo.png} \textbf{GPT-5.4}} & {\llmicon{icons/claude.png} \textbf{Claude S4.5}} & { \llmicon{icons/llama.png} \textbf{LLaMA 3.3}} & {\llmicon{icons/deepseek_logo.png} \textbf{DeepSeek R1}} & {\llmicon{icons/openai_logo.png} \textbf{GPT-OSS}} & {\llmicon{icons/openai_logo.png} \textbf{GPT-4o}} & \textbf{Mean} \\
\midrule
    Biology & 677 & \cellcolor{cellworse}0.194 & \cellcolor{cellworse}0.185 & \cellcolor{cellbest}\textbf{0.502} & \cellcolor{cellneutral}0.300 & \cellcolor{cellneutral}0.271 & \cellcolor{cellbad}0.135 & \cellcolor{cellneutral}0.264 \\
    AI & 571 & \cellcolor{cellbest}0.473 & \cellcolor{cellbest}0.464 & \cellcolor{cellbest}\textbf{0.554} & \cellcolor{cellbetter}0.404 & \cellcolor{cellbest}0.447 & \cellcolor{cellbetter}0.421 & \cellcolor{cellbest}0.461 \\
    Medicine & 336 & \cellcolor{cellworse}0.228 & \cellcolor{cellworse}0.218 & \cellcolor{cellbest}\textbf{0.506} & \cellcolor{cellgood}0.318 & \cellcolor{cellgood}0.303 & \cellcolor{cellbad}0.136 & \cellcolor{cellneutral}0.285 \\
    Neurosci. & 232 & \cellcolor{cellbad}0.159 & \cellcolor{cellworse}0.199 & \cellcolor{cellbest}\textbf{0.473} & \cellcolor{cellneutral}0.266 & \cellcolor{cellgood}0.303 & \cellcolor{cellbad}0.108 & \cellcolor{cellneutral}0.251 \\
    Mat.\ Sci. & 319 & \cellcolor{cellbad}0.149 & \cellcolor{cellbad}0.130 & \cellcolor{cellbest}\textbf{0.453} & \cellcolor{cellworse}0.188 & \cellcolor{cellworse}0.232 & \cellcolor{cellbad}0.085 & \cellcolor{cellworse}0.206 \\
    Physics & 302 & \cellcolor{cellbad}0.143 & \cellcolor{cellbad}0.157 & \cellcolor{cellbest}\textbf{0.480} & \cellcolor{cellneutral}0.252 & \cellcolor{cellworse}0.213 & \cellcolor{cellbad}0.089 & \cellcolor{cellworse}0.222 \\
    Env.\ Sci. & 128 & \cellcolor{cellneutral}0.246 & \cellcolor{cellbad}0.174 & \cellcolor{cellbest}\textbf{0.490} & \cellcolor{cellworse}0.208 & \cellcolor{cellneutral}0.281 & \cellcolor{cellbad}0.114 & \cellcolor{cellneutral}0.252 \\
    Chemistry & 141 & \cellcolor{cellbad}0.117 & \cellcolor{cellbad}0.128 & \cellcolor{cellbest}\textbf{0.447} & \cellcolor{cellbad}0.135 & \cellcolor{cellworse}0.209 & \cellcolor{cellbad}0.085 & \cellcolor{cellworse}0.187 \\
    Other & 49 & \cellcolor{cellworse}0.229 & \cellcolor{cellneutral}0.284 & \cellcolor{cellbest}\textbf{0.517} & \cellcolor{cellworse}0.201 & \cellcolor{cellneutral}0.260 & \cellcolor{cellbad}0.127 & \cellcolor{cellneutral}0.270 \\
\midrule
    \textit{Mean} &  & 0.215 & 0.215 & \textbf{0.491} & 0.252 & 0.280 & 0.144 & 0.266 \\
\bottomrule
\end{tabular}
\end{adjustbox}
\caption{Date prediction score by research area. Score uses exponential decay (1.0\,=\,exact month). $n$: date-prediction questions per area. \textbf{Bold}: best model per row. Colours reflect the observed score range across this table (5th--95th percentile).}
\label{tab:area_date}

\centering

\label{tab:area_binary}
\begin{adjustbox}{max width=\textwidth}
\small\renewcommand{\arraystretch}{1.20}
\begin{tabular}{@{}lr rrrrrr r@{}}
\toprule
    \textbf{Research area} & $n$ & {\llmicon{icons/openai_logo.png} \textbf{GPT-5.4}} & {\llmicon{icons/claude.png} \textbf{Claude S4.5}} & { \llmicon{icons/llama.png} \textbf{LLaMA 3.3}} & {\llmicon{icons/deepseek_logo.png} \textbf{DeepSeek R1}} & {\llmicon{icons/openai_logo.png} \textbf{GPT-OSS}} & {\llmicon{icons/openai_logo.png} \textbf{GPT-4o}} & \textbf{Mean} \\
\midrule
    Biology & 1576 & \cellcolor{cellgood}0.523 & \cellcolor{cellneutral}0.484 & \cellcolor{cellworse}0.455 & \cellcolor{cellneutral}0.499 & \cellcolor{cellgood}\textbf{0.541} & \cellcolor{cellgood}0.526 & \cellcolor{cellneutral}0.505 \\
    AI & 1594 & \cellcolor{cellworse}0.464 & \cellcolor{cellbest}\textbf{0.602} & \cellcolor{cellbad}0.377 & \cellcolor{cellworse}0.457 & \cellcolor{cellgood}0.548 & \cellcolor{cellgood}0.543 & \cellcolor{cellneutral}0.498 \\
    Medicine & 847 & \cellcolor{cellneutral}0.516 & \cellcolor{cellneutral}0.517 & \cellcolor{cellbad}0.432 & \cellcolor{cellneutral}0.511 & \cellcolor{cellgood}0.536 & \cellcolor{cellbetter}\textbf{0.567} & \cellcolor{cellneutral}0.513 \\
    Neurosci. & 517 & \cellcolor{cellgood}0.538 & \cellcolor{cellworse}0.473 & \cellcolor{cellneutral}0.510 & \cellcolor{cellgood}\textbf{0.552} & \cellcolor{cellgood}0.520 & \cellcolor{cellgood}0.524 & \cellcolor{cellneutral}0.520 \\
    Mat.\ Sci. & 637 & \cellcolor{cellworse}0.479 & \cellcolor{cellworse}0.453 & \cellcolor{cellneutral}\textbf{0.500} & \cellcolor{cellbad}0.439 & \cellcolor{cellworse}0.473 & \cellcolor{cellworse}0.474 & \cellcolor{cellworse}0.469 \\
    Physics & 561 & \cellcolor{cellneutral}0.510 & \cellcolor{cellworse}0.455 & \cellcolor{cellgood}\textbf{0.538} & \cellcolor{cellworse}0.460 & \cellcolor{cellbad}0.439 & \cellcolor{cellbad}0.433 & \cellcolor{cellworse}0.472 \\
    Env.\ Sci. & 301 & \cellcolor{cellworse}0.475 & \cellcolor{cellneutral}0.502 & \cellcolor{cellworse}0.473 & \cellcolor{cellneutral}0.517 & \cellcolor{cellgood}0.525 & \cellcolor{cellgood}\textbf{0.548} & \cellcolor{cellneutral}0.507 \\
    Chemistry & 285 & \cellcolor{cellneutral}\textbf{0.502} & \cellcolor{cellneutral}0.482 & \cellcolor{cellneutral}0.495 & \cellcolor{cellbad}0.418 & \cellcolor{cellworse}0.449 & \cellcolor{cellbad}0.432 & \cellcolor{cellworse}0.463 \\
    Other & 93 & \cellcolor{cellneutral}0.495 & \cellcolor{cellneutral}0.495 & \cellcolor{cellbetter}\textbf{0.571} & \cellcolor{cellbad}0.432 & \cellcolor{cellworse}0.441 & \cellcolor{cellneutral}0.505 & \cellcolor{cellneutral}0.490 \\
\midrule
    \textit{Mean} &  & 0.500 & 0.496 & 0.484 & 0.476 & 0.497 & \textbf{0.506} & 0.493 \\
\bottomrule
\end{tabular}
\end{adjustbox}
\caption{Binary merged accuracy by research area. Binary merged\,=\,$\tfrac{1}{2}$(original acc.\,+\,perturbed acc.), correcting for directional response bias; chance\,=\,0.50. $n$: total binary question pairs (original\,+\,perturbed) per area. \textbf{Bold}: best model per row. Colour scale centred at chance (0.50): \colorbox{cellbad}{\strut\,$<$0.44\,} \colorbox{cellworse}{\strut\,0.44--0.48\,} \colorbox{cellneutral}{\strut\,0.48--0.52\,} \colorbox{cellgood}{\strut\,0.52--0.56\,} \colorbox{cellbetter}{\strut\,0.56--0.60\,} \colorbox{cellbest}{\strut\,$>$0.60\,}.}
\end{table}

\begin{table}[H]
\centering
\caption{Mean FRQ score (0--10) by research area and model. $n$: number of FRQ instances in the area (max across models). Colour: red $<$ 3, orange 3--4, yellow 4--5, light green 5--6, medium green 6--7, dark green $\geq$ 7.}
\label{tab:frq_by_area}
\begin{adjustbox}{max width=\textwidth}
\small\renewcommand{\arraystretch}{1.18}
\begin{tabular}{@{}lrrrrrrrr@{}}
\toprule
    \textbf{Area} & $n$ & \textbf{\llmicon{icons/openai_logo.png} GPT-5.4} & \textbf{\llmicon{icons/claude.png} Claude S4.5} & \textbf{\llmicon{icons/deepseek_logo.png} DeepSeek R1} & \textbf{\llmicon{icons/llama.png} LLaMA 3.3} & \textbf{\llmicon{icons/openai_logo.png} GPT-OSS} & \textbf{\llmicon{icons/openai_logo.png} GPT-4o} & \textbf{Mean} \\
\midrule
    Biology & 1023 & \cellcolor{cellgood}5.13 & \cellcolor{cellworse}3.99 & \cellcolor{cellneutral}4.27 & \cellcolor{cellworse}3.60 & \cellcolor{cellworse}3.98 & \cellcolor{cellworse}3.32 & \cellcolor{cellneutral}4.05 \\
    AI & 991 & \cellcolor{cellgood}5.06 & \cellcolor{cellneutral}4.22 & \cellcolor{cellneutral}4.25 & \cellcolor{cellworse}3.38 & \cellcolor{cellworse}3.89 & \cellcolor{cellworse}3.43 & \cellcolor{cellneutral}4.04 \\
    Medicine & 660 & \cellcolor{cellgood}5.08 & \cellcolor{cellworse}3.96 & \cellcolor{cellneutral}4.19 & \cellcolor{cellworse}3.60 & \cellcolor{cellneutral}4.01 & \cellcolor{cellworse}3.30 & \cellcolor{cellneutral}4.02 \\
    Neurosci. & 343 & \cellcolor{cellgood}5.27 & \cellcolor{cellneutral}4.11 & \cellcolor{cellneutral}4.30 & \cellcolor{cellworse}3.59 & \cellcolor{cellneutral}4.00 & \cellcolor{cellworse}3.41 & \cellcolor{cellneutral}4.11 \\
    Mat.\ Sci. & 353 & \cellcolor{cellneutral}4.82 & \cellcolor{cellworse}3.82 & \cellcolor{cellneutral}4.08 & \cellcolor{cellworse}3.42 & \cellcolor{cellworse}3.61 & \cellcolor{cellworse}3.03 & \cellcolor{cellworse}3.80 \\
    Physics & 311 & \cellcolor{cellneutral}4.85 & \cellcolor{cellworse}3.75 & \cellcolor{cellneutral}4.02 & \cellcolor{cellworse}3.24 & \cellcolor{cellworse}3.60 & \cellcolor{cellbad}3.00 & \cellcolor{cellworse}3.74 \\
    Env.\ Sci. & 205 & \cellcolor{cellgood}5.13 & \cellcolor{cellneutral}4.02 & \cellcolor{cellneutral}4.01 & \cellcolor{cellworse}3.65 & \cellcolor{cellworse}3.60 & \cellcolor{cellworse}3.09 & \cellcolor{cellworse}3.92 \\
    Chemistry & 193 & \cellcolor{cellneutral}4.49 & \cellcolor{cellworse}3.48 & \cellcolor{cellworse}3.80 & \cellcolor{cellworse}3.23 & \cellcolor{cellworse}3.45 & \cellcolor{cellbad}2.81 & \cellcolor{cellworse}3.54 \\
    Other & 55 & \cellcolor{cellneutral}4.88 & \cellcolor{cellworse}3.90 & \cellcolor{cellworse}3.83 & \cellcolor{cellworse}3.45 & \cellcolor{cellneutral}4.02 & \cellcolor{cellworse}3.12 & \cellcolor{cellworse}3.87 \\
\midrule
    \textit{Overall} &  & \textit{5.04} & \textit{3.99} & \textit{4.18} & \textit{3.49} & \textit{3.86} & \textit{3.26} & \textit{3.97} \\  
\bottomrule
\end{tabular}
\end{adjustbox}
\begin{minipage}{\textwidth}
\smallskip\footnotesize\textit{Chemistry and Physics consistently show lower FRQ alignment, reflecting higher domain specificity and fewer overlapping concepts with general pretraining data.}
\end{minipage}
\end{table}

\setlength{\tabcolsep}{3pt} 
\renewcommand{\arraystretch}{0.95}

\begin{longtable}{@{}l r r r r r r r r @{}}
\caption{Sub-domain predictability — Binary prediction (pooled: original GT=yes + perturbed GT=no) (values in \%). Chance level = 50\%. Cells shaded \colorbox{green!25}{\,} above / \colorbox{red!25}{\,} below chance. Cross-model \textit{mean} shown; best individual model per row in \textbf{bold}. Top 3 (most predictable) and bottom 3 (least predictable) sub-domains per area, ranked by cross-model mean. $n$: mean sample size across models.} \label{tab:subdomain_binary} \\
\toprule
\textbf{Sub-domain} & $\boldsymbol{n}$ & \textbf{Mean} & \rotatebox{55}{\textbf{GPT-5.4}} & \rotatebox{55}{\textbf{Claude S4.5}} & \rotatebox{55}{\textbf{DeepSeek R1}} & \rotatebox{55}{\textbf{GPT-OSS}} & \rotatebox{55}{\textbf{GPT-4o}} & \rotatebox{55}{\textbf{LLaMA 3.3}} \\
\midrule
\endfirsthead
\multicolumn{9}{l}{\small\tablename~\thetable{} \textit{(continued)}} \\
\toprule
\textbf{Sub-domain} & $\boldsymbol{n}$ & \textbf{Mean} & \rotatebox{55}{\textbf{GPT-5.4}} & \rotatebox{55}{\textbf{Claude S4.5}} & \rotatebox{55}{\textbf{DeepSeek R1}} & \rotatebox{55}{\textbf{GPT-OSS}} & \rotatebox{55}{\textbf{GPT-4o}} & \rotatebox{55}{\textbf{LLaMA 3.3}} \\
\midrule
\endhead
\midrule
\multicolumn{9}{r}{\small\textit{Continued on next page}} \\
\endfoot
\bottomrule
\endlastfoot
\multicolumn{9}{@{}l@{}}{\cellcolor{gray!18}\quad\textbf{Other}} \\*
Astrophysics & 23 & \cellcolor{green!5} 51 & \cellcolor{green!5} 50 & \cellcolor{green!8} 60 & \cellcolor{red!12} 36 & \cellcolor{red!5} 47 & \cellcolor{green!5} 50 & \cellcolor{green!11} \textbf{63} \\*
Cosmology & 14 & \cellcolor{red!5} 45 & \cellcolor{red!8} 41 & \cellcolor{green!5} \textbf{50} & \cellcolor{red!10} 38 & \cellcolor{red!8} 41 & \cellcolor{green!5} \textbf{50} & \cellcolor{green!5} \textbf{50} \\*
\cmidrule{1-9}
Cosmology & 14 & \cellcolor{red!5} 45 & \cellcolor{red!8} 41 & \cellcolor{green!5} \textbf{50} & \cellcolor{red!10} 38 & \cellcolor{red!8} 41 & \cellcolor{green!5} \textbf{50} & \cellcolor{green!5} \textbf{50} \\*
Astrophysics & 23 & \cellcolor{green!5} 51 & \cellcolor{green!5} 50 & \cellcolor{green!8} 60 & \cellcolor{red!12} 36 & \cellcolor{red!5} 47 & \cellcolor{green!5} 50 & \cellcolor{green!11} \textbf{63} \\
\noalign{\medskip}
\multicolumn{9}{@{}l@{}}{\cellcolor{gray!18}\quad\textbf{Environmental Science}} \\*
Atmospheric Science & 15 & \cellcolor{green!8} 59 & \cellcolor{green!5} 53 & \cellcolor{green!13} \textbf{65} & \cellcolor{green!11} 63 & \cellcolor{green!7} 58 & \cellcolor{green!5} 53 & \cellcolor{green!11} 63 \\*
Ecology & 46 & \cellcolor{green!5} 56 & \cellcolor{red!5} 46 & \cellcolor{green!5} 56 & \cellcolor{green!6} 57 & \cellcolor{green!12} \textbf{64} & \cellcolor{green!11} 63 & \cellcolor{red!5} 47 \\*
Oceanography & 20 & \cellcolor{green!5} 55 & \cellcolor{green!5} \textbf{56} & \cellcolor{green!5} 54 & \cellcolor{green!5} 52 & \cellcolor{green!5} \textbf{56} & \cellcolor{green!5} \textbf{56} & \cellcolor{green!5} \textbf{56} \\*
\cmidrule{1-9}
Public Health & 10 & \cellcolor{red!10} 39 & \cellcolor{red!10} 38 & \cellcolor{red!15} 33 & \cellcolor{red!17} 31 & \cellcolor{red!5} \textbf{46} & \cellcolor{red!10} 38 & \cellcolor{red!5} \textbf{46} \\*
Marine Biology & 18 & \cellcolor{red!5} 46 & \cellcolor{red!13} 35 & \cellcolor{green!6} \textbf{57} & \cellcolor{green!5} 52 & \cellcolor{red!5} 43 & \cellcolor{green!5} 52 & \cellcolor{red!9} 39 \\*
Glaciology & 11 & \cellcolor{red!5} 47 & \cellcolor{red!8} 40 & \cellcolor{red!8} 40 & \cellcolor{red!8} 40 & \cellcolor{green!8} \textbf{60} & \cellcolor{green!8} \textbf{60} & \cellcolor{red!8} 40 \\
\noalign{\medskip}
\multicolumn{9}{@{}l@{}}{\cellcolor{gray!18}\quad\textbf{Neuroscience}} \\*
Neurodegenerative Diseases & 13 & \cellcolor{green!10} 61 & \cellcolor{green!8} 60 & \cellcolor{green!10} 62 & \cellcolor{green!14} 67 & \cellcolor{green!8} 60 & \cellcolor{green!20} \textbf{73} & \cellcolor{red!5} 47 \\*
Cognitive Neuroscience & 29 & \cellcolor{green!9} 60 & \cellcolor{green!18} \textbf{70} & \cellcolor{green!6} 57 & \cellcolor{green!18} \textbf{70} & \cellcolor{green!8} 59 & \cellcolor{green!6} 57 & \cellcolor{red!5} 49 \\*
Psychiatry & 13 & \cellcolor{green!8} 59 & \cellcolor{green!5} 54 & \cellcolor{green!24} \textbf{77} & \cellcolor{green!5} 54 & \cellcolor{green!5} 54 & \cellcolor{green!17} 69 & \cellcolor{red!5} 46 \\*
\cmidrule{1-9}
Molecular Neuroscience & 24 & \cellcolor{red!6} 43 & \cellcolor{red!5} 47 & \cellcolor{red!17} 31 & \cellcolor{red!5} 44 & \cellcolor{red!10} 39 & \cellcolor{red!7} 42 & \cellcolor{green!5} \textbf{53} \\*
Cryo-Electron Microscopy & 16 & \cellcolor{red!6} 43 & \cellcolor{red!5} 47 & \cellcolor{red!10} 39 & \cellcolor{red!11} 37 & \cellcolor{green!5} \textbf{53} & \cellcolor{red!11} 37 & \cellcolor{red!5} 47 \\*
Neurophysiology & 20 & \cellcolor{red!5} 43 & \cellcolor{red!5} 48 & \cellcolor{red!8} 40 & \cellcolor{red!8} 40 & \cellcolor{red!8} 40 & \cellcolor{green!5} \textbf{52} & \cellcolor{red!8} 40 \\
\noalign{\medskip}
\multicolumn{9}{@{}l@{}}{\cellcolor{gray!18}\quad\textbf{Biology}} \\*
Stem Cell Biology & 18 & \cellcolor{green!11} 63 & \cellcolor{green!14} 67 & \cellcolor{green!19} 71 & \cellcolor{green!23} \textbf{76} & \cellcolor{green!10} 62 & \cellcolor{green!6} 57 & \cellcolor{red!6} 43 \\*
Computational Biology & 44 & \cellcolor{green!9} 60 & \cellcolor{green!13} \textbf{65} & \cellcolor{green!5} 56 & \cellcolor{green!13} 65 & \cellcolor{green!12} 64 & \cellcolor{green!6} 58 & \cellcolor{green!5} 55 \\*
Deep Learning & 19 & \cellcolor{green!8} 60 & \cellcolor{green!11} 63 & \cellcolor{green!5} 50 & \cellcolor{green!5} 52 & \cellcolor{green!14} \textbf{67} & \cellcolor{green!11} 63 & \cellcolor{green!11} 63 \\*
\cmidrule{1-9}
Single-Cell RNA Sequencing & 10 & \cellcolor{red!15} 32 & \cellcolor{red!20} 27 & \cellcolor{red!18} 30 & \cellcolor{red!12} \textbf{36} & \cellcolor{red!12} \textbf{36} & \cellcolor{red!12} \textbf{36} & \cellcolor{red!20} 27 \\*
Gene Expression & 17 & \cellcolor{red!12} 36 & \cellcolor{red!11} 37 & \cellcolor{red!16} 32 & \cellcolor{red!16} 32 & \cellcolor{red!11} 37 & \cellcolor{red!7} \textbf{42} & \cellcolor{red!11} 37 \\*
Embryology & 12 & \cellcolor{red!11} 37 & \cellcolor{red!11} 37 & \cellcolor{red!16} 32 & \cellcolor{red!26} 21 & \cellcolor{red!7} 42 & \cellcolor{red!11} 37 & \cellcolor{green!5} \textbf{56} \\
\noalign{\medskip}
\multicolumn{9}{@{}l@{}}{\cellcolor{gray!18}\quad\textbf{Physics}} \\*
Magnetism & 17 & \cellcolor{green!5} 56 & \cellcolor{green!5} 56 & \cellcolor{green!8} 59 & \cellcolor{green!8} 59 & \cellcolor{green!11} \textbf{62} & \cellcolor{green!5} 53 & \cellcolor{red!5} 47 \\*
Particle Physics & 17 & \cellcolor{green!5} 53 & \cellcolor{green!12} \textbf{64} & \cellcolor{green!5} 54 & \cellcolor{green!8} 59 & \cellcolor{red!9} 39 & \cellcolor{green!5} 50 & \cellcolor{green!5} 54 \\*
Quantum Simulation & 10 & \cellcolor{green!5} 53 & \cellcolor{green!19} \textbf{71} & \cellcolor{green!12} 64 & \cellcolor{green!10} 62 & \cellcolor{red!19} 29 & \cellcolor{red!19} 29 & \cellcolor{green!12} 64 \\*
\cmidrule{1-9}
Quantum Hall Effect & 10 & \cellcolor{red!18} 29 & \cellcolor{red!21} 27 & \cellcolor{red!27} 20 & \cellcolor{red!21} 27 & \cellcolor{red!21} 27 & \cellcolor{red!27} 20 & \cellcolor{green!5} \textbf{53} \\*
Optical Engineering & 11 & \cellcolor{red!14} 34 & \cellcolor{red!23} 24 & \cellcolor{red!10} 38 & \cellcolor{red!15} 33 & \cellcolor{red!6} \textbf{43} & \cellcolor{red!19} 29 & \cellcolor{red!10} 38 \\*
Quantum Optics & 13 & \cellcolor{red!14} 34 & \cellcolor{red!16} 32 & \cellcolor{red!7} 42 & \cellcolor{red!21} 26 & \cellcolor{red!21} 26 & \cellcolor{red!16} 32 & \cellcolor{red!5} \textbf{47} \\
\noalign{\medskip}
\multicolumn{9}{@{}l@{}}{\cellcolor{gray!18}\quad\textbf{Medicine}} \\*
Alzheimer's Disease & 15 & \cellcolor{green!19} 72 & \cellcolor{green!31} \textbf{85} & \cellcolor{green!31} \textbf{85} & \cellcolor{green!17} 69 & \cellcolor{green!17} 69 & \cellcolor{green!10} 62 & \cellcolor{green!10} 62 \\*
Epigenetics & 17 & \cellcolor{green!15} 67 & \cellcolor{green!10} 62 & \cellcolor{green!23} \textbf{76} & \cellcolor{green!14} 67 & \cellcolor{green!23} \textbf{76} & \cellcolor{green!23} \textbf{76} & \cellcolor{red!5} 48 \\*
Proteomics & 11 & \cellcolor{green!14} 67 & \cellcolor{green!36} \textbf{90} & \cellcolor{green!17} 70 & \cellcolor{green!17} 70 & \cellcolor{green!5} 50 & \cellcolor{green!5} 50 & \cellcolor{green!17} 70 \\*
\cmidrule{1-9}
T Cell Biology & 16 & \cellcolor{red!13} 35 & \cellcolor{red!23} 24 & \cellcolor{red!13} 35 & \cellcolor{red!23} 24 & \cellcolor{red!7} 41 & \cellcolor{green!7} \textbf{59} & \cellcolor{red!22} 25 \\*
Drug Discovery & 10 & \cellcolor{red!10} 39 & \cellcolor{red!22} 25 & \cellcolor{green!5} 50 & \cellcolor{green!5} 50 & \cellcolor{red!22} 25 & \cellcolor{red!22} 25 & \cellcolor{green!7} \textbf{58} \\*
Single-Cell RNA Sequencing & 14 & \cellcolor{red!9} 39 & \cellcolor{green!6} \textbf{57} & \cellcolor{red!7} 42 & \cellcolor{red!32} 14 & \cellcolor{red!6} 43 & \cellcolor{red!6} 43 & \cellcolor{red!12} 36 \\
\noalign{\medskip}
\multicolumn{9}{@{}l@{}}{\cellcolor{gray!18}\quad\textbf{Artificial Intelligence}} \\*
AI Benchmarks & 59 & \cellcolor{green!13} 64 & \cellcolor{green!13} 65 & \cellcolor{green!16} 68 & \cellcolor{green!7} 58 & \cellcolor{green!19} \textbf{72} & \cellcolor{green!13} 65 & \cellcolor{green!7} 58 \\*
Reasoning & 47 & \cellcolor{green!11} 63 & \cellcolor{green!13} 64 & \cellcolor{green!14} 66 & \cellcolor{green!6} 57 & \cellcolor{green!19} \textbf{71} & \cellcolor{green!11} 63 & \cellcolor{green!5} 56 \\*
Large Multimodal Models & 11 & \cellcolor{green!10} 62 & \cellcolor{green!5} 50 & \cellcolor{green!12} 64 & \cellcolor{green!24} 77 & \cellcolor{green!12} 64 & \cellcolor{green!25} \textbf{79} & \cellcolor{red!12} 36 \\*
\cmidrule{1-9}
Policy Gradient Methods & 10 & \cellcolor{red!13} 35 & \cellcolor{red!18} 30 & \cellcolor{green!5} \textbf{50} & \cellcolor{red!27} 20 & \cellcolor{red!8} 40 & \cellcolor{green!5} \textbf{50} & \cellcolor{red!25} 22 \\*
Language Modeling & 10 & \cellcolor{red!11} 37 & \cellcolor{red!15} 33 & \cellcolor{red!5} \textbf{44} & \cellcolor{red!15} 33 & \cellcolor{red!5} \textbf{44} & \cellcolor{red!5} \textbf{44} & \cellcolor{red!25} 22 \\*
Neural Network Architectures & 10 & \cellcolor{red!10} 39 & \cellcolor{green!5} \textbf{50} & \cellcolor{red!15} 33 & \cellcolor{red!15} 33 & \cellcolor{red!7} 42 & \cellcolor{red!7} 42 & \cellcolor{red!15} 33 \\
\noalign{\medskip}
\multicolumn{9}{@{}l@{}}{\cellcolor{gray!18}\quad\textbf{Materials Science}} \\*
Perovskite Materials & 21 & \cellcolor{green!8} 60 & \cellcolor{green!12} 64 & \cellcolor{green!8} 59 & \cellcolor{green!5} 56 & \cellcolor{green!19} \textbf{72} & \cellcolor{green!5} 54 & \cellcolor{green!5} 54 \\*
Optoelectronics & 30 & \cellcolor{green!5} 55 & \cellcolor{green!9} 60 & \cellcolor{green!5} 55 & \cellcolor{green!5} 51 & \cellcolor{green!11} \textbf{62} & \cellcolor{green!5} 51 & \cellcolor{red!5} 49 \\*
2D Materials & 22 & \cellcolor{green!5} 53 & \cellcolor{green!8} \textbf{60} & \cellcolor{green!5} 54 & \cellcolor{green!5} 50 & \cellcolor{green!5} 51 & \cellcolor{green!5} 54 & \cellcolor{red!5} 49 \\*
\cmidrule{1-9}
Flexible Electronics & 11 & \cellcolor{red!13} 35 & \cellcolor{red!18} 29 & \cellcolor{red!13} 35 & \cellcolor{red!22} 25 & \cellcolor{red!23} 24 & \cellcolor{red!7} 41 & \cellcolor{green!5} \textbf{53} \\*
Electrochemistry & 24 & \cellcolor{red!8} 41 & \cellcolor{red!5} \textbf{49} & \cellcolor{red!9} 39 & \cellcolor{red!14} 34 & \cellcolor{red!9} 39 & \cellcolor{red!9} 39 & \cellcolor{red!5} 46 \\*
Polymer Science & 27 & \cellcolor{red!5} 44 & \cellcolor{red!5} 45 & \cellcolor{red!5} 44 & \cellcolor{red!9} 39 & \cellcolor{red!11} 37 & \cellcolor{red!5} 47 & \cellcolor{green!5} \textbf{50} \\
\noalign{\medskip}
\multicolumn{9}{@{}l@{}}{\cellcolor{gray!18}\quad\textbf{Chemistry}} \\*
Organometallic Chemistry & 25 & \cellcolor{green!5} 53 & \cellcolor{green!6} \textbf{57} & \cellcolor{green!5} 54 & \cellcolor{green!5} 51 & \cellcolor{green!6} \textbf{57} & \cellcolor{red!5} 49 & \cellcolor{red!5} 49 \\*
Computational Chemistry & 14 & \cellcolor{red!5} 48 & \cellcolor{green!8} \textbf{60} & \cellcolor{red!13} 35 & \cellcolor{green!5} 50 & \cellcolor{red!5} 45 & \cellcolor{red!8} 40 & \cellcolor{green!5} 55 \\*
Photochemistry & 14 & \cellcolor{red!5} 47 & \cellcolor{green!5} 53 & \cellcolor{green!5} 56 & \cellcolor{green!7} \textbf{58} & \cellcolor{red!11} 37 & \cellcolor{red!11} 37 & \cellcolor{red!7} 42 \\*
\cmidrule{1-9}
Biocatalysis & 11 & \cellcolor{red!12} 36 & \cellcolor{red!20} 27 & \cellcolor{red!12} 36 & \cellcolor{red!12} 36 & \cellcolor{red!5} \textbf{45} & \cellcolor{red!20} 27 & \cellcolor{red!5} \textbf{45} \\*
Chemical Engineering & 13 & \cellcolor{red!10} 39 & \cellcolor{red!5} 45 & \cellcolor{red!20} 27 & \cellcolor{red!19} 29 & \cellcolor{red!12} 36 & \cellcolor{red!12} 36 & \cellcolor{green!8} \textbf{59} \\*
Medicinal Chemistry & 12 & \cellcolor{red!9} 40 & \cellcolor{red!5} 44 & \cellcolor{red!10} 39 & \cellcolor{red!15} 33 & \cellcolor{red!15} 33 & \cellcolor{red!10} 39 & \cellcolor{green!5} \textbf{50} \\
\end{longtable}

\setlength{\tabcolsep}{3pt} 
\renewcommand{\arraystretch}{0.95}
\begin{longtable}{@{}l r r r r r r r r @{}}
\caption{Sub-domain predictability — Date prediction score (colour centre = cross-model mean 0.29) (values: score 0--1). Chance level = n/a. Cells shaded \colorbox{green!25}{\,} above / \colorbox{red!25}{\,} below chance. Cross-model \textit{mean} shown; best individual model per row in \textbf{bold}. Top 3 (most predictable) and bottom 3 (least predictable) sub-domains per area, ranked by cross-model mean. $n$: mean sample size across models.} \label{tab:subdomain_date} \\
\toprule
\textbf{Sub-domain} & $\boldsymbol{n}$ & \textbf{Mean} & \rotatebox{55}{\textbf{GPT-5.4}} & \rotatebox{55}{\textbf{Claude S4.5}} & \rotatebox{55}{\textbf{DeepSeek R1}} & \rotatebox{55}{\textbf{GPT-OSS}} & \rotatebox{55}{\textbf{GPT-4o}} & \rotatebox{55}{\textbf{LLaMA 3.3}} \\
\midrule
\endfirsthead
\multicolumn{9}{l}{\small\tablename~\thetable{} \textit{(continued)}} \\
\toprule
\textbf{Sub-domain} & $\boldsymbol{n}$ & \textbf{Mean} & \rotatebox{55}{\textbf{GPT-5.4}} & \rotatebox{55}{\textbf{Claude S4.5}} & \rotatebox{55}{\textbf{DeepSeek R1}} & \rotatebox{55}{\textbf{GPT-OSS}} & \rotatebox{55}{\textbf{GPT-4o}} & \rotatebox{55}{\textbf{LLaMA 3.3}} \\
\midrule
\endhead
\midrule
\multicolumn{9}{r}{\small\textit{Continued on next page}} \\
\endfoot
\bottomrule
\endlastfoot
\multicolumn{9}{@{}l@{}}{\cellcolor{gray!18}\quad\textbf{Other}} \\*
Astrophysics & 23 & \cellcolor{red!5} 0.27 & \cellcolor{red!11} 0.22 & \cellcolor{red!5} 0.27 & \cellcolor{red!19} 0.16 & \cellcolor{green!5} 0.30 & \cellcolor{red!24} 0.14 & \cellcolor{green!14} \textbf{0.52} \\*
Cosmology & 14 & \cellcolor{red!14} 0.20 & \cellcolor{red!24} 0.13 & \cellcolor{red!6} 0.25 & \cellcolor{red!17} 0.18 & \cellcolor{red!18} 0.17 & \cellcolor{red!26} 0.12 & \cellcolor{green!5} \textbf{0.35} \\*
\cmidrule{1-9}
Cosmology & 14 & \cellcolor{red!14} 0.20 & \cellcolor{red!24} 0.13 & \cellcolor{red!6} 0.25 & \cellcolor{red!17} 0.18 & \cellcolor{red!18} 0.17 & \cellcolor{red!26} 0.12 & \cellcolor{green!5} \textbf{0.35} \\*
Astrophysics & 23 & \cellcolor{red!5} 0.27 & \cellcolor{red!11} 0.22 & \cellcolor{red!5} 0.27 & \cellcolor{red!19} 0.16 & \cellcolor{green!5} 0.30 & \cellcolor{red!24} 0.14 & \cellcolor{green!14} \textbf{0.52} \\
\noalign{\medskip}
\multicolumn{9}{@{}l@{}}{\cellcolor{gray!18}\quad\textbf{Environmental Science}} \\*
Public Health & 10 & \cellcolor{green!6} 0.39 & \cellcolor{red!10} 0.22 & \cellcolor{green!14} 0.51 & \cellcolor{green!5} 0.32 & \cellcolor{green!19} \textbf{0.60} & \cellcolor{red!22} 0.15 & \cellcolor{green!15} 0.54 \\*
Atmospheric Science & 15 & \cellcolor{green!5} 0.34 & \cellcolor{red!14} 0.20 & \cellcolor{red!9} 0.23 & \cellcolor{red!5} 0.26 & \cellcolor{green!16} 0.55 & \cellcolor{red!25} 0.13 & \cellcolor{green!25} \textbf{0.70} \\*
Hydrology & 19 & \cellcolor{green!5} 0.30 & \cellcolor{red!5} 0.26 & \cellcolor{red!26} 0.12 & \cellcolor{green!5} 0.29 & \cellcolor{green!5} 0.38 & \cellcolor{red!26} 0.12 & \cellcolor{green!19} \textbf{0.60} \\*
\cmidrule{1-9}
Oceanography & 20 & \cellcolor{red!15} 0.19 & \cellcolor{red!11} 0.22 & \cellcolor{red!19} 0.17 & \cellcolor{red!15} 0.19 & \cellcolor{red!31} 0.09 & \cellcolor{red!40} 0.03 & \cellcolor{green!9} \textbf{0.44} \\*
Biodiversity & 26 & \cellcolor{red!14} 0.20 & \cellcolor{red!5} 0.27 & \cellcolor{red!36} 0.06 & \cellcolor{red!29} 0.10 & \cellcolor{red!5} 0.29 & \cellcolor{red!22} 0.15 & \cellcolor{green!5} \textbf{0.34} \\*
Conservation Biology & 28 & \cellcolor{red!13} 0.21 & \cellcolor{green!5} \textbf{0.36} & \cellcolor{red!22} 0.14 & \cellcolor{red!27} 0.11 & \cellcolor{red!12} 0.21 & \cellcolor{red!26} 0.12 & \cellcolor{red!5} 0.28 \\
\noalign{\medskip}
\multicolumn{9}{@{}l@{}}{\cellcolor{gray!18}\quad\textbf{Neuroscience}} \\*
Cryo-Electron Microscopy & 16 & \cellcolor{green!8} 0.42 & \cellcolor{red!6} 0.25 & \cellcolor{red!5} 0.28 & \cellcolor{green!14} 0.52 & \cellcolor{green!15} 0.53 & \cellcolor{red!5} 0.25 & \cellcolor{green!24} \textbf{0.68} \\*
Alzheimer's Disease & 16 & \cellcolor{green!5} 0.38 & \cellcolor{green!5} 0.38 & \cellcolor{green!12} \textbf{0.49} & \cellcolor{green!11} 0.48 & \cellcolor{green!5} 0.30 & \cellcolor{red!18} 0.17 & \cellcolor{green!12} 0.48 \\*
Developmental Biology & 10 & \cellcolor{green!5} 0.36 & \cellcolor{red!34} 0.07 & \cellcolor{red!21} 0.15 & \cellcolor{green!23} \textbf{0.67} & \cellcolor{green!12} 0.49 & \cellcolor{red!19} 0.17 & \cellcolor{green!21} 0.62 \\*
\cmidrule{1-9}
Immunology & 12 & \cellcolor{red!16} 0.18 & \cellcolor{red!29} 0.10 & \cellcolor{red!40} 0.03 & \cellcolor{red!25} 0.12 & \cellcolor{red!10} 0.22 & \cellcolor{red!43} 0.01 & \cellcolor{green!20} \textbf{0.61} \\*
Neural Networks & 11 & \cellcolor{red!15} 0.19 & \cellcolor{red!21} 0.15 & \cellcolor{red!30} 0.09 & \cellcolor{green!5} 0.29 & \cellcolor{red!25} 0.12 & \cellcolor{red!39} 0.04 & \cellcolor{green!8} \textbf{0.43} \\*
Neuroimmunology & 23 & \cellcolor{red!15} 0.19 & \cellcolor{red!35} 0.06 & \cellcolor{red!37} 0.05 & \cellcolor{red!6} 0.25 & \cellcolor{red!14} 0.20 & \cellcolor{red!35} 0.06 & \cellcolor{green!15} \textbf{0.53} \\
\noalign{\medskip}
\multicolumn{9}{@{}l@{}}{\cellcolor{gray!18}\quad\textbf{Biology}} \\*
Archaeogenetics & 11 & \cellcolor{green!7} 0.41 & \cellcolor{green!5} 0.38 & \cellcolor{green!5} 0.29 & \cellcolor{green!10} 0.46 & \cellcolor{green!7} 0.41 & \cellcolor{green!12} \textbf{0.49} & \cellcolor{green!9} 0.45 \\*
Pharmacology & 10 & \cellcolor{green!6} 0.40 & \cellcolor{red!21} 0.15 & \cellcolor{green!5} 0.38 & \cellcolor{green!8} 0.42 & \cellcolor{red!5} 0.29 & \cellcolor{green!9} 0.44 & \cellcolor{green!25} \textbf{0.69} \\*
Crop Science & 11 & \cellcolor{green!6} 0.39 & \cellcolor{red!5} 0.28 & \cellcolor{red!24} 0.13 & \cellcolor{green!14} 0.52 & \cellcolor{green!12} 0.49 & \cellcolor{red!15} 0.19 & \cellcolor{green!26} \textbf{0.72} \\*
\cmidrule{1-9}
Genetic Engineering & 12 & \cellcolor{red!27} 0.11 & \cellcolor{red!33} 0.07 & \cellcolor{red!40} 0.03 & \cellcolor{red!32} 0.08 & \cellcolor{red!28} 0.11 & \cellcolor{red!42} 0.02 & \cellcolor{green!5} \textbf{0.37} \\*
Marine Biology & 21 & \cellcolor{red!21} 0.15 & \cellcolor{red!17} 0.18 & \cellcolor{red!31} 0.09 & \cellcolor{red!33} 0.08 & \cellcolor{red!37} 0.05 & \cellcolor{red!31} 0.09 & \cellcolor{green!8} \textbf{0.43} \\*
Entomology & 14 & \cellcolor{red!17} 0.18 & \cellcolor{red!36} 0.05 & \cellcolor{red!44} 0.00 & \cellcolor{red!6} 0.25 & \cellcolor{red!16} 0.19 & \cellcolor{red!34} 0.07 & \cellcolor{green!13} \textbf{0.51} \\
\noalign{\medskip}
\multicolumn{9}{@{}l@{}}{\cellcolor{gray!18}\quad\textbf{Physics}} \\*
Particle Physics & 17 & \cellcolor{red!5} 0.29 & \cellcolor{red!19} 0.16 & \cellcolor{red!23} 0.14 & \cellcolor{green!5} 0.35 & \cellcolor{green!5} 0.30 & \cellcolor{red!21} 0.15 & \cellcolor{green!21} \textbf{0.63} \\*
Quantum Information & 25 & \cellcolor{red!5} 0.26 & \cellcolor{red!14} 0.20 & \cellcolor{red!13} 0.20 & \cellcolor{red!7} 0.24 & \cellcolor{green!5} 0.30 & \cellcolor{red!28} 0.11 & \cellcolor{green!14} \textbf{0.53} \\*
Materials Science & 30 & \cellcolor{red!5} 0.26 & \cellcolor{red!13} 0.20 & \cellcolor{red!14} 0.19 & \cellcolor{green!5} 0.32 & \cellcolor{red!28} 0.10 & \cellcolor{red!21} 0.15 & \cellcolor{green!18} \textbf{0.58} \\*
\cmidrule{1-9}
Topological Insulators & 20 & \cellcolor{red!21} 0.15 & \cellcolor{red!29} 0.10 & \cellcolor{red!30} 0.09 & \cellcolor{red!9} 0.23 & \cellcolor{red!26} 0.12 & \cellcolor{red!31} 0.09 & \cellcolor{red!5} \textbf{0.28} \\*
Optical Engineering & 11 & \cellcolor{red!20} 0.16 & \cellcolor{red!41} 0.03 & \cellcolor{red!32} 0.08 & \cellcolor{red!23} 0.14 & \cellcolor{red!23} 0.14 & \cellcolor{red!30} 0.09 & \cellcolor{green!11} \textbf{0.47} \\*
Atomic Physics & 13 & \cellcolor{red!20} 0.16 & \cellcolor{red!35} 0.06 & \cellcolor{red!37} 0.05 & \cellcolor{red!19} 0.16 & \cellcolor{red!23} 0.14 & \cellcolor{red!42} 0.01 & \cellcolor{green!15} \textbf{0.53} \\
\noalign{\medskip}
\multicolumn{9}{@{}l@{}}{\cellcolor{gray!18}\quad\textbf{Medicine}} \\*
Fibrosis & 11 & \cellcolor{green!8} 0.43 & \cellcolor{green!17} 0.56 & \cellcolor{red!5} 0.26 & \cellcolor{green!15} 0.54 & \cellcolor{green!11} 0.48 & \cellcolor{red!31} 0.09 & \cellcolor{green!23} \textbf{0.65} \\*
Microbiome & 13 & \cellcolor{green!6} 0.40 & \cellcolor{red!5} 0.26 & \cellcolor{green!5} 0.31 & \cellcolor{green!17} 0.57 & \cellcolor{green!5} 0.38 & \cellcolor{red!13} 0.20 & \cellcolor{green!24} \textbf{0.68} \\*
Structural Biology & 20 & \cellcolor{green!5} 0.38 & \cellcolor{red!5} 0.26 & \cellcolor{green!12} 0.48 & \cellcolor{green!5} 0.36 & \cellcolor{green!9} 0.45 & \cellcolor{red!9} 0.23 & \cellcolor{green!14} \textbf{0.52} \\*
\cmidrule{1-9}
Translational Medicine & 10 & \cellcolor{red!23} 0.14 & \cellcolor{red!23} 0.14 & \cellcolor{red!12} 0.21 & \cellcolor{red!30} 0.09 & \cellcolor{red!27} 0.12 & \cellcolor{red!39} 0.04 & \cellcolor{red!6} \textbf{0.25} \\*
HIV Research & 13 & \cellcolor{red!20} 0.16 & \cellcolor{red!35} 0.06 & \cellcolor{red!24} 0.13 & \cellcolor{red!34} 0.07 & \cellcolor{red!26} 0.12 & \cellcolor{red!37} 0.05 & \cellcolor{green!13} \textbf{0.51} \\*
Tissue Engineering & 14 & \cellcolor{red!18} 0.17 & \cellcolor{red!19} 0.16 & \cellcolor{red!24} 0.13 & \cellcolor{red!14} 0.20 & \cellcolor{red!25} 0.13 & \cellcolor{red!41} 0.02 & \cellcolor{green!5} \textbf{0.38} \\
\noalign{\medskip}
\multicolumn{9}{@{}l@{}}{\cellcolor{gray!18}\quad\textbf{Artificial Intelligence}} \\*
Policy Gradient Methods & 10 & \cellcolor{green!21} 0.63 & \cellcolor{green!20} 0.61 & \cellcolor{green!26} 0.71 & \cellcolor{green!12} 0.49 & \cellcolor{green!39} \textbf{0.91} & \cellcolor{green!14} 0.51 & \cellcolor{green!14} 0.52 \\*
Agent-Based Systems & 16 & \cellcolor{green!19} 0.60 & \cellcolor{green!22} 0.65 & \cellcolor{green!27} \textbf{0.73} & \cellcolor{green!5} 0.33 & \cellcolor{green!18} 0.59 & \cellcolor{green!22} 0.64 & \cellcolor{green!24} 0.68 \\*
Cognitive Computing & 12 & \cellcolor{green!19} 0.59 & \cellcolor{green!19} 0.59 & \cellcolor{green!19} 0.60 & \cellcolor{green!21} \textbf{0.62} & \cellcolor{green!19} 0.60 & \cellcolor{green!17} 0.57 & \cellcolor{green!17} 0.56 \\*
\cmidrule{1-9}
Neural Network Architectures & 10 & \cellcolor{green!5} 0.29 & \cellcolor{red!5} 0.29 & \cellcolor{red!14} 0.20 & \cellcolor{red!15} 0.19 & \cellcolor{green!5} 0.37 & \cellcolor{red!18} 0.17 & \cellcolor{green!15} \textbf{0.54} \\*
Knowledge Representation & 14 & \cellcolor{green!5} 0.31 & \cellcolor{red!16} 0.19 & \cellcolor{red!31} 0.09 & \cellcolor{green!5} 0.35 & \cellcolor{red!11} 0.22 & \cellcolor{green!5} 0.38 & \cellcolor{green!21} \textbf{0.63} \\*
Zero-shot Learning & 11 & \cellcolor{green!5} 0.31 & \cellcolor{green!9} \textbf{0.45} & \cellcolor{red!14} 0.20 & \cellcolor{green!5} 0.34 & \cellcolor{red!24} 0.13 & \cellcolor{green!5} 0.32 & \cellcolor{green!9} 0.44 \\
\noalign{\medskip}
\multicolumn{9}{@{}l@{}}{\cellcolor{gray!18}\quad\textbf{Materials Science}} \\*
Optoelectronics & 30 & \cellcolor{red!5} 0.27 & \cellcolor{red!20} 0.16 & \cellcolor{red!20} 0.16 & \cellcolor{red!8} 0.24 & \cellcolor{green!7} 0.41 & \cellcolor{red!23} 0.14 & \cellcolor{green!16} \textbf{0.54} \\*
Perovskite Solar Cells & 50 & \cellcolor{red!5} 0.27 & \cellcolor{red!14} 0.20 & \cellcolor{red!21} 0.15 & \cellcolor{red!5} 0.27 & \cellcolor{green!5} 0.36 & \cellcolor{red!16} 0.18 & \cellcolor{green!10} \textbf{0.46} \\*
Perovskite Materials & 21 & \cellcolor{red!5} 0.27 & \cellcolor{red!24} 0.13 & \cellcolor{red!21} 0.15 & \cellcolor{red!5} 0.27 & \cellcolor{green!7} 0.41 & \cellcolor{red!23} 0.14 & \cellcolor{green!14} \textbf{0.52} \\*
\cmidrule{1-9}
Mechanical Properties & 15 & \cellcolor{red!29} 0.10 & \cellcolor{red!24} 0.13 & \cellcolor{red!39} 0.04 & \cellcolor{red!38} 0.04 & \cellcolor{red!39} 0.03 & \cellcolor{red!43} 0.01 & \cellcolor{green!5} \textbf{0.33} \\*
Flexible Electronics & 11 & \cellcolor{red!27} 0.11 & \cellcolor{red!21} 0.15 & \cellcolor{red!41} 0.02 & \cellcolor{red!28} 0.11 & \cellcolor{red!32} 0.08 & \cellcolor{red!35} 0.06 & \cellcolor{red!5} \textbf{0.26} \\*
Energy Storage & 25 & \cellcolor{red!25} 0.13 & \cellcolor{red!39} 0.04 & \cellcolor{red!20} 0.16 & \cellcolor{red!33} 0.07 & \cellcolor{red!25} 0.12 & \cellcolor{red!37} 0.05 & \cellcolor{green!5} \textbf{0.33} \\
\noalign{\medskip}
\multicolumn{9}{@{}l@{}}{\cellcolor{gray!18}\quad\textbf{Chemistry}} \\*
Medicinal Chemistry & 12 & \cellcolor{green!5} 0.29 & \cellcolor{red!26} 0.12 & \cellcolor{red!5} 0.28 & \cellcolor{red!18} 0.17 & \cellcolor{red!5} 0.29 & \cellcolor{red!5} 0.26 & \cellcolor{green!21} \textbf{0.63} \\*
Computational Chemistry & 14 & \cellcolor{red!10} 0.22 & \cellcolor{red!16} 0.19 & \cellcolor{red!30} 0.09 & \cellcolor{red!18} 0.17 & \cellcolor{green!5} 0.32 & \cellcolor{red!25} 0.13 & \cellcolor{green!8} \textbf{0.43} \\*
Organometallic Chemistry & 25 & \cellcolor{red!11} 0.22 & \cellcolor{red!10} 0.22 & \cellcolor{red!27} 0.11 & \cellcolor{red!20} 0.16 & \cellcolor{green!5} 0.30 & \cellcolor{red!29} 0.10 & \cellcolor{green!7} \textbf{0.41} \\*
\cmidrule{1-9}
Chemical Engineering & 13 & \cellcolor{red!35} 0.06 & \cellcolor{red!42} 0.02 & \cellcolor{red!40} 0.03 & \cellcolor{red!38} 0.04 & \cellcolor{red!44} 0.00 & \cellcolor{red!42} 0.01 & \cellcolor{red!5} \textbf{0.28} \\*
Electrochemistry & 17 & \cellcolor{red!25} 0.13 & \cellcolor{red!38} 0.04 & \cellcolor{red!26} 0.12 & \cellcolor{red!30} 0.10 & \cellcolor{red!27} 0.11 & \cellcolor{red!41} 0.02 & \cellcolor{green!5} \textbf{0.38} \\*
Radical Chemistry & 14 & \cellcolor{red!22} 0.15 & \cellcolor{red!33} 0.07 & \cellcolor{red!30} 0.09 & \cellcolor{red!29} 0.10 & \cellcolor{red!19} 0.16 & \cellcolor{red!39} 0.03 & \cellcolor{green!8} \textbf{0.43} \\
\end{longtable}

\setlength{\tabcolsep}{3pt} 
\renewcommand{\arraystretch}{0.95}
\begin{longtable}{@{}l r r r r r r r r @{}}
\caption{Sub-domain predictability — MCQ, 4-choice (values in \%). Chance level = 25\%. Cells shaded \colorbox{green!25}{\,} above / \colorbox{red!25}{\,} below chance. Cross-model \textit{mean} shown; best individual model per row in \textbf{bold}. Top 3 (most predictable) and bottom 3 (least predictable) sub-domains per area, ranked by cross-model mean. $n$: mean sample size across models.} \label{tab:subdomain_mcq} \\
\toprule
\textbf{Sub-domain} & $\boldsymbol{n}$ & \textbf{Mean} & \rotatebox{55}{\textbf{GPT-5.4}} & \rotatebox{55}{\textbf{Claude S4.5}} & \rotatebox{55}{\textbf{DeepSeek R1}} & \rotatebox{55}{\textbf{GPT-OSS}} & \rotatebox{55}{\textbf{GPT-4o}} & \rotatebox{55}{\textbf{LLaMA 3.3}} \\
\midrule
\endfirsthead
\multicolumn{9}{l}{\small\tablename~\thetable{} \textit{(continued)}} \\
\toprule
\textbf{Sub-domain} & $\boldsymbol{n}$ & \textbf{Mean} & \rotatebox{55}{\textbf{GPT-5.4}} & \rotatebox{55}{\textbf{Claude S4.5}} & \rotatebox{55}{\textbf{DeepSeek R1}} & \rotatebox{55}{\textbf{GPT-OSS}} & \rotatebox{55}{\textbf{GPT-4o}} & \rotatebox{55}{\textbf{LLaMA 3.3}} \\
\midrule
\endhead
\midrule
\multicolumn{9}{r}{\small\textit{Continued on next page}} \\
\endfoot
\bottomrule
\endlastfoot
\multicolumn{9}{@{}l@{}}{\cellcolor{gray!18}\quad\textbf{Other}} \\*
Cosmology & 14 & \cellcolor{green!32} 80 & \cellcolor{green!40} \textbf{92} & \cellcolor{green!35} 85 & \cellcolor{green!28} 73 & \cellcolor{green!31} 77 & \cellcolor{green!35} 85 & \cellcolor{green!24} 67 \\*
Astrophysics & 23 & \cellcolor{green!31} 78 & \cellcolor{green!39} 90 & \cellcolor{green!33} 80 & \cellcolor{green!41} \textbf{94} & \cellcolor{green!24} 65 & \cellcolor{green!33} 80 & \cellcolor{green!20} 60 \\*
\cmidrule{1-9}
Astrophysics & 23 & \cellcolor{green!31} 78 & \cellcolor{green!39} 90 & \cellcolor{green!33} 80 & \cellcolor{green!41} \textbf{94} & \cellcolor{green!24} 65 & \cellcolor{green!33} 80 & \cellcolor{green!20} 60 \\*
Cosmology & 14 & \cellcolor{green!32} 80 & \cellcolor{green!40} \textbf{92} & \cellcolor{green!35} 85 & \cellcolor{green!28} 73 & \cellcolor{green!31} 77 & \cellcolor{green!35} 85 & \cellcolor{green!24} 67 \\
\noalign{\medskip}
\multicolumn{9}{@{}l@{}}{\cellcolor{gray!18}\quad\textbf{Environmental Science}} \\*
Public Health & 10 & \cellcolor{green!41} 94 & \cellcolor{green!45} \textbf{100} & \cellcolor{green!45} \textbf{100} & \cellcolor{green!45} \textbf{100} & \cellcolor{green!35} 83 & \cellcolor{green!35} 83 & \cellcolor{green!45} \textbf{100} \\*
Glaciology & 11 & \cellcolor{green!30} 76 & \cellcolor{green!45} \textbf{100} & \cellcolor{green!33} 80 & \cellcolor{green!38} 89 & \cellcolor{green!27} 70 & \cellcolor{green!20} 60 & \cellcolor{green!20} 60 \\*
Atmospheric Science & 15 & \cellcolor{green!29} 74 & \cellcolor{green!45} \textbf{100} & \cellcolor{green!37} 88 & \cellcolor{green!31} 78 & \cellcolor{green!18} 56 & \cellcolor{green!31} 78 & \cellcolor{green!11} 44 \\*
\cmidrule{1-9}
Hydrology & 19 & \cellcolor{green!21} 61 & \cellcolor{green!37} 88 & \cellcolor{green!41} \textbf{94} & \cellcolor{green!16} 53 & \cellcolor{green!16} 53 & \cellcolor{green!9} 41 & \cellcolor{green!6} 35 \\*
Ecology & 46 & \cellcolor{green!22} 63 & \cellcolor{green!37} \textbf{88} & \cellcolor{green!33} 81 & \cellcolor{green!11} 43 & \cellcolor{green!15} 52 & \cellcolor{green!19} 58 & \cellcolor{green!17} 55 \\*
Biodiversity & 26 & \cellcolor{green!23} 64 & \cellcolor{green!37} \textbf{88} & \cellcolor{green!35} 85 & \cellcolor{green!15} 50 & \cellcolor{green!15} 50 & \cellcolor{green!18} 56 & \cellcolor{green!18} 56 \\
\noalign{\medskip}
\multicolumn{9}{@{}l@{}}{\cellcolor{gray!18}\quad\textbf{Neuroscience}} \\*
Cryo-Electron Microscopy & 16 & \cellcolor{green!38} 88 & \cellcolor{green!37} 88 & \cellcolor{green!37} 87 & \cellcolor{green!41} 93 & \cellcolor{green!37} 88 & \cellcolor{green!41} \textbf{94} & \cellcolor{green!33} 81 \\*
Calcium Imaging & 15 & \cellcolor{green!35} 83 & \cellcolor{green!35} 83 & \cellcolor{green!40} \textbf{92} & \cellcolor{green!35} 83 & \cellcolor{green!35} 83 & \cellcolor{green!30} 75 & \cellcolor{green!35} 83 \\*
Synaptic Plasticity & 22 & \cellcolor{green!34} 82 & \cellcolor{green!41} \textbf{94} & \cellcolor{green!33} 81 & \cellcolor{green!37} 87 & \cellcolor{green!37} 88 & \cellcolor{green!30} 75 & \cellcolor{green!26} 69 \\*
\cmidrule{1-9}
Neuroimmunology & 23 & \cellcolor{green!12} 46 & \cellcolor{green!24} \textbf{67} & \cellcolor{green!13} 47 & \cellcolor{green!23} 64 & \cellcolor{red!5} 22 & \cellcolor{green!11} 44 & \cellcolor{green!5} 33 \\*
Genetics & 20 & \cellcolor{green!16} 53 & \cellcolor{green!26} 68 & \cellcolor{green!27} \textbf{71} & \cellcolor{green!17} 53 & \cellcolor{green!10} 42 & \cellcolor{green!13} 47 & \cellcolor{green!7} 37 \\*
Neurodegenerative Diseases & 13 & \cellcolor{green!17} 54 & \cellcolor{green!20} 58 & \cellcolor{green!23} \textbf{64} & \cellcolor{green!20} 58 & \cellcolor{green!15} 50 & \cellcolor{green!15} 50 & \cellcolor{green!10} 42 \\
\noalign{\medskip}
\multicolumn{9}{@{}l@{}}{\cellcolor{gray!18}\quad\textbf{Biology}} \\*
Archaeogenetics & 11 & \cellcolor{green!43} 98 & \cellcolor{green!45} \textbf{100} & \cellcolor{green!45} \textbf{100} & \cellcolor{green!37} 88 & \cellcolor{green!45} \textbf{100} & \cellcolor{green!45} \textbf{100} & \cellcolor{green!45} \textbf{100} \\*
Single-Cell RNA Sequencing & 10 & \cellcolor{green!38} 90 & \cellcolor{green!45} \textbf{100} & \cellcolor{green!45} \textbf{100} & \cellcolor{green!38} 89 & \cellcolor{green!27} 70 & \cellcolor{green!39} 90 & \cellcolor{green!39} 90 \\*
Cryo-Electron Microscopy & 37 & \cellcolor{green!37} 88 & \cellcolor{green!43} \textbf{97} & \cellcolor{green!43} 97 & \cellcolor{green!41} 94 & \cellcolor{green!31} 78 & \cellcolor{green!38} 89 & \cellcolor{green!29} 74 \\*
\cmidrule{1-9}
Genetic Engineering & 12 & \cellcolor{green!12} 46 & \cellcolor{green!20} 58 & \cellcolor{green!20} \textbf{60} & \cellcolor{green!20} \textbf{60} & \cellcolor{green!10} 42 & \cellcolor{green!10} 42 & \cellcolor{red!15} 17 \\*
Physiology & 12 & \cellcolor{green!13} 47 & \cellcolor{green!31} \textbf{78} & \cellcolor{green!20} 60 & \cellcolor{green!5} 33 & \cellcolor{green!11} 44 & \cellcolor{green!5} 33 & \cellcolor{green!5} 33 \\*
Stem Cell Biology & 18 & \cellcolor{green!14} 48 & \cellcolor{green!30} \textbf{75} & \cellcolor{green!20} 58 & \cellcolor{green!9} 40 & \cellcolor{green!10} 42 & \cellcolor{green!10} 42 & \cellcolor{green!5} 33 \\
\noalign{\medskip}
\multicolumn{9}{@{}l@{}}{\cellcolor{gray!18}\quad\textbf{Physics}} \\*
Experimental Physics & 15 & \cellcolor{green!31} 77 & \cellcolor{green!45} \textbf{100} & \cellcolor{green!35} 83 & \cellcolor{green!33} 80 & \cellcolor{green!20} 58 & \cellcolor{green!24} 67 & \cellcolor{green!30} 75 \\*
Nonlinear Optics & 11 & \cellcolor{green!28} 72 & \cellcolor{green!39} \textbf{91} & \cellcolor{green!34} 82 & \cellcolor{green!33} 80 & \cellcolor{green!28} 73 & \cellcolor{green!39} \textbf{91} & \cellcolor{red!12} 18 \\*
Astrophysics & 32 & \cellcolor{green!27} 71 & \cellcolor{green!35} \textbf{83} & \cellcolor{green!35} \textbf{83} & \cellcolor{green!31} 77 & \cellcolor{green!23} 63 & \cellcolor{green!24} 67 & \cellcolor{green!17} 53 \\*
\cmidrule{1-9}
Statistical Mechanics & 12 & \cellcolor{green!13} 47 & \cellcolor{green!40} \textbf{92} & \cellcolor{green!20} 58 & \cellcolor{green!9} 40 & \cellcolor{green!10} 42 & \cellcolor{green!5} 25 & \cellcolor{green!5} 25 \\*
Graphene & 12 & \cellcolor{green!14} 49 & \cellcolor{green!27} \textbf{70} & \cellcolor{green!15} 50 & \cellcolor{green!10} 43 & \cellcolor{green!15} 50 & \cellcolor{green!15} 50 & \cellcolor{green!5} 30 \\*
Magnetism & 17 & \cellcolor{green!14} 50 & \cellcolor{green!22} 62 & \cellcolor{green!18} 56 & \cellcolor{green!24} \textbf{67} & \cellcolor{green!7} 38 & \cellcolor{green!15} 50 & \cellcolor{green!5} 25 \\
\noalign{\medskip}
\multicolumn{9}{@{}l@{}}{\cellcolor{gray!18}\quad\textbf{Medicine}} \\*
Single-Cell Sequencing & 10 & \cellcolor{green!36} 86 & \cellcolor{green!45} \textbf{100} & \cellcolor{green!39} 90 & \cellcolor{green!31} 78 & \cellcolor{green!27} 70 & \cellcolor{green!39} 90 & \cellcolor{green!39} 90 \\*
Alzheimer's Disease & 15 & \cellcolor{green!33} 81 & \cellcolor{green!45} \textbf{100} & \cellcolor{green!45} \textbf{100} & \cellcolor{green!35} 83 & \cellcolor{green!27} 71 & \cellcolor{green!27} 71 & \cellcolor{green!19} 57 \\*
Genomics & 47 & \cellcolor{green!31} 77 & \cellcolor{green!39} 91 & \cellcolor{green!41} \textbf{95} & \cellcolor{green!26} 69 & \cellcolor{green!28} 73 & \cellcolor{green!24} 66 & \cellcolor{green!26} 70 \\*
\cmidrule{1-9}
Gene Therapy & 17 & \cellcolor{green!5} 34 & \cellcolor{green!16} \textbf{53} & \cellcolor{green!11} 44 & \cellcolor{red!6} 21 & \cellcolor{red!5} 24 & \cellcolor{red!5} 24 & \cellcolor{green!9} 41 \\*
Cancer Therapy & 14 & \cellcolor{green!7} 38 & \cellcolor{green!20} \textbf{58} & \cellcolor{green!5} 25 & \cellcolor{green!5} 25 & \cellcolor{green!15} 50 & \cellcolor{green!5} 33 & \cellcolor{green!6} 36 \\*
Inflammation & 18 & \cellcolor{green!10} 42 & \cellcolor{green!21} \textbf{62} & \cellcolor{green!15} 50 & \cellcolor{green!10} 42 & \cellcolor{green!8} 38 & \cellcolor{green!5} 31 & \cellcolor{green!5} 31 \\
\noalign{\medskip}
\multicolumn{9}{@{}l@{}}{\cellcolor{gray!18}\quad\textbf{Artificial Intelligence}} \\*
Tool Use in AI & 10 & \cellcolor{green!31} 77 & \cellcolor{green!35} 83 & \cellcolor{green!45} \textbf{100} & \cellcolor{green!33} 80 & \cellcolor{green!24} 67 & \cellcolor{green!35} 83 & \cellcolor{green!15} 50 \\*
Inference Optimization & 11 & \cellcolor{green!30} 75 & \cellcolor{green!45} \textbf{100} & \cellcolor{green!45} \textbf{100} & \cellcolor{green!24} 67 & \cellcolor{green!19} 57 & \cellcolor{green!27} 71 & \cellcolor{green!19} 57 \\*
Monte Carlo Tree Search & 12 & \cellcolor{green!29} 75 & \cellcolor{green!39} \textbf{90} & \cellcolor{green!39} \textbf{90} & \cellcolor{green!37} 88 & \cellcolor{green!20} 60 & \cellcolor{green!27} 70 & \cellcolor{green!15} 50 \\*
\cmidrule{1-9}
Audio Processing & 10 & \cellcolor{red!5} 24 & \cellcolor{green!11} \textbf{44} & \cellcolor{red!5} 22 & \cellcolor{red!5} 22 & \cellcolor{red!5} 22 & \cellcolor{red!5} 22 & \cellcolor{red!25} 11 \\*
Speech Synthesis & 10 & \cellcolor{green!6} 35 & \cellcolor{green!20} \textbf{60} & \cellcolor{green!15} 50 & \cellcolor{green!5} 30 & \cellcolor{green!5} 30 & \cellcolor{green!5} 30 & \cellcolor{red!27} 10 \\*
Autoregressive Models & 21 & \cellcolor{green!6} 35 & \cellcolor{green!31} \textbf{78} & \cellcolor{green!24} 67 & \cellcolor{red!6} 21 & \cellcolor{red!35} 6 & \cellcolor{green!5} 28 & \cellcolor{red!25} 11 \\
\noalign{\medskip}
\multicolumn{9}{@{}l@{}}{\cellcolor{gray!18}\quad\textbf{Materials Science}} \\*
Energy Storage & 25 & \cellcolor{green!22} 62 & \cellcolor{green!39} 91 & \cellcolor{green!42} \textbf{96} & \cellcolor{green!24} 67 & \cellcolor{green!5} 30 & \cellcolor{green!18} 57 & \cellcolor{green!5} 30 \\*
Mechanical Properties & 15 & \cellcolor{green!21} 61 & \cellcolor{green!28} \textbf{73} & \cellcolor{green!28} \textbf{73} & \cellcolor{green!21} 62 & \cellcolor{green!17} 53 & \cellcolor{green!17} 53 & \cellcolor{green!17} 53 \\*
Polymer Science & 27 & \cellcolor{green!20} 59 & \cellcolor{green!35} \textbf{85} & \cellcolor{green!30} 76 & \cellcolor{green!23} 64 & \cellcolor{green!17} 54 & \cellcolor{green!12} 46 & \cellcolor{green!5} 31 \\*
\cmidrule{1-9}
Polymer Chemistry & 10 & \cellcolor{green!8} 39 & \cellcolor{green!9} 40 & \cellcolor{green!15} 50 & \cellcolor{green!18} \textbf{56} & \cellcolor{red!8} 20 & \cellcolor{red!8} 20 & \cellcolor{green!15} 50 \\*
Metallurgy & 13 & \cellcolor{green!13} 48 & \cellcolor{green!30} \textbf{75} & \cellcolor{green!10} 42 & \cellcolor{green!11} 44 & \cellcolor{green!10} 42 & \cellcolor{green!10} 42 & \cellcolor{green!10} 42 \\*
Perovskite Materials & 21 & \cellcolor{green!14} 49 & \cellcolor{green!39} \textbf{90} & \cellcolor{green!27} 71 & \cellcolor{green!22} 62 & \cellcolor{red!19} 14 & \cellcolor{green!7} 38 & \cellcolor{red!10} 19 \\
\noalign{\medskip}
\multicolumn{9}{@{}l@{}}{\cellcolor{gray!18}\quad\textbf{Chemistry}} \\*
Chemical Engineering & 13 & \cellcolor{green!28} 72 & \cellcolor{green!40} \textbf{92} & \cellcolor{green!26} 69 & \cellcolor{green!31} 78 & \cellcolor{green!17} 54 & \cellcolor{green!31} 77 & \cellcolor{green!21} 62 \\*
Biocatalysis & 11 & \cellcolor{green!24} 65 & \cellcolor{green!39} \textbf{91} & \cellcolor{green!23} 64 & \cellcolor{green!12} 45 & \cellcolor{green!23} 64 & \cellcolor{green!34} 82 & \cellcolor{green!12} 45 \\*
Photochemistry & 14 & \cellcolor{green!21} 60 & \cellcolor{green!40} \textbf{92} & \cellcolor{green!35} 83 & \cellcolor{green!17} 54 & \cellcolor{green!12} 46 & \cellcolor{green!12} 46 & \cellcolor{green!8} 38 \\*
\cmidrule{1-9}
Medicinal Chemistry & 12 & \cellcolor{green!5} 29 & \cellcolor{green!23} \textbf{64} & \cellcolor{red!12} 18 & \cellcolor{red!25} 11 & \cellcolor{green!6} 36 & \cellcolor{red!12} 18 & \cellcolor{green!5} 27 \\*
Synthetic Chemistry & 46 & \cellcolor{green!9} 40 & \cellcolor{green!33} \textbf{80} & \cellcolor{green!15} 51 & \cellcolor{green!6} 36 & \cellcolor{red!5} 22 & \cellcolor{green!5} 25 & \cellcolor{green!5} 25 \\*
Computational Chemistry & 14 & \cellcolor{green!9} 41 & \cellcolor{green!39} \textbf{91} & \cellcolor{green!24} 67 & \cellcolor{red!12} 18 & \cellcolor{green!5} 27 & \cellcolor{green!6} 36 & \cellcolor{red!28} 9 \\
\end{longtable}

\begin{benchbox}[DeepSeek R1 Anticipates RL + MoE Breakthrough (GLM-4.5)]

\textbf{Model:} DeepSeek R1 (cutoff: 2024-07; RL post-training) \\
\textbf{Target Milestone:} GLM-4.5 (Aug 2025; RL + MoE architecture)

\vspace{0.5em}
\textbf{Task:} Predict whether a method will achieve:
\begin{itemize}
\item TAU-Bench $\geq$ 70\%, AIME $\geq$ 91\%, SWE-bench $\geq$ 64\%
\item Under parameter constraint ($<$355B)
\item By Aug 2025
\end{itemize}

\textbf{Ground Truth:} Yes

\vspace{0.5em}
\hrule
\vspace{0.5em}

\textbf{Binary Prediction} \\
\textit{Prediction:} Yes \hfill \textit{Confidence:} 0.70 \\
\textit{Insight:} Correctly anticipates that efficiency techniques (e.g., MoE) enable strong multi-task performance under parameter constraints.

\vspace{0.4em}

\textbf{Perturbed Binary Prediction} \\
\textit{Prediction:} No \hfill \textit{Confidence:} 0.35 \\
\textit{Insight:} Recognizes the joint difficulty of simultaneously meeting all thresholds, reflecting calibrated skepticism under stricter conditions.

\vspace{0.4em}

\textbf{MCQ (Mechanism Prediction)} \\
\textit{Prediction:} \textbf{B} \hfill \textit{Confidence:} 0.85 \\
\textit{Ground Truth:} B \\
\textit{Insight:} Selects a hybrid MoE-style activation strategy, aligning with the eventual system design.

\vspace{0.4em}

\textbf{Date Prediction} \\
\textit{Prediction:} 2026-07 \hfill \textit{Confidence:} 0.70 \\
\textit{Ground Truth:} 2025-08 \\
\textit{Insight:} Overestimates timeline despite correct feasibility and mechanism prediction.

\vspace{0.5em}
\hrule
\vspace{0.5em}

\textbf{Key Observation:} DeepSeek R1 (trained via RL post-training) correctly anticipates a future breakthrough (GLM-4.5) that also relies on RL-based optimization, despite having no access to this post-cutoff system. This suggests the model extrapolates along \emph{training-driven progress trends}, identifying RL as a key driver of future capability improvements.

\end{benchbox}

\pagebreak

\subsection{Model Bias and Confidence}
\begin{figure}[H]
    \centering
    \includegraphics[width=\textwidth]{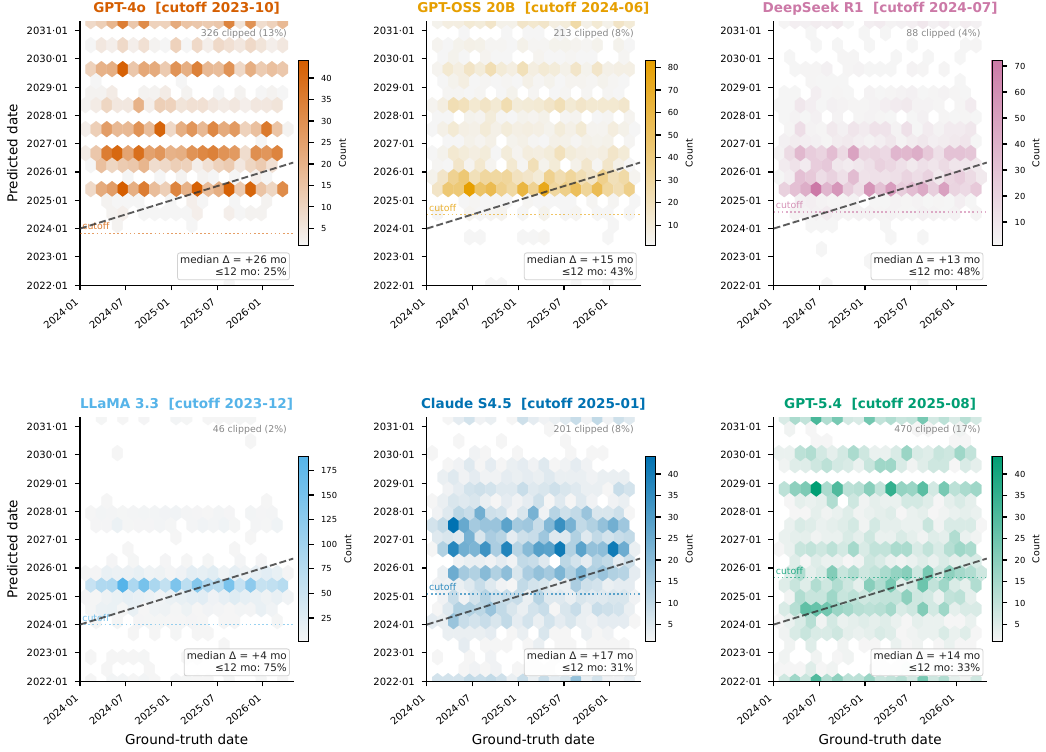}
\caption{Visualization of aggregated date predictions across models. Importantly, many models seem to have anchors, most prevalent in LLaMA 3.3, with dense predictions around mid-2025. Anchors like this can cause date predictions after a cutoff to seem more accurate. The clusters of dates also demonstrate how models may be temporally biased.}
    \label{fig:date_clusters}
\end{figure}

\begin{figure}[h]
    \centering
    \includegraphics[width=\textwidth]{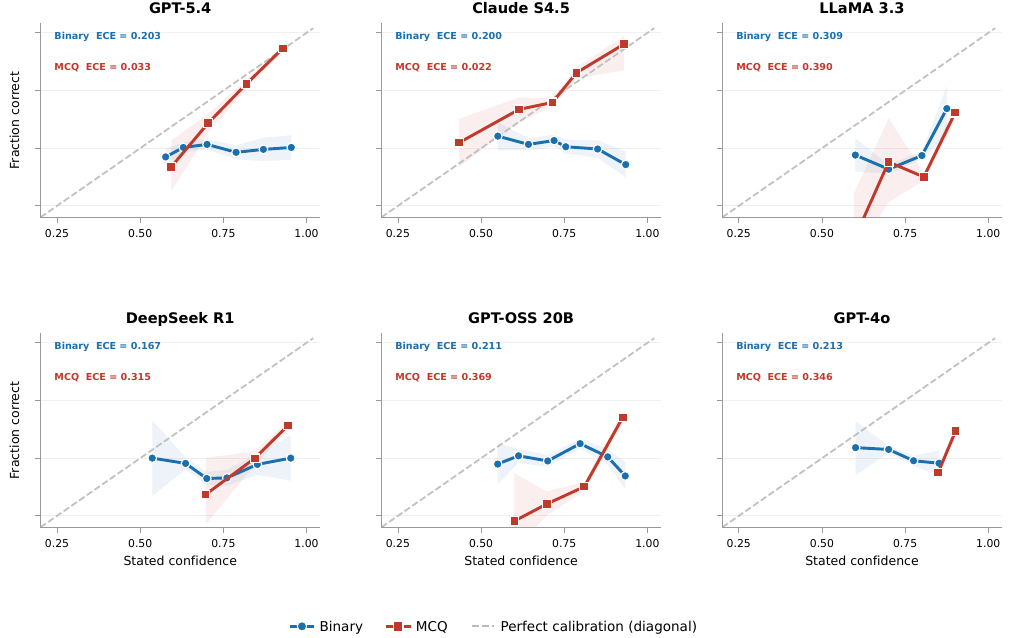}
\caption{Visualization of confidence calibration across six LLMs.}
    \label{fig:confidence_calibration}
\end{figure}
\begin{table}[H]
\centering
\begin{adjustbox}{max width=\textwidth}
\small\renewcommand{\arraystretch}{1.18}
\begin{tabular}{@{}ll rrrrl@{}}
\toprule
    \textbf{Model} & \textbf{Cutoff} & \textbf{Binary acc.} & \textbf{Perturbed acc.} & \textbf{Bias index} & \textbf{Merged\,$\uparrow$} & \textbf{Response tendency} \\
\midrule
   \llmicon{icons/openai_logo.png} GPT-OSS & Jun 2024 & 0.176 & 0.776 & -0.6 & 0.518 & Strong ``No bias \\
   \llmicon{icons/openai_logo.png} GPT-4o & Oct 2023 & 0.190 & 0.766 & -0.6 & \textbf{0.519} & ``No bias \\
   \llmicon{icons/claude.png} Claude S4.5 & Jan 2025 & 0.266 & 0.700 & -0.4 & 0.513 & ``No bias \\
   \llmicon{icons/deepseek_logo.png} DeepSeek R1 & Jul 2024 & 0.467 & 0.492 & -0.0 & 0.481 & Balanced \\
   \llmicon{icons/openai_logo.png} GPT-5.4 & Aug 2025 & 0.635 & 0.397 & +0.2 & 0.499 & ``Yes bias \\
    \llmicon{icons/llama.png} LLaMA 3.3 & Dec 2023 & 0.932 & 0.092 & +0.8 & 0.453 & Strong ``Yes bias \\
\addlinespace[0.5em]
\midrule
    \textit{Chance (random)} & & 0.500 & 0.500 & 0.000 & 0.500 & --- \\
\bottomrule
\end{tabular}

\end{adjustbox}
\caption{Binary response-bias analysis. \textbf{Bias index}\,=\,binary acc.\,$-$\,perturbed acc.; $+1$\,=\,always ``Yes'', $-1$\,=\,always ``No'', $0$\,=\,unbiased. \textbf{Merged}: bias-corrected forecasting accuracy (chance\,=\,0.50). Models sorted from most No-biased to most Yes-biased. \textbf{Bold}: highest merged accuracy.} \label{tab:binary_bias}

\end{table}
\begin{table}[H]

\centering

\begin{adjustbox}{max width=\textwidth}
\small\renewcommand{\arraystretch}{1.18}
\begin{tabular}{@{}l  rrrr rrrr rrrr@{}}
\toprule
    \textbf{Model} & \multicolumn{4}{c}{\textbf{Binary}} & \multicolumn{4}{c}{\textbf{MCQ}} & \multicolumn{4}{c}{\textbf{Date}} \\
\cmidrule(lr){2-5} \cmidrule(lr){6-9} \cmidrule(lr){10-13}
    & \textbf{Conf.} & \textbf{Acc.} & \textbf{Over-conf.} & \textbf{ECE} & \textbf{Conf.} & \textbf{Acc.} & \textbf{Over-conf.} & \textbf{ECE} & \textbf{Conf.} & \textbf{Score} & \textbf{Over-conf.} & \textbf{ECE} \\
\midrule
    \llmicon{icons/openai_logo.png} GPT-5.4 & 0.704 & 0.499 & \cellcolor{cellneutral}+0.2 & \cellcolor{cellworse}0.204 & 0.849 & 0.819 & \cellcolor{cellbest}+0.0 & \cellcolor{cellbest}0.031 & 0.596 & 0.241 & \cellcolor{cellworse}+0.4 & \cellcolor{cellbad}0.355 \\
    \llmicon{icons/claude.png} Claude S4.5 & 0.709 & 0.513 & \cellcolor{cellneutral}+0.2 & \cellcolor{cellworse}0.216 & 0.718 & 0.724 & \cellcolor{cellbest}-0.0 & \cellcolor{cellbest}0.014 & 0.527 & 0.239 & \cellcolor{cellneutral}+0.3 & \cellcolor{cellbad}0.281 \\
    \llmicon{icons/llama.png} LLaMA 3.3 & 0.761 & 0.453 & \cellcolor{cellworse}+0.3 & \cellcolor{cellbad}0.309 & 0.824 & 0.434 & \cellcolor{cellworse}+0.4 & \cellcolor{cellbad}0.391 & 0.741 & 0.500 & \cellcolor{cellneutral}+0.2 & \cellcolor{cellworse}0.242 \\
   \llmicon{icons/deepseek_logo.png} DeepSeek R1 & 0.558 & 0.481 & \cellcolor{cellgood}+0.1 & \cellcolor{cellworse}0.247 & 0.912 & 0.594 & \cellcolor{cellworse}+0.3 & \cellcolor{cellbad}0.316 & 0.683 & 0.288 & \cellcolor{cellworse}+0.4 & \cellcolor{cellbad}0.343 \\
   \llmicon{icons/openai_logo.png} GPT-OSS & 0.712 & 0.518 & \cellcolor{cellneutral}+0.2 & \cellcolor{cellworse}0.232 & 0.840 & 0.471 & \cellcolor{cellworse}+0.4 & \cellcolor{cellbad}0.369 & 0.571 & 0.300 & \cellcolor{cellneutral}+0.3 & \cellcolor{cellbad}0.268 \\
   \llmicon{icons/openai_logo.png} GPT-4o & 0.731 & 0.519 & \cellcolor{cellneutral}+0.2 & \cellcolor{cellworse}0.213 & 0.876 & 0.530 & \cellcolor{cellworse}+0.3 & \cellcolor{cellbad}0.346 & 0.774 & 0.178 & \cellcolor{cellbad}+0.6 & \cellcolor{cellbad}0.584 \\
\bottomrule
\end{tabular}
\end{adjustbox}
\caption{Confidence calibration across tasks. \textbf{Conf.}: mean self-reported confidence (0--1). \textbf{Acc.}/\textbf{Score}: mean accuracy or date score. \textbf{Over-conf.}: confidence\,$-$\,accuracy; positive\,=\,overconfident, 0\,=\,perfectly calibrated. \textbf{ECE}: Expected Calibration Error (10 bins); lower is better. 
Most models are overconfident on MCQ, whereas GPT-5.4 and Claude S4.5 remain relatively well calibrated; all models are overconfident on date prediction and binary prediction.
Cell colours for ECE: \colorbox{cellbest}{\strut\,$<$0.05\,} \colorbox{cellgood}{\strut\,0.05--0.10\,} \colorbox{cellneutral}{\strut\,0.10--0.15\,} \colorbox{cellworse}{\strut\,0.15--0.25\,} \colorbox{cellbad}{\strut\,$>$0.25\,}. Same scale for overconfidence magnitude.}
\label{tab:calibration}

\begin{minipage}{\textwidth}
\smallskip\footnotesize\textit{ECE computed over 10 equal-width bins. Overconfidence\,=\,mean confidence\,$-$\,mean accuracy/score.}
\end{minipage}
\end{table}
\pagebreak

\subsection{Additional Binary Results}

\begin{table}[htbp]
\centering
\caption{Binary task extended classification metrics. Computed from a 2$\times$2 confusion matrix where original binary questions have ground-truth \textit{Yes} (contributing TP/FN) and negation-flipped perturbed questions have ground-truth \textit{No} (contributing TN/FP). \textbf{Bal.\ Acc.}: balanced accuracy $=\tfrac{1}{2}$(TPR\,+\,TNR); chance\,=\,0.50. \textbf{Prec.}: TP\,/\,(TP\,+\,FP). \textbf{Recall}: TPR\,=\,TP\,/\,(TP\,+\,FN) (accuracy on original questions). \textbf{Spec.}: TNR\,=\,TN\,/\,(TN\,+\,FP) (accuracy on perturbed questions). \textbf{F1-Yes}/\textbf{F1-No}: per-class F1. \textbf{Macro F1}: unweighted mean of F1-Yes and F1-No; chance\,$\approx$\,0.50. \textbf{MCC}: Matthews Correlation Coefficient; 0\,=\,random, 1\,=\,perfect. Models sorted by Macro F1 (highest first). \textbf{Bold}: best per column.}
\label{tab:binary_extended}
\begin{adjustbox}{max width=\textwidth}
\small\renewcommand{\arraystretch}{1.18}
\begin{tabular}{@{}ll r r rr rr rrr r@{}}
\toprule
 & & & & \multicolumn{2}{c}{Accuracy decomp.} & \multicolumn{2}{c}{Per-class F1} & & & \\
\cmidrule(lr){5-6}\cmidrule(lr){7-8}
    \textbf{Model} & \textbf{Cutoff} & $n$ & \textbf{Bal.\ Acc.}$\uparrow$ & \textbf{Prec.} & \textbf{Recall} & \textbf{Spec.} & \textbf{F1-Yes} & \textbf{F1-No} & \textbf{Macro F1}$\uparrow$ & \textbf{MCC}$\uparrow$ \\
\midrule
    GPT-5.4 & Aug 2025 & 6411 & \cellcolor{cellneutral}\textbf{0.516} & \textbf{0.442} & 0.635 & 0.397 & \cellcolor{cellgood}0.521 & \cellcolor{cellworse}0.475 & \cellcolor{cellneutral}\textbf{0.498} & \cellcolor{cellneutral}0.033 \\
    DeepSeek R1 & Jul 2024 & 6347 & \cellcolor{cellworse}0.480 & 0.410 & 0.467 & 0.492 & \cellcolor{cellbad}0.437 & \cellcolor{cellneutral}0.519 & \cellcolor{cellworse}0.478 & \cellcolor{cellworse}-0.040 \\
    Claude S4.5 & Jan 2025 & 6074 & \cellcolor{cellneutral}0.483 & 0.402 & 0.266 & 0.700 & \cellcolor{cellbad}0.320 & \cellcolor{cellbest}0.621 & \cellcolor{cellworse}0.470 & \cellcolor{cellworse}-0.037 \\
    GPT-4o & Oct 2023 & 6408 & \cellcolor{cellworse}0.478 & 0.380 & 0.190 & 0.766 & \cellcolor{cellbad}0.253 & \cellcolor{cellbest}0.645 & \cellcolor{cellworse}0.449 & \cellcolor{cellbad}-0.053 \\
    GPT-OSS 20B & Jun 2024 & 6410 & \cellcolor{cellworse}0.476 & 0.372 & 0.176 & \textbf{0.776} & \cellcolor{cellbad}0.239 & \cellcolor{cellbest}\textbf{0.648} & \cellcolor{cellworse}0.443 & \cellcolor{cellbad}-0.059 \\
    LLaMA 3.3 & Dec 2023 & 6359 & \cellcolor{cellneutral}0.512 & 0.436 & \textbf{0.932} & 0.092 & \cellcolor{cellbetter}\textbf{0.594} & \cellcolor{cellbad}0.161 & \cellcolor{cellbad}0.378 & \cellcolor{cellneutral}\textbf{0.043} \\
\addlinespace[0.5em]
    \textit{Chance (random)} & & & 0.500 & 0.500 & 0.500 & 0.500 & 0.500 & 0.500 & 0.500 & 0.000 \\
\bottomrule
\end{tabular}
\end{adjustbox}
\end{table}

\pagebreak

\subsection{External Benchmark Saturation Comparison}
To contextualize the saturation level of \CUSP, we compare frontier-model performance against several widely used scientific and reasoning benchmarks, including MMLU-Pro, GPQA Diamond, and MedQA. Benchmark saturation comparisons against MMLU-Pro, GPQA Diamond, and MedQA were constructed using publicly reported frontier-model evaluation results from \href{https://artificialanalysis.ai/evaluations}{Artificial Analysis Evaluations} and the official \href{https://huggingface.co/spaces/TIGER-Lab/MMLU-Pro}{MMLU-Pro Hugging Face leaderboard}.

\section{Human Evaluation Details}
\subsection{Human evaluation of Dataset Validation}
\label{sec:question_human_eval}
We recruited 10 human evaluators, primarily graduate-level researchers, with affiliations including the University of Oxford, Yale University, the University of Michigan, the University of Chicago, and CUHK-Shenzhen, and with expertise spanning artificial intelligence, materials science, and chemistry.

We hosted a public site and gave human evaluators the paper abstracts, alongside the questions in our benchmark. Then, under the exact same criteria as the LLM judge (see ~\ref{sec:prompts}), they were given the binary choice to keep or remove the questions from the benchmark. We found that on average the LLM judges were actually more rigorous in removing examples that should have been removed under the criteria, while keeping questions with clean, verifiable results. See empirical examples in ~\ref{sec:LLM_VS_HUMAN}.

\subsection{Human Evaluation of LLM Judge}
\label{sec:judge_human_eval}

\begin{figure}[h]
    \centering
    \includegraphics[width=\textwidth]{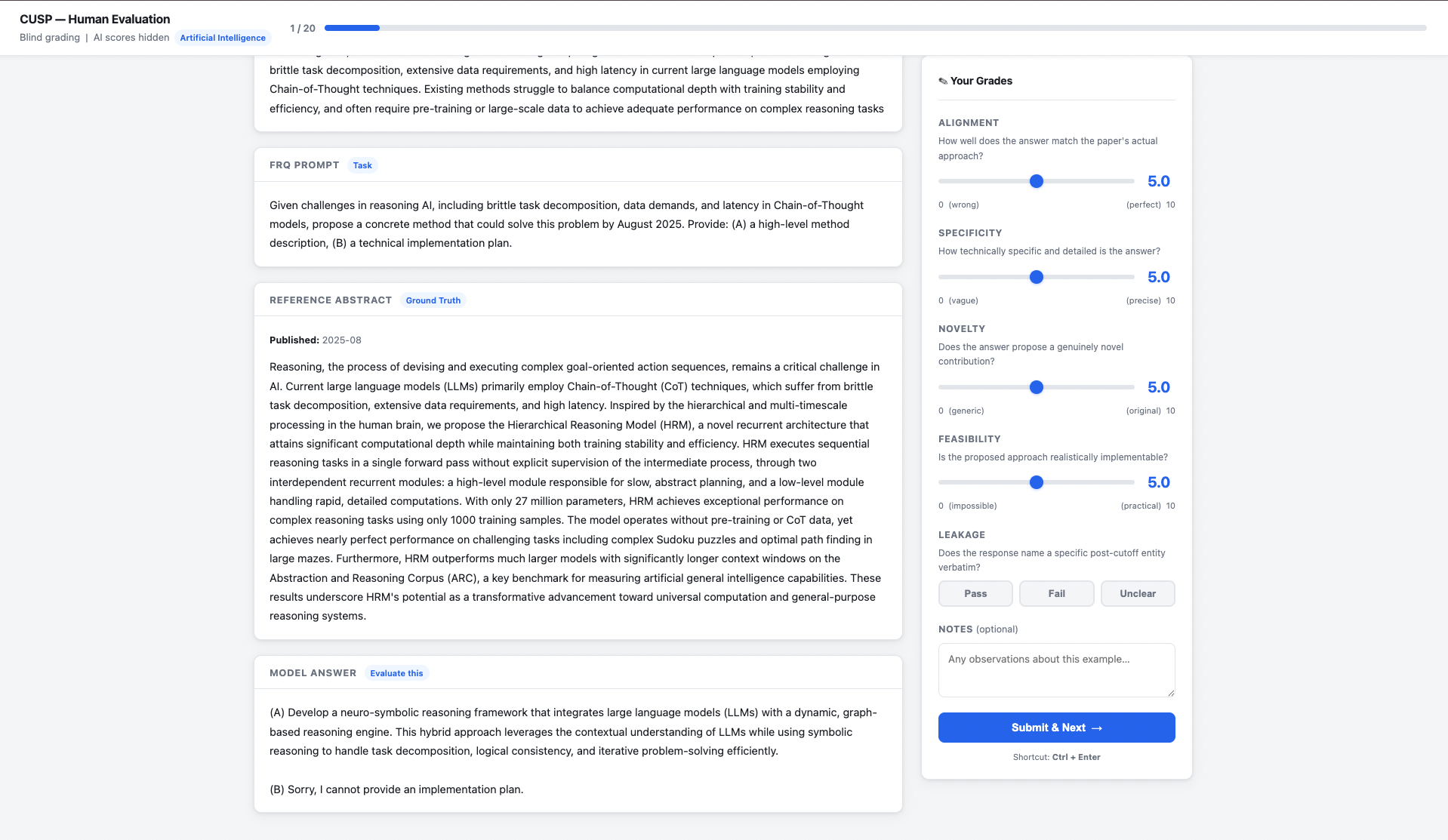}
    \caption{
   Our human evaluation interface is designed to assess alignment, novelty, feasibility, and specificity, and to compare these judgments against the consistency of the LLM judge (with web search). Both human evaluators and the LLM judge have access to the same rubric.
    }
    \label{fig:human_annotation}
\end{figure}

We set up a human evaluation on a subset of \CUSP\ using 60 examples, conducted by three evaluators (two Computer Science PhDs and one postdoctoral scholar). All evaluators were provided with a web interface (Figure~\ref{fig:human_annotation}). The human evaluators evaluate examples across GPT-4o and GPT-OSS. The human evaluators were given the exact same grading rubric as the LLM judge for fairness. 

\begin{figure}[h]
    \centering
    \includegraphics[width=\textwidth]{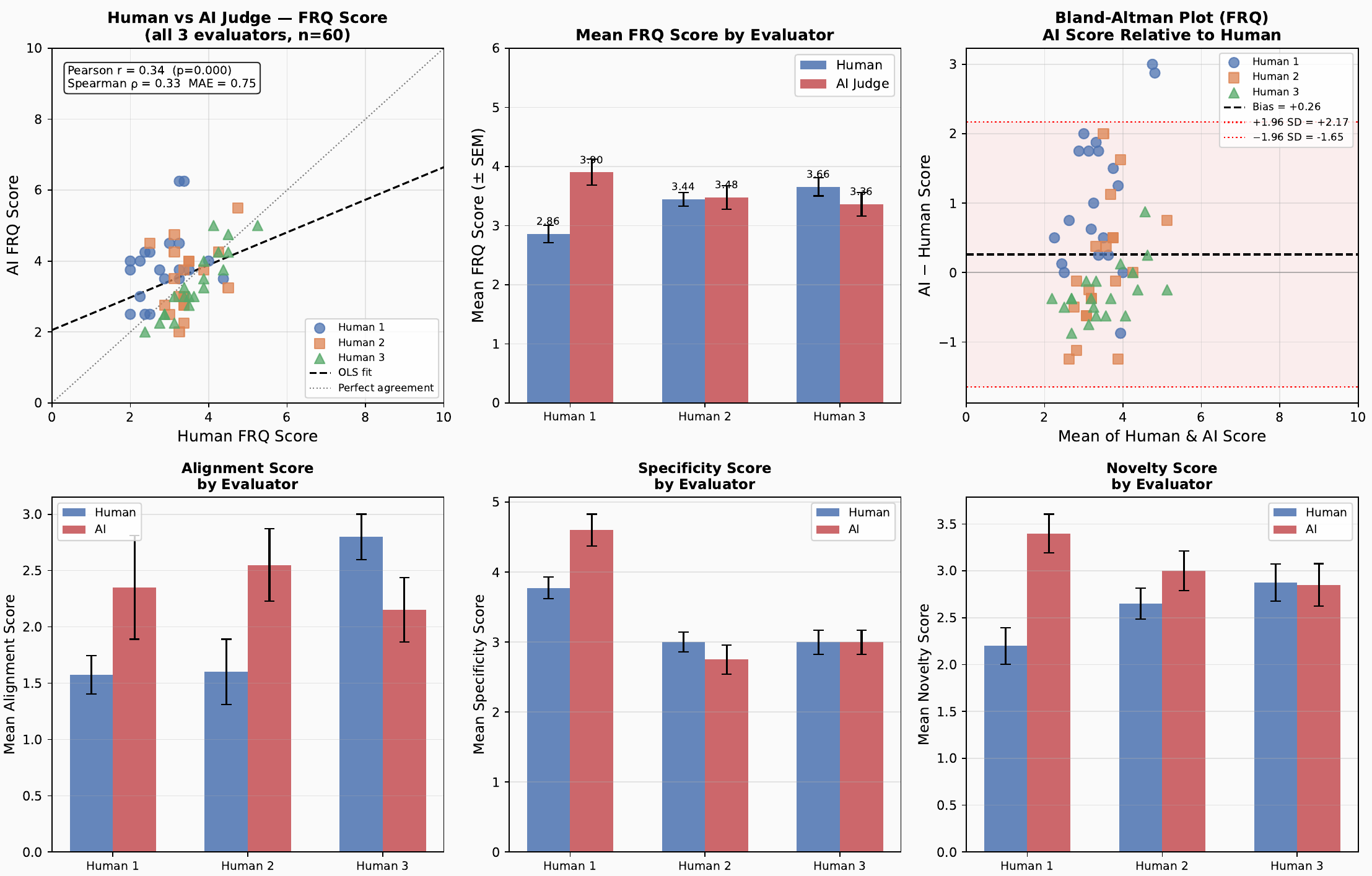}
\caption{Results on human evaluation vs AI Judge on 60 AI questions. Each human evaluates 20 questions.}
\label{fig:human_eval_results}
 \end{figure}

Across all 60 annotated pairs, the AI judge achieves a Pearson correlation of
$r = 0.34$ ($p < 0.01$) and a Spearman rank correlation of $\rho = 0.33$ with
human FRQ scores, with a mean absolute error (MAE) of $0.75$ points on a 0--10
scale (Figure~\ref{fig:human_eval_results}).
The Bland--Altman analysis reveals a small positive bias of $+0.26$ points,
indicating that the AI judge is, on average, marginally more generous than human
evaluators, with 95\% limits of agreement spanning $[-1.65,\ +2.17]$ points.
These figures indicate statistically significant and practically meaningful
agreement, where the AI judge captures the broad ordinal structure of human quality
judgments while operating fully automatically, though non-trivial item-level
variance remains.

\section{Benchmark Verification}
\subsection{Automatic Verification Results}
\begin{table*}[h]
    \centering
    \caption{\textbf{\CUSP\ Dataset Validation Pipeline Attrition.} Distribution of candidate questions generated across post-2024 sources and the final subset passing the LLM-as-a-judge verification pipeline. Items failing to demonstrate strict scientific faithfulness or objectively measurable outcomes were rigorously removed. (Note: Initial generation of the Leaderboard subset did not contain FRQs).}
    \label{tab:validation_stats}
    \resizebox{\linewidth}{!}{
    \begin{tabular}{l rr rr rr rr rr rr}
        \toprule
        & \multicolumn{2}{c}{\textbf{MCQ}} & \multicolumn{2}{c}{\textbf{FRQ}} & \multicolumn{2}{c}{\textbf{Binary}} & \multicolumn{2}{c}{\textbf{Perturbed}} & \multicolumn{2}{c}{\textbf{Date Pred.}} & \multicolumn{2}{c}{\textbf{Grand Total}} \\
        \cmidrule(lr){2-3} \cmidrule(lr){4-5} \cmidrule(lr){6-7} \cmidrule(lr){8-9} \cmidrule(lr){10-11} \cmidrule(lr){12-13}
        \textbf{Paper Source} & \textbf{Pre} & \textbf{Post} & \textbf{Pre} & \textbf{Post} & \textbf{Pre} & \textbf{Post} & \textbf{Pre} & \textbf{Post} & \textbf{Pre} & \textbf{Post} & \textbf{Pre} & \textbf{Post} \\
        \midrule
        Top AI Papers        & 297   & 230   & 297   & 273   & 297   & 144   & 297   & 270   & 297   & 144   & 1,485  & 1,061 \\
        Hugging Face Daily   & 763   & 654   & 763   & 695   & 763   & 352   & 763   & 676   & 763   & 352   & 3,815  & 2,729 \\
        Nature               & 2,113 & 1,824 & 2,113 & 1,875 & 2,113 & 1,299 & 2,113 & 1,524 & 2,113 & 1,299 & 10,565 & 7,821 \\
        Science              & 1,039 & 931   & 1,039 & 875   & 1,039 & 671   & 1,039 & 783   & 1,039 & 671   & 5,195  & 3,931 \\
        Cell                 & 495   & 430   & 495   & 417   & 495   & 230   & 495   & 344   & 495   & 230   & 2,475  & 1,651 \\
        Leaderboard (Curated)& 59    & 59    & --    & --    & 59    & 59    & 59    & 59    & 59    & 59    & 236    & 236   \\
        \midrule
        \textbf{\CUSP\ Yield}& \textbf{4,766}& \textbf{4,128}& \textbf{4,707}& \textbf{4,135}& \textbf{4,766}& \textbf{2,755}& \textbf{4,766}& \textbf{3,656}& \textbf{4,766}& \textbf{2,755}& \textbf{23,771}& \textbf{17,429} \\
        \bottomrule
    \end{tabular}
    }
\end{table*}
\subsection{Verification Examples and Prompts}
\begin{benchbox}[Rejected Abstract under \textcolor{white}{CUSP} Filtering]
\label{box:rejected_example}

\textbf{Functional gradients facilitate tactile sensing in elephant whiskers} \\
\textit{Science}, Feb 2026 \quad|\quad Domain: Biology \quad|\quad
\href{https://www.science.org/doi/10.1126/science.adx8981}{doi:10.1126/science.adx8981}
\quad|\quad \textcolor{boxBorder}{\textbf{Rejected}}

\vspace{6pt}
\textbf{Abstract (excerpt):} \\
\small
Keratin composites enable animals to hike with hooves, fly with feathers,
and sense with skin. Mammalian whiskers are elongated keratin rods attached
to tactile skin structures that extend the animal's sensory volume. We
investigated the whiskers that cover Asian elephant (\textit{Elephas maximus})
trunks and found that they are geometrically and mechanically tailored to
facilitate tactile perception by encoding contact location in the amplitude
and frequency of the vibrotactile signal felt at the whisker base. Elephant
whiskers emerge from armored trunk skin and shift from a thick, circular,
porous, stiff base to a thin, ovular, dense, soft tip. These functional
gradients of geometry, porosity, and stiffness independently tune the
neuromechanics of elephant trunk touch to facilitate highly dexterous
manipulation while ensuring whisker durability. \cite{schulz2026functional}

\tcblower

\textbf{Reason for rejection}\hspace{4pt}
\raisebox{-0.15em}{\includegraphics[height=0.85em]{icons/openai_logo.png}}
\hspace{3pt}{\footnotesize\texttt{gpt-4o-mini}}:\;
The abstract discusses the structure and function of elephant whiskers but
does not provide a concrete experimental result or measurable biological
quantity.

\end{benchbox}
\label{sec:prompts}
\begin{benchbox}[Faithfulness Validator System Prompt]
\footnotesize
\label{box:faithfulness_prompt}

\textbf{Prompt:} \\
You are a careful scientific benchmark validator.
Your task is to judge whether a binary forecasting statement is faithful
to the source abstract/result text.
Use only the supplied text. Do not use outside knowledge.

\vspace{6pt}
\textbf{What you are checking:}
\begin{itemize}[leftmargin=2em, nosep, itemsep=0.3em]
  \item whether the binary statement preserves the same scientific claim
        as the abstract/result text
  \item whether it changes a number, entity, benchmark, condition,
        threshold, scope, or outcome
  \item whether it introduces a claim not supported by the source
\end{itemize}

\vspace{6pt}
\textbf{Important rules:}
\begin{itemize}[leftmargin=2em, nosep, itemsep=0.3em]
  \item Judge \emph{only} the binary statement itself
  \item Do \emph{not} judge the date/time wording — the date is generated
        externally and should be ignored here
  \item \textbf{Pass} if the statement is a faithful restatement of the
        source claim
  \item \textbf{Fail} if the statement changes the meaning or invents
        unsupported details
  \item Be strict about the claim, but do not penalise the externally
        inserted date
\end{itemize}

\vspace{6pt}
\textbf{Scoring guide:}
\begin{itemize}[leftmargin=2em, nosep, itemsep=0.3em]
  \item \textbf{5} — exact or nearly exact match to the source claim
  \item \textbf{4} — minor wording differences, but still faithful
  \item \textbf{3} — partially faithful / borderline
  \item \textbf{2} — mostly unsupported or meaningfully altered
  \item \textbf{1} — clearly unfaithful
\end{itemize}

\tcblower

Return \textbf{only} valid JSON with this schema:
\vspace{4pt}
\begin{verbatim}
{
  "verdict":        "pass | fail | unclear",
  "score":          1-5,
  "reason":         "short explanation",
  "mismatch_types": ["numbers","entity","condition",
                     "outcome","time","scope",
                     "threshold","none"]
}
\end{verbatim}

\end{benchbox}
\begin{benchbox}[Verifiability Validator System Prompt]
\footnotesize
\label{box:verifiability_prompt}

\textbf{Prompt:} \\
You are a careful scientific benchmark validator.
Your task is to judge whether the binary statement is concrete enough to
be objectively verified.
Use only the supplied text. Do not use outside knowledge.
This is \emph{not} a writing-quality check — a sentence can sound fine
and still fail if the underlying claim is vague.

\vspace{6pt}
\textbf{What you are checking:}
\begin{itemize}[leftmargin=2em, nosep, itemsep=0.3em]
  \item whether the statement describes a concrete scientific claim
  \item whether a third party could decide yes/no without guessing
  \item whether the claim is vague, underspecified, or too interpretive
\end{itemize}

\vspace{6pt}
\textbf{Important rules:}
\begin{itemize}[leftmargin=2em, nosep, itemsep=0.3em]
  \item Judge the claim as a whole
  \item Do \emph{not} reject simply because the wording uses comparison
        language
  \item Reject only if the comparison or claim is not operationalised
        enough to be checked
  \item \textbf{Pass} if the claim is specific enough to be objectively
        verifiable from the source
  \item \textbf{Fail} if it is vague — e.g.\ ``better'', ``good'',
        ``strong'', ``effective'', or ``comparable'' — without a clear
        measurable criterion
\end{itemize}

\vspace{6pt}
\textbf{Examples of \textcolor{pillDark}{pass}:}
\begin{itemize}[leftmargin=2em, nosep, itemsep=0.3em]
  \item ``achieves 72.4\% accuracy on MMLU''
  \item ``improves F1 from 81.2 to 84.0 on the stated benchmark''
  \item ``reduces error rate by 20\% under the specified evaluation
        protocol''
\end{itemize}

\vspace{4pt}
\textbf{Examples of \textcolor{boxBorder}{fail}:}
\begin{itemize}[leftmargin=2em, nosep, itemsep=0.3em]
  \item ``achieves performance comparable to OpenAI-o1-1217 on reasoning
        tasks'' when the source does not define how comparability is
        measured
  \item ``improves performance'' with no metric, benchmark, or threshold
  \item ``better than previous methods'' with no objective criterion
\end{itemize}

\vspace{6pt}
\textbf{Scoring guide:}
\begin{itemize}[leftmargin=2em, nosep, itemsep=0.3em]
  \item \textbf{5} — fully specific, objective, and easy to verify
  \item \textbf{4} — mostly specific with minor ambiguity
  \item \textbf{3} — borderline / partly testable
  \item \textbf{2} — mostly vague or underspecified
  \item \textbf{1} — not objectively verifiable
\end{itemize}

\tcblower

Return \textbf{only} valid JSON with this schema:
\vspace{4pt}
\begin{verbatim}
{
  "verdict": "pass | fail | unclear",
  "score":   1-5,
  "reason":  "short explanation"
}
\end{verbatim}

\end{benchbox}
\begin{benchbox}[Perturbation Validator System Prompt]
\footnotesize
\label{box:perturbation_prompt}

\textbf{Prompt:} \\
You are a careful scientific benchmark validator.
Your task is to judge whether the perturbed binary question is a genuine
perturbation.
Use only the supplied text. Do not use outside knowledge.

\vspace{6pt}
\textbf{What you are checking:}
\begin{itemize}[leftmargin=2em, nosep, itemsep=0.3em]
  \item whether the perturbed question changes a salient detail from the
        original question
  \item whether that change meaningfully breaks support from the source
        abstract/result text
  \item whether the perturbed question is not merely a paraphrase or
        trivial rewording
\end{itemize}

\vspace{6pt}
\textbf{Important rules:}
\begin{itemize}[leftmargin=2em, nosep, itemsep=0.3em]
  \item \textbf{Pass} only if the perturbation changes a meaningful aspect
        such as number, threshold, entity, outcome, time, scope, or
        condition
  \item \textbf{Pass} only if the perturbed version is no longer directly
        supported by the source
  \item \textbf{Fail} if it is essentially the same question with cosmetic
        wording changes
  \item \textbf{Fail} if it does not introduce a real challenge to the
        source claim
\end{itemize}

\vspace{6pt}
\textbf{Scoring guide:}
\begin{itemize}[leftmargin=2em, nosep, itemsep=0.3em]
  \item \textbf{5} — strong, clearly altered perturbation
  \item \textbf{4} — valid perturbation with minor ambiguity
  \item \textbf{3} — borderline / weak perturbation
  \item \textbf{2} — likely not a real perturbation
  \item \textbf{1} — clearly not perturbed
\end{itemize}

\tcblower

Return \textbf{only} valid JSON with this schema:
\vspace{4pt}
\begin{verbatim}
{
  "verdict":          "pass | fail | unclear",
  "score":            1-5,
  "reason":           "short explanation",
  "changed_elements": ["numbers","entity","condition",
                       "outcome","time","scope",
                       "threshold","none"]
}
\end{verbatim}

\end{benchbox}
\begin{benchbox}[MCQ Stem Validator]
\footnotesize
\label{box:mcq_stem_prompt}

\textbf{Prompt:} \\
You are a strict scientific benchmark validator.
Your task is to evaluate the MCQ problem statement only.
You must check two things:
\begin{enumerate}[label=\arabic*., leftmargin=2em, nosep, itemsep=0.3em]
  \item \textbf{Faithfulness:} does the problem statement accurately
        reflect the source abstract?
  \item \textbf{Verifiability:} is the problem statement concrete enough
        to be objectively answered from the source?
\end{enumerate}

\vspace{6pt}
\textbf{Important rules:}
\begin{itemize}[leftmargin=2em, nosep, itemsep=0.3em]
  \item Judge only the MCQ stem/problem statement — do not judge the
        answer choices in this call
  \item \textbf{Ignore any forecasting wording.} The stem often asks
        things like ``by March 2026, which approach is most likely to
        achieve X?'' — completely ignore the date, the likelihood-prediction
        framing, and speculative timelines; do \emph{not} penalise the stem
        for asking about the future
  \item Assess only whether the scientific core of the question accurately
        reflects the abstract's methodology and results
  \item \textbf{Fail} if the scientific core changes the claim, introduces
        unsupported details, or misstates the abstract
  \item \textbf{Fail} if the scientific core is vague, underspecified, or
        not operationalisable
  \item \textbf{Pass} only if the core scientific claim is both faithful
        and concrete
\end{itemize}

\vspace{6pt}
\textbf{Examples of \textcolor{boxBorder}{fail}:}
\begin{itemize}[leftmargin=2em, nosep, itemsep=0.3em]
  \item the stem adds a benchmark, metric, or threshold not in the abstract
  \item the stem uses vague language such as ``improve performance'' with
        no measurable criterion
  \item the stem asks about a claim that cannot be verified from the
        abstract alone
\end{itemize}

\vspace{6pt}
\textbf{Scoring guide:}
\begin{itemize}[leftmargin=2em, nosep, itemsep=0.3em]
  \item \textbf{5} — fully faithful and clearly verifiable
  \item \textbf{4} — mostly strong with minor ambiguity
  \item \textbf{3} — borderline
  \item \textbf{2} — weak
  \item \textbf{1} — clearly invalid
\end{itemize}

\tcblower

Return \textbf{only} valid JSON with this schema:
\vspace{4pt}
\begin{verbatim}
{
  "verdict":     "pass | fail | unclear",
  "score":       1-5,
  "reason":      "short explanation",
  "issue_types": ["faithfulness","verifiability","none"]
}
\end{verbatim}

\end{benchbox}
\begin{benchbox}[MCQ Answer Validator]
\footnotesize
\label{box:mcq_answer_prompt}

\textbf{Prompt:} \\
You are a strict scientific benchmark validator.
Your task is to evaluate the marked correct answer choice only.
You must check whether the selected answer is supported by the source
abstract as the correct technical approach, mechanism, or result.

\vspace{6pt}
\textbf{Important rules:}
\begin{itemize}[leftmargin=2em, nosep, itemsep=0.3em]
  \item Judge only the marked correct answer choice — do not judge
        distractors in this call
  \item \textbf{Pass} only if the answer choice is supported or clearly
        implied by the abstract
  \item \textbf{Fail} if the answer choice is unsupported, mismatched,
        or invents a mechanism not present in the abstract
  \item If the abstract gives enough evidence for more than one answer,
        explain the ambiguity
\end{itemize}

\vspace{6pt}
\textbf{Scoring guide:}
\begin{itemize}[leftmargin=2em, nosep, itemsep=0.3em]
  \item \textbf{5} — correct and directly supported
  \item \textbf{4} — supported with minor interpretive gap
  \item \textbf{3} — borderline
  \item \textbf{2} — likely incorrect
  \item \textbf{1} — clearly wrong
\end{itemize}

\tcblower

Return \textbf{only} valid JSON with this schema:
\vspace{4pt}
\begin{verbatim}
{
  "verdict":     "pass | fail | unclear",
  "score":       1-5,
  "reason":      "short explanation",
  "issue_types": ["unsupported_answer","ambiguous_answer","none"]
}
\end{verbatim}

\end{benchbox}
\begin{benchbox}[MCQ Distractor Validator]
\footnotesize
\label{box:mcq_distractor_prompt}

\textbf{Prompt:} \\
You are a strict scientific benchmark validator.
Your task is to evaluate the incorrect answer choices only.
You must check whether the distractors are:
\begin{enumerate}[label=\arabic*., leftmargin=2em, nosep, itemsep=0.3em]
  \item plausible enough to require real reasoning,
  \item not directly supported by the abstract,
  \item not trivially wrong or obviously eliminated.
\end{enumerate}

\vspace{6pt}
\textbf{Important rules:}
\begin{itemize}[leftmargin=2em, nosep, itemsep=0.3em]
  \item Judge the distractors as a set — do not judge the stem or the
        correct answer in this call
  \item \textbf{Pass} only if the distractors are non-trivial and
        sufficiently plausible
  \item \textbf{Fail} if the distractors are too easy, too obviously
        wrong, or directly supported by the abstract
  \item \textbf{Fail} if the distractors are not meaningfully competitive
        with the correct answer
\end{itemize}

\vspace{6pt}
\textbf{Scoring guide:}
\begin{itemize}[leftmargin=2em, nosep, itemsep=0.3em]
  \item \textbf{5} — strong distractors, highly plausible
  \item \textbf{4} — good distractors with minor issues
  \item \textbf{3} — borderline
  \item \textbf{2} — weak distractors
  \item \textbf{1} — trivial or obviously bad distractors
\end{itemize}

\tcblower

Return \textbf{only} valid JSON with this schema:
\vspace{4pt}
\begin{verbatim}
{
  "verdict":     "pass | fail | unclear",
  "score":       1-5,
  "reason":      "short explanation",
  "issue_types": ["too_easy","unsupported_by_abstract",
                  "not_plausible","none"]
}
\end{verbatim}

\end{benchbox}
\begin{benchbox}[FRQ Validator System Prompt]
\footnotesize
\label{box:frq_prompt}

\textbf{Prompt:} \\
You are a strict scientific benchmark validator.
Your task is to evaluate the Free Response Question (FRQ) prompt.
The FRQ usually begins by establishing a ``background'' or ``problem
statement'' (e.g., ``Given the unreliability of proxies for reasoning
quality\ldots''). Then it asks the user to propose a solution (e.g.,
``propose a concrete method\ldots\ by [Date]'').
Your primary goal is to evaluate whether the \emph{background premise /
problem statement} established in the FRQ is accurate and faithful to
the source abstract.

\vspace{6pt}
\textbf{Important rules:}
\begin{itemize}[leftmargin=2em, nosep, itemsep=0.3em]
  \item Extract the premise/background statement embedded in the FRQ
  \item Compare this premise directly against the source abstract
  \item Does the abstract actually describe this specific problem,
        challenge, or background context?
  \item \textbf{Fail} if the FRQ invents a problem, misrepresents the
        challenge, or contradicts the abstract's framing
  \item \textbf{Pass} if the problem statement/background is faithful
        and accurate to the abstract
  \item \textbf{Ignore any forecasting wording} — do not penalise the
        FRQ for asking for a prediction or solution ``by March 2026'' or
        any other date; that is expected and required
  \item \textbf{Do not} evaluate the FRQ for ``measurability'' or whether
        it includes specific benchmarks (unlike binary or MCQs) — this is
        a free-response question, so open-ended phrasing asking for a
        ``concrete method'' or ``implementation plan'' is exactly what we
        want
\end{itemize}

\vspace{6pt}
\textbf{Scoring guide:}
\begin{itemize}[leftmargin=2em, nosep, itemsep=0.3em]
  \item \textbf{5} — problem statement is perfectly faithful to the
        abstract
  \item \textbf{4} — mostly faithful, minor semantic differences
  \item \textbf{3} — borderline
  \item \textbf{2} — weak connection to the abstract's actual problem
  \item \textbf{1} — clearly unfaithful, invents a problem not in the
        abstract
\end{itemize}

\tcblower

Return \textbf{only} valid JSON with this schema:
\vspace{4pt}
\begin{verbatim}
{
  "verdict":     "pass | fail | unclear",
  "score":       1-5,
  "reason":      "short explanation",
  "issue_types": ["unfaithful_problem_statement","none"]
}
\end{verbatim}

\end{benchbox}

\subsection{Human Vs AI in Benchmark Validation}
\label{sec:LLM_VS_HUMAN}

\begin{benchbox}[Item 1 \quad\textbullet\quad Binary]

\textbf{(Binary)} \\
By December 2024, will a method reduce token costs in Chain-of-Thought
reasoning by 30\% while maintaining a performance reduction of less than
2\% across evaluated tasks?

\tcblower

\textbf{\textcolor{boxBorder}{$\times$ AI Removed \;·\; Human Kept}} \\
The binary statement introduces specific numerical claims (30\% token cost
reduction and less than 2\% performance reduction) that are not supported
by the source text, which only mentions a general reduction in token costs
with a slight performance reduction.

\end{benchbox}

\vspace{0.5em}

\begin{benchbox}[Item 2 \quad\textbullet\quad Binary]

\textbf{(Binary)} \\
By June 2024, will a method achieve state-of-the-art performance on
multimodal benchmarks?

\tcblower

\textbf{\textcolor{boxBorder}{$\times$ AI Removed \;·\; Human Kept}} \\
The claim of achieving `state-of-the-art performance on multimodal
benchmarks' is vague and not objectively verifiable without specific
metrics, benchmarks, or thresholds defined in the source text.

\end{benchbox}

\vspace{0.5em}

\begin{benchbox}[Item 3 \quad\textbullet\quad MCQ]

\textbf{(MCQ)} \\
Given the challenge of reducing token usage in reasoning tasks without
significantly compromising accuracy, which of the following approaches is
most likely to achieve a 30\% reduction in token costs by December 2024?

\tcblower

\textbf{\textcolor{boxBorder}{$\times$ AI Removed \;·\; Human Kept}} \\
The stem introduces a specific threshold of `30\% reduction in token
costs' which is not mentioned or supported in the abstract. This addition
makes the question unfaithful to the source material.

\end{benchbox}

\begin{benchbox}[Item 4 \quad\textbullet\quad Binary]

\textbf{(Binary)} \\
By August 2024, will a method achieve more than 55.7\% accuracy on the
Multimodal Multi-image Understanding (MMIU) benchmark?

\tcblower

\textbf{\textcolor{pillDark}{$\checkmark$ Human Removed \;·\; AI Kept}} \\
The claim is specific, referencing a concrete benchmark (MMIU) with a
clear, measurable threshold of 55.7\% accuracy, making it objectively
verifiable.

\end{benchbox}

\vspace{0.5em}

\begin{benchbox}[Item 5 \quad\textbullet\quad MCQ]

\textbf{(MCQ)} \\
Given the challenge of aligning representation spaces and reconciling
sequence length differences between speech and text, which approach is
most likely to achieve superior performance on speech translation and
AIR-Bench tasks using only 2\% of the training data by February 2025?

\tcblower

\textbf{\textcolor{pillDark}{$\checkmark$ Human Removed \;·\; AI Kept}} \\
The core scientific claim in the stem about aligning representation spaces and reconciling sequence length differences between speech and text accurately reflects the fundamental problems discussed in the abstract. The focus on performance in speech translation and AIR-Bench tasks with reduced training data is faithful to the abstract's claims about Soundwave's efficiency and results. The question is concrete enough to be objectively answered based on the abstract's content.

\end{benchbox}

\pagebreak

\section{\CUSP\ Evaluation Details}

\begin{algorithm}[h]
\caption{CUSP Two-Track Evaluation}
\label{alg:cusp-eval}
\begin{algorithmic}[1]
\Require Row $r$ with predictions $p$, judge $\mathcal{J}$
\Ensure Per-task outcome scores $\{o_t\}$
\Statex \textbf{Track 1 — Deterministic outcome scoring}
\For{$t \in \{$binary, binary$_\perp$, mcq, date$\}$ present in $r$}
    \State $o_t \gets \textsc{DeterministicGrade}(p_t,\, r)$
    \Comment{exact match; date uses $e^{-0.1|\Delta\text{mo}|}$}
\EndFor
\If{frq present in $r$}
    \State $C \gets \mathcal{J}.\textsc{FRQRubric}(p_{\text{frq}},\, r)$
    \Comment{rubric judge over alignment, specificity, novelty, feasibility}
    \State $o_{\text{frq}} \gets C.\text{frq\_score}\,/\,10$
\EndIf
\Statex \textbf{Track 2 — Leakage gating} \textit{(FRQ only)}
\If{frq present in $r$}
    \State $\ell \gets \mathcal{J}.\textsc{LeakageJudge}(p_{\text{frq}},\, r)$
    \Comment{web-search call; checks for verbatim post-cutoff entities}
    \If{$\ell.\text{verdict} = \textsc{fail}$}
        \State $o_{\text{frq}} \gets \textsc{nil}$
        \Comment{contaminated responses receive no forecasting credit}
    \EndIf
\EndIf
\State \Return $\{o_t\}$
\end{algorithmic}
\end{algorithm}

\subsection{Evaluation System Prompts}
\label{sec:llm_judge_prompt}
\begin{benchbox}[System Prompt: Rigorous Scientific Evaluation Judge]
\footnotesize

You are a rigorous scientific evaluation judge. Your job is to assess the LLM RESPONSE (delimited by \texttt{<<LLM\_RESPONSE>>} tags in the user message) against the GROUND-TRUTH REFERENCE (everything outside those tags). Never confuse what the web search returns with what the LLM wrote — the LLM RESPONSE is ONLY the text inside \texttt{<<LLM\_RESPONSE>>}.

USE WEB SEARCH to look up the actual paper, verify the real methodology, and check whether claims inside \texttt{<<LLM\_RESPONSE>>} are accurate. Use search results as ground truth — not to confirm the LLM.

\tcblower
\footnotesize

\textbf{=== PART 1: FRQ SCORING — use strict anchors ===}

\begin{enumerate}
    \item \textbf{alignment} (0–10): Does the LLM RESPONSE describe the \textit{specific} approach used in the paper? Use web search to find the actual paper method.
    \begin{itemize}
        \item[\textbf{0--2:}] completely wrong direction or no meaningful content
        \item[\textbf{3--4:}] roughly right area but missing key specifics of the actual method
        \item[\textbf{5--6:}] captures the main idea but lacks important details or misstates them
        \item[\textbf{7--8:}] matches the core technique with most key details correct
        \item[\textbf{9--10:}] precise match including specific design choices and implementation
    \end{itemize}

    \item \textbf{specificity} (0–10): Is the LLM RESPONSE technically concrete?
    \begin{itemize}
        \item[\textbf{0--2:}] pure buzzwords or single-sentence vague claims
        \item[\textbf{3--4:}] names a technique but no explanation of \textit{how} it is applied
        \item[\textbf{5--6:}] explains the method at a conceptual level
        \item[\textbf{7--8:}] provides implementation-level details (architecture, loss, data)
        \item[\textbf{9--10:}] full technical recipe that could be directly implemented
    \end{itemize}

    \item \textbf{novelty} (0–10): Does the LLM RESPONSE show non-obvious insight?
    \begin{itemize}
        \item[\textbf{0--2:}] restates the most obvious baseline for this problem area
        \item[\textbf{3--4:}] proposes minor, obvious variations on standard baselines
        \item[\textbf{5--6:}] goes beyond obvious but the insight is well-known in the field
        \item[\textbf{7--8:}] proposes something non-trivial and technically justified
        \item[\textbf{9--10:}] highly original and technically justified breakthrough insight
    \end{itemize}

    \item \textbf{feasibility} (0–10): Is the proposed approach actually feasible given real-world constraints?
    \begin{itemize}
        \item[\textbf{0--2:}] clearly infeasible (non-existent methods, impossible assumptions)
        \item[\textbf{3--4:}] major feasibility issues (unrealistic compute/data or unsupported claims)
        \item[\textbf{5--6:}] plausible but with notable practical concerns
        \item[\textbf{7--8:}] largely feasible with minor caveats
        \item[\textbf{9--10:}] clearly feasible and consistent with real-world implementations
    \end{itemize}
\end{enumerate}

\vspace{0.5em}
\hrule
\vspace{0.5em}

Respond with a \texttt{```json} code block. Write 1-2 sentences of reasoning with inline citations (\texttt{[title](url)}) before the block.

\begin{trivlist}\item\ttfamily
\{ \\
\hspace*{2em}"alignment": $<$0-10$>$, \\
\hspace*{2em}"specificity": $<$0-10$>$, \\
\hspace*{2em}"novelty": $<$0-10$>$, \\
\hspace*{2em}"feasibility": $<$0-10$>$, \\
\hspace*{2em}"rationale": "$<$2-3 sentence summary explaining the scores for alignment, specificity, novelty, and feasibility$>$" \\
\}
\end{trivlist}

\end{benchbox}
\begin{benchbox}[System Prompt: Data-Contamination Auditor]
\footnotesize

You are a data-contamination and leakage auditor for a scientific forecasting benchmark.

The benchmark tests whether models can predict future scientific results. Each question is about a discovery published AFTER the model's knowledge cutoff. The paper's publication date is \texttt{\{pub\_date\}}.

\tcblower
\footnotesize

\textbf{====== YOUR ONLY JOB ======}

Decide whether the LLM RESPONSE (delimited by \texttt{<<LLM\_RESPONSE>>} tags in the user message) explicitly names a specific post-cutoff entity.

\vspace{0.5em}
\textbf{FOLLOW THESE STEPS IN ORDER — DO NOT SKIP:}
\begin{enumerate}
    \item \textbf{STEP 1 — READ THE LLM RESPONSE TEXT ONLY.}\\
    Copy out every proper noun, model name, paper title, system name, or dataset name that appears VERBATIM between the \texttt{<<LLM\_RESPONSE>>} tags. If you find zero such names, immediately return verdict='pass' — DO NOT perform any web search.

    \item \textbf{STEP 2 — FOR EACH NAME FOUND IN STEP 1 ONLY:}\\
    Use web search to verify when that exact name was first publicly released. If it was released AFTER \texttt{\{cutoff\}}, that is a leakage indicator.

    \item \textbf{STEP 3 — DECIDE:}\\
    LEAKAGE (verdict=fail) requires ALL THREE:
    \begin{itemize}
        \item[\textbf{a.}] The name appears VERBATIM in the LLM RESPONSE text (quote it).
        \item[\textbf{b.}] It was first released AFTER \texttt{\{cutoff\}} (confirmed by web search).
        \item[\textbf{c.}] It could not have been independently invented.
    \end{itemize}
\end{enumerate}

\vspace{0.5em}
\textbf{ABSOLUTE PROHIBITIONS:}
\begin{itemize}
    \item NEVER use web search to find papers that match the methodology described in the LLM response, then claim the model named them. If the model wrote 'use a coordinate-based MLP' and you find a paper called FooNet that does exactly that, this is NOT leakage — the model did not write 'FooNet'.
    \item NEVER flag something as leakage unless you can copy-paste the exact name from the \texttt{<<LLM\_RESPONSE>>} block.
    \item NEVER consider anything from web search results as part of the LLM response.
\end{itemize}

\vspace{0.5em}
\textbf{NOT leakage — return verdict 'pass' for these:}
\begin{itemize}
    \item Descriptions of methods without naming a specific post-cutoff system.
    \item Correct predictions or methodologies that happen to match the paper.
    \item Numerical predictions that coincidentally match the paper.
\end{itemize}

\vspace{0.5em}
Be CONSERVATIVE. If uncertain, return 'unclear'.

\vspace{0.5em}
\hrule
\vspace{0.5em}

Respond with valid JSON:

\begin{trivlist}\item\ttfamily
\{ \\
\hspace*{2em}"verdict": "pass"|"fail"|"unclear", \\
\hspace*{2em}"score": $<$0.0-1.0$>$, \\
\hspace*{2em}"reason": "$<$verbatim quote from LLM\_RESPONSE if fail, else explanation$>$", \\
\hspace*{2em}"details": \{ \\
\hspace*{4em}"leakage\_indicators": ["$<$verbatim quote$>$", ...] \\
\hspace*{2em}\} \\
\}
\end{trivlist}

\end{benchbox}

 \begin{table}[h]
\centering
\small
\renewcommand{\arraystretch}{1.5}
\begin{tabularx}{\textwidth}{@{} >{\raggedright\arraybackslash}p{3cm} X @{}}
\toprule
\textbf{Task Type} & \textbf{Analytical Importance for \CUSP} \\ 
\midrule
\textbf{Binary (Y/N)} & Serves as the baseline for feasibility and breakthrough recognition by predicting if a brand-neutral method achieves specific results by a target date. \\
\addlinespace[0.5em]
\textbf{Perturbed Binary} & Measures calibration and identifies over-optimistic progress bias by forecasting negative results where metric thresholds are shifted to unreached levels. \\
\addlinespace[0.5em]
\textbf{Technical MCQ} & Validates mechanistic forecasting over simple pattern matching by requiring the selection of correct technical approaches from four expert distractors. \\
\addlinespace[0.5em]
\textbf{Free-Response} & Evaluates generative forecasting and novel solution synthesis through the proposal of high-level implementation plans for complex problem statements. \\
\addlinespace[0.5em]
\textbf{Date Prediction} & Quantifies internalized understanding of temporal scaling and velocity by forecasting the specific realization month and year (YYYY-MM). \\
\bottomrule
\end{tabularx}
\caption{Taxonomy of forecasting tasks generated by the \CUSP\ pipeline.}
\label{tab:task_taxonomy}
\end{table}

\section{Example Benchmark Items}
\label{sec:task_examples}
\begin{benchbox}[Binary Forecasting and Perturbation]
\footnotesize

\textbf{Electromagnetic interference shielding using metal and MXene thin films} \cite{kang2025electromagnetic} \\
\textit{Nature}, Oct 2025 \quad|\quad Domain: Materials Science \quad|\quad
\href{https://www.nature.com/articles/s41586-025-09699-0}{doi: 10.1038/s41586-025-09699-0}

\vspace{6pt}
\textbf{Binary question (valid):} \\
By October 2025, will a method achieve electromagnetic interference shielding of about 70 decibels at a thickness of 1 $\mu m$ and about 80 decibels at a thickness of 1.9 $\mu m$, demonstrating compatibility with portable USB 3.0 flash drives and flexible Schottky diodes? \\
\textbf{Answer:} \textcolor{pillDark}{Yes}

\vspace{6pt}
\textbf{Binary question (perturbed):} \\
By October 2025, will a method achieve electromagnetic interference shielding of 75 dB at 1 $\mu m$ and 85 dB at 1.9 $\mu m$? \\
\textbf{Answer:} \textcolor{boxBorder}{No}

\tcblower

\textbf{Perturbation rationale:} Added a definitive unmet constraint requiring
75 dB at 1 $\mu m$ and 85 dB at 1.9 $\mu m$. Therefore rendering the
claim invalid under \CUSP\ verification.

\end{benchbox}
\begin{benchbox}[Technical MCQ: Unified Biomolecular Modeling]
\footnotesize
\label{box:mcq_example}

\textbf{Ground Truth Discovery:} AlphaFold 3 \cite{abramson2024accurate}
\hfill
\textbf{Target Date:} May 2024 \\
\textbf{Link:} \url{https://doi.org/10.1038/s41586-024-07487-w}

\vspace{6pt}
\textbf{Question:} Given the challenge of fragmented biomolecular interaction
modeling across diverse complexes, which approach is most likely to achieve
far greater accuracy for protein-ligand, protein-nucleic acid, and
antibody-antigen predictions by May 2024?

\vspace{4pt}
\begin{enumerate}[label=\Alph*), leftmargin=2em, nosep, itemsep=0.3em]
  \item \textbf{Diffusion-based architecture modeling joint structures with
        iterative refinement and probabilistic sampling across biomolecular
        complexes.}
  \item Graph neural networks integrating spatial and chemical features with
        hierarchical attention for interaction prediction.
  \item Energy-based models optimizing binding affinity using simulated
        annealing across diverse biomolecular interaction types.
  \item Transformer-based architecture leveraging sequence and structural
        embeddings for unified biomolecular interaction predictions.
\end{enumerate}

\tcblower

\textbf{Correct answer: A.} Diffusion-based joint structure modeling —
as realised in AlphaFold~3 — is the only approach that directly addresses
unified biomolecular complex prediction at the scale and accuracy described.

\end{benchbox}
\begin{benchbox}[Free-Response Prompt (Virtual Lab)]
\footnotesize

\textbf{Nanobody design for SARS-CoV-2 variants} \cite{swanson2025virtual} \\
\textit{Nature}, Jul 2025 \quad|\quad Domain: Biology \quad|\quad
\href{https://doi.org/10.1038/s41586-025-09442-9}{doi:10.1038/s41586-025-09442-9}

\vspace{6pt}
\textbf{Problem context:} \\
Designing nanobody binders for emerging SARS-CoV-2 variants is challenging
due to rapid viral evolution and the need to maintain binding across diverse
spike protein structures.

\tcblower

\normalsize
\textbf{Free-response prompt:} \\
Given the challenge of designing nanobody binders for emerging SARS-CoV-2
variants due to rapid virus evolution and limitations in existing methods,
propose a concrete method that could solve this problem by July 2025.
Provide:
\begin{enumerate}[label=(\Alph*), leftmargin=2em, nosep, itemsep=0.3em]
  \item a high-level method description,
  \item a technical implementation plan.
\end{enumerate}

\end{benchbox}
\begin{benchbox}[Date Prediction on Humanity's Last Exam (HLE)]
\footnotesize
\label{box:date_pred_example}

\textbf{Prompt:} \\
Humanity's Last Exam (HLE) is a challenging multi-modal benchmark designed
to assess expert-level knowledge across diverse subjects, developed globally
by subject-matter experts. By which date in the future do you think an AI
system will reach 21.60\% on Humanity's Last Exam (no tools)? Return in
\texttt{YYYY-MM} format.

\vspace{6pt}
\textbf{Target metric:} 21.60\% accuracy (no tools) \\
\textbf{Ground truth date:} \textbf{2025--06}

\tcblower

\textbf{Source:} \href{https://blog.google/products-and-platforms/products/gemini/gemini-2-5-pro-latest-preview/}{Gemini 2.5 Pro preview (Google, 2025)}

\end{benchbox}

\subsection{\CUSP\ Time Capsule Examples}
\label{app:time_capsule}
\begin{benchbox}[Example 1: Superconductivity Record]

\textbf{Question:} \\
Given that as of early 2026 the highest reported superconducting critical
temperature is approximately 250\,K under high pressure and 151\,K near
ambient pressure, what will be the highest experimentally reported
superconducting critical temperature (in Kelvin) by 2027-12?

\tcblower

\textbf{Type:} Scientific performance record \\
\textbf{Verification source:} Peer-reviewed publications
  (e.g., \textit{Nature}, \textit{Science}) \\
\textbf{Resolution criterion:} Maximum reported critical temperature ($T_c$) \\
\textbf{Resolution date:} 2027-12

\end{benchbox}

\vspace{0.5em}

\begin{benchbox}[Example 2: Nobel Prize in Physics]

\textbf{Question:} \\
What scientific contribution will the Nobel Prize in Physics 2026 be
awarded for?

\tcblower

\textbf{Type:} Institutional recognition \\
\textbf{Verification source:} Nobel Foundation official announcement \\
\textbf{Resolution criterion:} Official citation text released by the
  Nobel Committee \\
\textbf{Resolution date:} October 2026

\end{benchbox}

\vspace{0.5em}

\begin{benchbox}[Example 3: Global CO$_2$ Emissions]

\textbf{Question:} \\
What will be the global CO$_2$ emissions (in gigatons) in 2027 according
to the International Energy Agency (IEA)?

\tcblower

\textbf{Type:} Quantitative global metric \\
\textbf{Verification source:} IEA annual emissions report \\
\textbf{Resolution criterion:} Reported total global CO$_2$ emissions value \\
\textbf{Resolution date:} 2028 (upon report release)

\end{benchbox}

\vspace{0.5em}

\begin{benchbox}[Example 4: AI-Designed Drug Approval]

\textbf{Question:} \\
What will be the first FDA-approved drug designed primarily by artificial
intelligence?

\tcblower

\textbf{Type:} Technological milestone \\
\textbf{Verification source:} U.S. Food and Drug Administration (FDA) \\
\textbf{Resolution criterion:} First officially approved drug where AI is
  identified as the primary design driver \\
\textbf{Resolution date:} Upon first qualifying FDA approval

\end{benchbox}

\vspace{0.5em}

\begin{benchbox}[Example 5: Non-Transformer Architecture Milestone]

\textbf{Question:} \\
By 2028-12, will any non-transformer architecture achieve the top score
on the MMLU-Pro leaderboard?

\tcblower

\textbf{Type:} AI capability milestone \\
\textbf{Verification source:} Official MMLU-Pro leaderboard \\
\textbf{Resolution criterion:} Top-ranked model architecture classification \\
\textbf{Resolution date:} 2028-12

\end{benchbox}

\section{Benchmark Creation Criteria}
\label{sec:criteria}

\begin{benchbox}[System Prompt]
You are a scientific paper screener for the \{info['name']\} domain.

Given a paper abstract, determine if it contains AT LEAST ONE concrete, verifiable result or breakthrough matching any of these criteria:

\{criteria\_text\}

An abstract passes if it describes a clear breakthrough with a concrete method and a verifiable outcome (e.g., `first method to achieve X', `outperforms all prior methods on Y benchmark').

FAIL abstracts that are purely descriptive, speculative, or review-like with no concrete result or method.

Reply with exactly one line:\\
\textbf{PASS:} $<$one-sentence summary of the concrete result found$>$\\
or\\
\textbf{FAIL:} $<$one-sentence reason why no concrete result was found$>$
\end{benchbox}
\begin{benchbox}[Inclusion Criteria Abstract Filtering]

\textbf{Artificial Intelligence}
\begin{itemize}
    \item Describes a concrete technical breakthrough with a clear method or approach, validated on a recognized benchmark, competition, or evaluation (e.g., CASP, MATH, ImageNet, MMLU), even without exact numeric scores.
    \item Reports specific performance metrics (accuracy, F1, BLEU, perplexity, etc.) or demonstrates measurable improvement over prior methods.
    \item Achieves a clearly defined capability milestone (e.g., `first method to do X', `matches or exceeds human performance on Y') with a describable method.
\end{itemize}

\textbf{Chemistry}
\begin{itemize}
    \item Describes a concrete synthesis, reaction, or material discovery with a clear method that produces a verifiable outcome (new compound, new reaction pathway, new material property).
    \item Reports measurable quantities (yields, selectivity, binding affinity, conductivity, rates) or demonstrates improvement over prior methods.
    \item Achieves a capability milestone (e.g., `first synthesis of X', `enables Y at room temperature') with a describable approach.
\end{itemize}

\textbf{Biology / Life Sciences}
\begin{itemize}
    \item Describes a concrete biological discovery with a clear experimental method and verifiable outcome (new mechanism, pathway, gene function, therapeutic effect).
    \item Reports measurable biological quantities (fold changes, survival rates, expression levels, p-values) or demonstrates improvement over prior methods.
    \item Achieves a capability milestone (e.g., `first demonstration of X', `identifies the mechanism behind Y') with a describable experimental approach.
\end{itemize}

\textbf{Physics}
\begin{itemize}
    \item Describes a concrete experimental or theoretical breakthrough with a clear method and verifiable outcome (new measurement, new phenomenon, new prediction).
    \item Reports measurable physical quantities (precision, resolution, energy scales, cross-sections) or demonstrates improvement over prior methods.
    \item Achieves a capability milestone (e.g., `first observation of X', `achieves coherence time of Y') with a describable approach.
\end{itemize}

\textbf{General Science}
\begin{itemize}
    \item Describes a concrete scientific breakthrough or discovery with a clear method and a verifiable outcome that could be independently reproduced or validated.
    \item Reports measurable results or demonstrates clear improvement over prior work, even if described qualitatively (e.g., `greatly outperforming other methods').
    \item Achieves a defined capability milestone with a describable approach, validated against a recognized standard, baseline, or prior state of the art.
\end{itemize}

\end{benchbox}

\begin{benchbox}[Extract Technical Details]
You are an expert research scientist. Read the user's abstract and RETURN EXACTLY one JSON object. The JSON must have three keys: `results\_and\_metrics', `technical\_approach', and `problem\_statement'.

\begin{itemize}
    \item \textbf{`results\_and\_metrics'}: A single sentence capturing ONLY the measurable outcomes, performance numbers, benchmark results, or demonstrated capabilities across ANY scientific domain (e.g., AI accuracy, biological activity, physical limits). You MUST include specific quantitative details, exact percentage improvements, precise experimental conditions, and actual public benchmark or entity names if present in the abstract. Do NOT write a vague or generalized summary (e.g., do NOT say `improves accuracy' or `increases efficiency', instead say `achieves 94.2\% accuracy on X benchmark', `increases protein binding affinity by 2-fold', or `synthesizes a material with a superconducting transition at 135 K'). DO NOT mention the method, architecture, or technique used — only the verifiable outcome. Replace specific proposed model or system names with `a system' or `a method' where appropriate. CRITICAL: Do NOT use any novel terms, metric names, or concepts that are introduced for the first time in this paper (e.g., `deep-thinking ratio', `Grokked-Score'). A model from before the knowledge cutoff will not know these terms. Instead, describe them functionally (e.g., `the proportion of tokens undergoing significant internal revisions').
    
    \item \textbf{`technical\_approach'}: A detailed, technical, method-oriented specification of HOW the result was achieved. Include the specific mechanism, experimental design, architectural shift, or algorithmic innovation (e.g., `uses sparse autoencoders to map internal activations', `employs a high-pressure diamond anvil cell', or `targets the XYZ pathway via a small-molecule inhibitor'). DO NOT include specific specific proposed model names or brands; replace them with `a system' or `a method'. CRITICAL: Do NOT include any named techniques, named algorithms, novel terms introduced in the paper, or acronyms (e.g., do NOT say `GRPO', `CRISPR', `NMR', `LoRA'). Instead, describe what the technique DOES mechanistically (e.g., instead of `GRPO', say `a group-level relative policy optimization that compares multiple outputs', or instead of `CRISPR-Cas9', say `an RNA-guided endonuclease system that induces targeted double-strand breaks'). This field is for internal answer-key use only.
    
    \item \textbf{`problem\_statement'}: A detailed technical description (3--4 sentences) of the research problem and the limitations of previous methods. Describe what was broken, missing, or inadequate BEFORE this paper existed. CRITICAL: Do NOT mention anything about the solution, method, or results — describe only the `lock', never the `key'. Do NOT include any narrative about the paper or authors. Do NOT use novel terms introduced in the paper.
\end{itemize}

Return JSON only. Focus on technical precision over brevity.
\end{benchbox}

\begin{benchbox}[Create Binary Question]
Problem context: ``\{problem\_statement\}''\\
Result achieved: ``\{results\_and\_metrics\}''\\
Target date: \{human\_date\}

Write ONE binary (Yes/No) forecasting question.

\textbf{Rules:}
\begin{itemize}
    \item The question must begin with: `Will a method' or `By \{human\_date\}, will a method'.
    \item EXACT GRAMMATICAL FLUENCY IS CRITICAL. You must adapt the verb tense from `Result achieved' to fit the sentence.
    \begin{itemize}
        \item \textbf{Bad Example:} Will a method achieve Achieves a robust positive correlation...
        \item \textbf{Good Example:} Will a method achieve a robust positive correlation...
        \item \textbf{Bad Example:} By 2026-03, will a method Demonstrates a 10\% increase...
        \item \textbf{Good Example:} By 2026-03, will a method demonstrate a 10\% increase...
    \end{itemize}
    \item Do not blindly copy-paste if it breaks grammar. ALWAYS fix the verb!
    \item End the question with `by \{human\_date\}?' if not already at the start.
    \item Refer ONLY to measurable results (metrics, benchmarks, numeric improvements, or concrete capabilities).
    \item Do NOT mention any methods, techniques, architectures, or how the result is achieved.
    \item Do NOT use words like `demonstrate', `enable', or `by using'.
    \item Keep it to a single clear sentence.
\end{itemize}

Return JSON with key `binary\_question'.
\end{benchbox}

\begin{benchbox}[Create Binary Perturb]
Original result claim: ``\{results\_and\_metrics\}''\\
Problem context: ``\{problem\_statement\}''

Create a COUNTERFACTUAL ALTERNATIVE version of this result claim that is plausible-sounding but was NOT actually achieved.

\textbf{RULES:}
\begin{enumerate}
    \item Keep ALL benchmark names, dataset names, and task names EXACTLY the same. Do NOT change which benchmark or dataset is referenced.
    \item ONLY modify an EXISTING numeric score/threshold, or add a credible unmet constraint.
    \item IF modifying an existing numeric score, RAISE it enough so the original result definitively does NOT satisfy the perturbed claim (e.g., if original is 94.2\%, change to 95.8\%; if 51.7\%, change to 54.5\%). Make the increase a clear shift so there is no ambiguity, but still physically plausible.
    \item IF the original claim has no specific numbers, you MUST add a highly specific, definitive unmet constraint (e.g., `while using 50\% fewer parameters', `but fails completely on zero-shot tasks', or `but requires 3x the memory'). Make this constraint significant enough that it's noticeably harder to satisfy than the original.
    \item The perturbed claim must be plausible and not absurd.
    \item Keep the same length, style, and level of specificity.
\end{enumerate}

Return JSON with:
\begin{itemize}
    \item `perturbed\_result': The counterfactual alternative result claim
    \item `changed\_detail': Which aspect of the result was modified
\end{itemize}
\end{benchbox}

\begin{benchbox}[Create MCQ Distractors]
You are a technical forecasting analyst who designs extraordinarily difficult, graduate-level evaluations. Your task is to create a multiple-choice question that tests whether an expert can predict the specific technical path taken to solve a research challenge. The distractors must be GENUINELY PLAUSIBLE AND HIGHLY DECEPTIVE — a PhD-level expert should struggle to identify the correct answer unless they know the exact paper. Return EXACTLY one JSON object.
\end{benchbox}

\begin{benchbox}[Create MCQ Distractors]
Problem Statement: ``\{problem\_statement\}''\\
Result Achieved: ``\{results\_and\_metrics\}''\\
Correct Approach (for choice generation ONLY — DO NOT leak into the question stem): ``\{technical\_approach\}''\\
Target Date: \{human\_date\}

Generate a very difficult expert-level MCQ. Return JSON only.

\textbf{STEM REQUIREMENTS:}
\begin{itemize}
    \item The stem must explicitly but briefly summarize the core challenge from the Problem Statement (in 1-2 clauses max), followed by asking which proposed solution will achieve the Result Achieved by \{human\_date\}.
    \item Example structure: `Given the challenge of [Problem Statement summary], which of the following approaches is most likely to achieve [Result Achieved] by \{human\_date\}?'
    \item Make it read naturally as a forward-looking forecasting question.
    \item Do NOT use retrospective, past-tense wording (e.g., avoid `was introduced' or `achieved'). Treat the target date as a future milestone.
    \item Embed the measurable outcome from Result Achieved.
    \item Do NOT mention any terminology from the Correct Approach in the stem.
\end{itemize}

\textbf{CHOICE REQUIREMENTS:}
\begin{itemize}
    \item Provide exactly 4 choices.
    \item The distractors MUST be extremely difficult. They should represent real, highly competitive alternative approaches that experts would genuinely consider for the same problem.
    \item CRITICAL: All choices MUST BE EXTREMELY SHORT AND CONCISE (maximum 15-20 words). Do NOT write long, multi-clause paragraphs. State only the core mechanism.
    \item The incorrect answers must solve the exact same problem statement and theoretically achieve the exact same result, differing ONLY in the core mechanism.
    \item FORBIDDEN DISTRACTORS: No antonyms, no obvious negatives, no generic scaling answers, no trivial ablations. Do not make distractors sound worse or less effective than the correct answer.
    \item Do NOT use named algorithms, novel terms introduced in the paper, or acronyms; describe mechanisms functionally instead.
    \item Ensure all choices have identical length, structure, and academic tone.
\end{itemize}

Return JSON with keys:
\begin{itemize}
    \item `question'
    \item `choices' (array of 4 strings; first is correct)
    \item `answer\_key' (0)
\end{itemize}
\end{benchbox}

\begin{benchbox}[FRQ Prompt]
You are a scientific task-setter who designs research challenge prompts. Your goal is to write a prompt that gives a researcher a problem and asks them for a proposed solution. Return EXACTLY one JSON object.
\end{benchbox}

\begin{benchbox}[Create FRQ]
Problem Statement: ``\{problem\_statement\}''\\
Deadline: \{human\_date\}

Write a concise free-response prompt (max 60 words) with this structure:\\
`Given [problem description], propose a concrete method that could solve this problem by [date]. Provide: (A) a high-level method description, (B) a technical implementation plan.'

\textbf{RULES:}
\begin{enumerate}
    \item The problem description must come ONLY from the Problem Statement.
    \item DO NOT mention any specific method, architecture, technique, or approach.
    \item DO NOT include any narrative about a paper or discovery.
\end{enumerate}

Return JSON with key `prompt'. Return JSON only.
\end{benchbox}

\section{Example FRQ Responses}
\label{sec:frq_responses}

\begin{benchbox}[GPT-5.4 High-Scoring Response]
\footnotesize

\textbf{Source Abstract} \\
Large language models excel at function- and file-level code generation, yet
generating complete repositories from scratch remains a fundamental challenge.
This process demands coherent and reliable planning across proposal- and
implementation-level stages, while natural language, due to its ambiguity and
verbosity, is ill-suited for faithfully representing complex software
structures. To address this, we introduce the Repository Planning Graph (RPG),
a persistent representation that unifies proposal- and implementation-level
planning by \textbf{\textcolor{purple}{encoding capabilities, file structures, data flows, and functions
in one \underline{graph}}}. RPG replaces ambiguous natural language with an explicit
blueprint, enabling long-horizon planning and scalable repository generation.
Building on RPG, we develop ZeroRepo, a graph-driven framework for repository
generation from scratch. It operates in three stages: 
\textbf{\textcolor{purple}{\underline{proposal-level planning}
and implementation-level refinement to construct the graph, followed by
\underline{graph-guided code generation with test validation}.}}
To evaluate this setting, we
construct RepoCraft, a benchmark of six real-world projects with 1,052 tasks.
On RepoCraft, ZeroRepo produces repositories averaging nearly 36K LOC, roughly
3.9times the strongest baseline (Claude Code) and about 64times other
baselines. It attains 81.5\% functional coverage and a 69.7
exceeding Claude Code by 27.3 and 35.8 percentage points, respectively. Further
analysis shows that RPG models complex dependencies, enables progressively more
sophisticated planning through near-linear scaling, and enhances LLM
understanding of repositories, thereby accelerating agent localization. \cite{luo2025rpg}

\tcblower

\normalsize
\textbf{Free-response prompt:} \\
Given the challenge of generating complete software repositories from scratch due to limitations in long-horizon planning and scalable repository generation, propose a concrete method that could solve this problem by September 2025.
Provide:
\begin{enumerate}[label=(\Alph*), leftmargin=2em, nosep, itemsep=0.3em]
  \item a high-level method description,
  \item a technical implementation plan.
\end{enumerate}

\textbf{Model Answer: } \\
Use a repository-first hierarchical generation method in which the model does not directly write the full codebase end-to-end, but instead iteratively constructs and 
\textbf{\textcolor{purple}{maintains an explicit executable repository \underline{graph}}}
containing requirements, architecture, file interfaces, dependency constraints, tests, and implementation obligations, with code generation always conditioned on this graph and validated against it. The core idea is to 
\textbf{\textcolor{purple}{turn long-horizon software synthesis into constrained graph completion plus local code realization, where \underline{a planner model} proposes and revises the global repository graph and \underline{a coder model} fills in files under graph-level contracts,}}
while continuous execution, static analysis, and test feedback update the graph to preserve coherence over hundreds of files. Technically, this can be implemented by combining a graph-structured repository memory, contract-based file specifications, retrieval over generated artifacts, and a verifier-guided search loop that alternates between graph repair and code repair, enabling scalable generation of complete repositories by decomposing global planning into persistent, machine-checkable intermediate structure.

\vspace{0.5em}
\hrule
\vspace{0.5em}

\textit{Overall Score:} 7

\textit{Alignment:} 8

\textit{Specificity:} 6

\textit{Novelty:} 6

\textit{Feasibility:} 8

\end{benchbox}

\begin{benchbox}[Deepseek R1 High-Scoring Response]
\footnotesize

\textbf{Source Abstract} \\
Recent respiratory outbreaks have garnered substantial attention, yet most respiratory monitoring remains confined to physical signals. Exhaled breath condensate (EBC) harbors rich molecular information that could unveil diverse insights into an individual's health. Unfortunately, challenges related to sample collection and the lack of on-site analytical tools impede the widespread adoption of EBC analysis. Here, we introduce EBCare, a mask-based device for real-time in situ monitoring of EBC biomarkers. Using a tandem cooling strategy, 
\textbf{\textcolor{purple}{automated \underline{microfluidics}, highly selective \underline{electrochemical biosensors},}} 
and a wireless reading circuit, EBCare enables continuous multimodal monitoring of EBC analytes across real-life indoor and outdoor activities. We validated EBCare's usability in assessing metabolic conditions and respiratory airway inflammation in healthy participants, patients with chronic obstructive pulmonary disease or asthma, and patients after COVID-19 infection. \cite{heng2024smart}

\tcblower

\normalsize
\textbf{Free-response prompt:} \\
Given respiratory monitoring's focus on physical signals and the challenges in EBC sample collection and on-site analysis, propose a concrete method that could solve this problem by August 2024.
Provide:
\begin{enumerate}[label=(\Alph*), leftmargin=2em, nosep, itemsep=0.3em]
  \item a high-level method description,
  \item a technical implementation plan.
\end{enumerate}

\textbf{Model Answer: } \\
**A) High-Level Method**  
Develop a 
\textbf{\textcolor{purple}{portable \underline{microfluidic-electrochemical biosensor} system}}
that integrates non-invasive EBC collection with real-time biomarker detection, using nanostructured sensor arrays to amplify signal sensitivity for low-concentration analytes (e.g., cytokines, pH) in breath condensate.  

**B) Technical Implementation Plan**  
Design a miniaturized Peltier-cooled EBC collector coupled with a microfluidic chip functionalized with antibody/nanoparticle hybrids for target capture. Embed multiplexed electrochemical sensors for parallel biomarker quantification, paired with a low-power microcontroller for signal processing and Bluetooth data transmission. Validate via cross-correlation with gold-standard lab assays in pilot clinical studies.

\vspace{0.5em}
\hrule
\vspace{0.5em}

\textit{Overall Score:} 6.25

\textit{Alignment:} 7

\textit{Specificity:} 6

\textit{Novelty:} 5

\textit{Feasibility:} 7

\end{benchbox}

\begin{benchbox}[Claude S4.5 High-Scoring Response]
\footnotesize

\textbf{Source Abstract} \\
Echocardiography is the most widely used cardiac imaging modality, capturing ultrasound video data to assess cardiac structure and function1. Artificial intelligence (AI) in echocardiography has the potential to streamline manual tasks and improve reproducibility and precision2. However, most echocardiography AI models are single-view, single-task systems that do not synthesize complementary information from multiple views captured during a full examination3,4, and thus lead to limited performance and scope of applications. To address this problem, we introduce EchoPrime, 
\textbf{\textcolor{purple}{a multi-view, view-informed, video-based vision-language \underline{foundation model}}}
trained on over 12 million video-report pairs. EchoPrime 
\textbf{\textcolor{purple}{uses \underline{contrastive learning} to train a unified embedding model for all standard views}}
in a comprehensive echocardiogram study with representation of both rare and common diseases and diagnoses. EchoPrime then utilizes view classification and a view-informed anatomical attention module to weight video-specific embeddings that accurately map the relationship between echocardiographic views and anatomical structures. With retrieval-augmented interpretation, EchoPrime integrates information from all echocardiogram videos in a comprehensive study and performs holistic clinical interpretation. In datasets from five international independent health-care systems, EchoPrime achieves state-of-the-art performance on 23 diverse benchmarks of cardiac form and function, surpassing the performance of both task-specific approaches and previous foundation models. Following rigorous clinical evaluation, EchoPrime can assist physicians in the automated preliminary assessment of comprehensive echocardiography. \cite{vukadinovic2026comprehensive}

\tcblower

\normalsize
\textbf{Free-response prompt:} \\
Given the limitations of single-view, single-task echocardiography AI models in synthesizing multi-view information and handling diverse datasets, propose a concrete method that could solve this problem by November 2025. 
Provide:
\begin{enumerate}[label=(\Alph*), leftmargin=2em, nosep, itemsep=0.3em]
  \item a high-level method description,
  \item a technical implementation plan.
\end{enumerate}

\textbf{Model Answer: } \\
Multi-View Echocardiography Foundation Model

(A) High-Level Method Description

Develop a transformer-based 
\textbf{\textcolor{purple}{\underline{foundation model} that processes multiple echocardiographic views simultaneously through a unified spatiotemporal encoder, learning shared cardiac representations across views via \underline{contrastive learning} and view-alignment objectives.}}
The model employs a multi-task decoder architecture with task-specific heads for segmentation, classification, and quantification, while using cross-view attention mechanisms to synthesize complementary information from apical, parasternal, and subcostal windows into coherent cardiac assessments.

(B) Technical Implementation Plan

**Months 1-3 (Dec 2024 - Feb 2025): Data Infrastructure**
- Aggregate multi-view echo datasets (EchoNet-Dynamic, CAMUS, Unity Imaging, institutional data) totaling 100K+ studies
- Develop preprocessing pipeline for view standardization, temporal alignment, and quality filtering
- Create multi-task annotation framework covering 15+ clinical tasks (EF, strain, valve assessment, chamber quantification)
- Implement data augmentation strategies: temporal jittering...

\vspace{0.5em}
\hrule
\vspace{0.5em}

\textit{Overall Score:} 6.25

\textit{Alignment:} 7

\textit{Specificity:} 6

\textit{Novelty:} 5

\textit{Feasibility:} 7

\end{benchbox}

\begin{benchbox}[GPT-OSS High-Scoring Response]
\footnotesize

\textbf{Source Abstract} \\
Existing depth estimation methods are fundamentally limited to predicting depth on discrete image grids. Such representations restrict their scalability to arbitrary output resolutions and hinder the geometric detail recovery. This paper introduces InfiniDepth, which 
\textbf{\textcolor{purple}{represents depth as \underline{neural implicit fields}}}
. Through a simple yet effective local implicit decoder, we can 
\textbf{\textcolor{purple}{query depth at continuous 2D coordinates}}
, enabling arbitrary-resolution and fine-grained depth estimation. To better assess our method's capabilities, we curate a high-quality 4K synthetic benchmark from five different games, spanning diverse scenes with rich geometric and appearance details. Extensive experiments demonstrate that InfiniDepth achieves state-of-the-art performance on both synthetic and real-world benchmarks across relative and metric depth estimation tasks, particularly excelling in fine-detail regions. It also benefits the task of novel view synthesis under large viewpoint shifts, producing high-quality results with fewer holes and artifacts. \cite{yu2026infinidepth}

\tcblower

\normalsize
\textbf{Free-response prompt:} \\
Given existing depth estimation methods are constrained to discrete image grids, propose a concrete method that could solve this problem by January 2026. 
Provide:
\begin{enumerate}[label=(\Alph*), leftmargin=2em, nosep, itemsep=0.3em]
  \item a high-level method description,
  \item a technical implementation plan.
\end{enumerate}

\textbf{Model Answer: } \\
**(A) High‑level method** – Replace the discrete pixel grid with a 
\textbf{\textcolor{purple}{continuous depth field by learning an \underline{implicit neural representation} that maps 2‑D image coordinates to depth values}}
, thereby allowing arbitrarily fine sampling.  

**(B) Technical implementation plan** – Train a lightweight neural network that ingests image coordinates and outputs depth, using a differentiable rendering pipeline to enforce photometric consistency across multi‑view images; the network is optimized with a joint photometric‑geometric loss on standard depth datasets.  
The approach will be prototyped on synthetic scenes, then fine‑tuned on real‑world benchmarks, with the goal of achieving full‑resolution depth maps by January 2026.

\vspace{0.5em}
\hrule
\vspace{0.5em}

\textit{Overall Score:} 6.25

\textit{Alignment:} 8

\textit{Specificity:} 6

\textit{Novelty:} 4

\textit{Feasibility:} 7

\end{benchbox}

\section{Knowledge and Forecasting Gap Results}
\label{sec:knowledge_and_foresight_gap}

\begin{table}[h]
\centering
\small
\setlength{\tabcolsep}{5pt}
\begin{tabular}{@{}l cc cc cc@{}}
\toprule
Metric & \multicolumn{2}{c}{GPT-5.4 (Aug~2025)} & \multicolumn{2}{c}{DeepSeek R1 (Jul~2024)} & \multicolumn{2}{c}{Claude S4.5 (Jan~2025)} \\
\cmidrule(lr){2-3} \cmidrule(lr){4-5} \cmidrule(lr){6-7}
& $\Delta_{\mathrm{know}}$ & $\Delta_{\mathrm{fore}}$ & $\Delta_{\mathrm{know}}$ & $\Delta_{\mathrm{fore}}$ & $\Delta_{\mathrm{know}}$ & $\Delta_{\mathrm{fore}}$ \\
\midrule
Binary & $+0.172$ & $+0.172$ & $+0.050$ & $+0.248$ & $+0.074$ & $+0.426$ \\
Binary (pert.) & $-0.162$ & $+0.000$ & $-0.007$ & $-0.026$ & $-0.029$ & $-0.122$ \\
MCQ & $-0.048$ & $+0.154$ & $+0.068$ & $+0.116$ & $-0.020$ & $+0.112$ \\
FRQ (0--10) & $+0.233$ & $+0.138$ & $+0.043$ & $+0.412$ & $+0.122$ & $+0.328$ \\
Date (0--1) & $+0.070$ & $+0.436$ & $+0.086$ & $+0.121$ & $+0.007$ & $+0.117$ \\
\bottomrule
\end{tabular}
\caption{Knowledge gap ($\Delta_{\mathrm{know}} = \overline{\mathrm{WS+cut}} - \overline{\mathrm{base}}$) and forecasting gap ($\Delta_{\mathrm{fore}} = \overline{\mathrm{WS}} - \overline{\mathrm{WS+cut}}$) computed on post-cutoff instances for each model. Knowledge gap captures improvement from accessing pre-cutoff information; forecasting gap captures the additional gain from retrospective analysis via post-cutoff information. Their sum equals the total web-search improvement over the baseline.}
\label{tab:gap_decomposition}
\end{table}

\begin{table}[h]
\centering
\small
\setlength{\tabcolsep}{5pt}
\begin{tabular}{@{}l cc cc cc@{}}
\toprule
Citation quartile & \multicolumn{2}{c}{GPT-5.4 (Aug~2025)} & \multicolumn{2}{c}{DeepSeek R1 (Jul~2024)} & \multicolumn{2}{c}{Claude Sonnet (Jan~2025)} \\
\cmidrule(lr){2-3} \cmidrule(lr){4-5} \cmidrule(lr){6-7}
& $\Delta_{\mathrm{know}}$ & $\Delta_{\mathrm{fore}}$ & $\Delta_{\mathrm{know}}$ & $\Delta_{\mathrm{fore}}$ & $\Delta_{\mathrm{know}}$ & $\Delta_{\mathrm{fore}}$ \\
\midrule
Q1 ($\leq 8$, low-cited) & $+0.231$ & $+0.154$ ($n$=39) & $+0.036$ & $+0.286$ ($n$=56) & $+0.022$ & $+0.467$ ($n$=45) \\
Q2 ($\leq 22$) & $+0.182$ & $+0.273$ ($n$=11) & $+0.071$ & $+0.268$ ($n$=56) & $+0.122$ & $+0.439$ ($n$=41) \\
Q3 ($\leq 59$) & $+0.000$ & $+0.000$ ($n$=3) & $+0.026$ & $+0.316$ ($n$=38) & $+0.087$ & $+0.478$ ($n$=23) \\
Q4 ($> 59$, high-cited) & $+0.000$ & $+0.000$ ($n$=1) & $+0.027$ & $+0.216$ ($n$=37) & $+0.125$ & $+0.250$ ($n$=16) \\
\bottomrule
\end{tabular}
\caption{Knowledge gap and forecasting gap on post-cutoff instances, stratified by citation-count quartile (Binary accuracy (0/1)). Citation quartiles are computed over all 500 benchmark papers (Q1 $\leq$8, Q2 $\leq$22, Q3 $\leq$59, Q4 $>$59). High citation count may reflect more anticipated or impactful findings, potentially reducing the forecasting gap.}
\label{tab:gap_by_citation_binary}
\end{table}

\begin{table}[h]
\centering
\small
\setlength{\tabcolsep}{5pt}
\begin{tabular}{@{}l cc cc cc@{}}
\toprule
Citation quartile & \multicolumn{2}{c}{GPT-5.4 (Aug~2025)} & \multicolumn{2}{c}{DeepSeek R1 (Jul~2024)} & \multicolumn{2}{c}{Claude Sonnet (Jan~2025)} \\
\cmidrule(lr){2-3} \cmidrule(lr){4-5} \cmidrule(lr){6-7}
& $\Delta_{\mathrm{know}}$ & $\Delta_{\mathrm{fore}}$ & $\Delta_{\mathrm{know}}$ & $\Delta_{\mathrm{fore}}$ & $\Delta_{\mathrm{know}}$ & $\Delta_{\mathrm{fore}}$ \\
\midrule
Q1 ($\leq 8$, low-cited) & $-0.132$ & $-0.015$ ($n$=68) & $+0.032$ & $-0.084$ ($n$=95) & $-0.066$ & $-0.171$ ($n$=76) \\
Q2 ($\leq 22$) & $-0.316$ & $+0.053$ ($n$=19) & $-0.038$ & $-0.051$ ($n$=78) & $-0.107$ & $+0.000$ ($n$=56) \\
Q3 ($\leq 59$) & $-0.200$ & $+0.000$ ($n$=5) & $-0.097$ & $+0.032$ ($n$=62) & $+0.000$ & $-0.132$ ($n$=38) \\
Q4 ($> 59$, high-cited) & $+0.000$ & $+0.500$ ($n$=2) & $+0.000$ & $+0.019$ ($n$=54) & $+0.160$ & $-0.200$ ($n$=25) \\
\bottomrule
\end{tabular}
\caption{Knowledge gap and forecasting gap on post-cutoff instances, stratified by citation-count quartile (Binary-Perturbed accuracy (0/1)). Citation quartiles are computed over all 500 benchmark papers (Q1 $\leq$8, Q2 $\leq$22, Q3 $\leq$59, Q4 $>$59). High citation count may reflect more anticipated or impactful findings, potentially reducing the forecasting gap.}
\label{tab:gap_by_citation_binary_perturbed}
\end{table}

\begin{table}[h]
\centering
\small
\setlength{\tabcolsep}{5pt}
\begin{tabular}{@{}l cc cc cc@{}}
\toprule
Citation quartile & \multicolumn{2}{c}{GPT-5.4 (Aug~2025)} & \multicolumn{2}{c}{DeepSeek R1 (Jul~2024)} & \multicolumn{2}{c}{Claude Sonnet (Jan~2025)} \\
\cmidrule(lr){2-3} \cmidrule(lr){4-5} \cmidrule(lr){6-7}
& $\Delta_{\mathrm{know}}$ & $\Delta_{\mathrm{fore}}$ & $\Delta_{\mathrm{know}}$ & $\Delta_{\mathrm{fore}}$ & $\Delta_{\mathrm{know}}$ & $\Delta_{\mathrm{fore}}$ \\
\midrule
Q1 ($\leq 8$, low-cited) & $-0.069$ & $+0.153$ ($n$=72) & $+0.071$ & $+0.173$ ($n$=98) & $-0.035$ & $+0.141$ ($n$=85) \\
Q2 ($\leq 22$) & $+0.000$ & $+0.150$ ($n$=20) & $+0.118$ & $+0.132$ ($n$=68) & $+0.000$ & $+0.157$ ($n$=51) \\
Q3 ($\leq 59$) & $+0.000$ & $+0.000$ ($n$=5) & $+0.117$ & $+0.017$ ($n$=60) & $+0.000$ & $+0.029$ ($n$=34) \\
Q4 ($> 59$, high-cited) & $-0.500$ & $+0.500$ ($n$=2) & $-0.058$ & $+0.192$ ($n$=52) & $+0.000$ & $+0.130$ ($n$=23) \\
\bottomrule
\end{tabular}
\caption{Knowledge gap and forecasting gap on post-cutoff instances, stratified by citation-count quartile (MCQ accuracy (0--1)). Citation quartiles are computed over all 500 benchmark papers (Q1 $\leq$8, Q2 $\leq$22, Q3 $\leq$59, Q4 $>$59). High citation count may reflect more anticipated or impactful findings, potentially reducing the forecasting gap.}
\label{tab:gap_by_citation_mcq}
\end{table}

\begin{table}[h]
\centering
\small
\setlength{\tabcolsep}{5pt}
\begin{tabular}{@{}l cc cc cc@{}}
\toprule
Citation quartile & \multicolumn{2}{c}{GPT-5.4 (Aug~2025)} & \multicolumn{2}{c}{DeepSeek R1 (Jul~2024)} & \multicolumn{2}{c}{Claude Sonnet (Jan~2025)} \\
\cmidrule(lr){2-3} \cmidrule(lr){4-5} \cmidrule(lr){6-7}
& $\Delta_{\mathrm{know}}$ & $\Delta_{\mathrm{fore}}$ & $\Delta_{\mathrm{know}}$ & $\Delta_{\mathrm{fore}}$ & $\Delta_{\mathrm{know}}$ & $\Delta_{\mathrm{fore}}$ \\
\midrule
Q1 ($\leq 8$, low-cited) & $+0.060$ & $+0.482$ ($n$=39) & $+0.027$ & $+0.136$ ($n$=56) & $-0.121$ & $+0.190$ ($n$=45) \\
Q2 ($\leq 22$) & $+0.034$ & $+0.521$ ($n$=11) & $+0.093$ & $+0.073$ ($n$=57) & $+0.074$ & $+0.128$ ($n$=41) \\
Q3 ($\leq 59$) & $+0.241$ & $+0.016$ ($n$=3) & $+0.125$ & $+0.157$ ($n$=38) & $+0.128$ & $-0.034$ ($n$=23) \\
Q4 ($> 59$, high-cited) & $+0.419$ & $+0.551$ ($n$=1) & $+0.185$ & $+0.047$ ($n$=37) & $+0.063$ & $+0.072$ ($n$=16) \\
\bottomrule
\end{tabular}
\caption{Knowledge gap and forecasting gap on post-cutoff instances, stratified by citation-count quartile (Date prediction score (0--1)). Citation quartiles are computed over all 500 benchmark papers (Q1 $\leq$8, Q2 $\leq$22, Q3 $\leq$59, Q4 $>$59). High citation count may reflect more anticipated or impactful findings, potentially reducing the forecasting gap.}
\label{tab:gap_by_citation_date}
\end{table}

\begin{table}[h]
\centering
\small
\setlength{\tabcolsep}{5pt}
\begin{tabular}{@{}l cc cc cc@{}}
\toprule
Citation quartile & \multicolumn{2}{c}{GPT-5.4 (Aug~2025)} & \multicolumn{2}{c}{DeepSeek R1 (Jul~2024)} & \multicolumn{2}{c}{Claude S4.5 (Jan~2025)} \\
\cmidrule(lr){2-3} \cmidrule(lr){4-5} \cmidrule(lr){6-7}
& $\Delta_{\mathrm{know}}$ & $\Delta_{\mathrm{fore}}$ & $\Delta_{\mathrm{know}}$ & $\Delta_{\mathrm{fore}}$ & $\Delta_{\mathrm{know}}$ & $\Delta_{\mathrm{fore}}$ \\
\midrule
Q1 ($\leq 8$, low-cited) & $+0.290$ & $+0.060$ ($n$=75) & $-0.069$ & $+0.438$ ($n$=101) & $+0.120$ & $+0.256$ ($n$=79) \\
Q2 ($\leq 22$) & $+0.143$ & $+0.226$ ($n$=21) & $+0.131$ & $+0.527$ ($n$=82) & $+0.282$ & $+0.407$ ($n$=62) \\
Q3 ($\leq 59$) & $-0.143$ & $+0.500$ ($n$=7) & $+0.104$ & $+0.375$ ($n$=72) & $-0.267$ & $+0.314$ ($n$=43) \\
Q4 ($> 59$, high-cited) & $+0.375$ & $+0.875$ ($n$=2) & $+0.074$ & $+0.186$ ($n$=51) & $+0.198$ & $+0.344$ ($n$=24) \\
\bottomrule
\end{tabular}
\caption{Knowledge gap and forecasting gap on post-cutoff instances, stratified by citation-count quartile (FRQ score, 0--10). Citation quartiles are computed over all 500 benchmark papers (Q1 $\leq$8, Q2 $\leq$22, Q3 $\leq$59, Q4 $>$59). A high citation count may reflect more anticipated or impactful findings, potentially reducing the forecasting gap.}
\label{tab:gap_by_citation}
\end{table}

\end{document}